# Effective catchment hydrologic representations are low-dimensional but interaction-dependent

Yuan-Heng Wang[1,2]*, Hoshin V. Gupta[2]

[1] Department of Environmental Sciences, University of California, Riverside, Riverside, CA, USA

[2] Department of Hydrology and Atmospheric Sciences, University of Arizona, Tucson, AZ, USA

*Corresponding author: Yuan-Heng Wang*

Email: yuanhenw@ucr.edu | yhwang0730@gmail.com | yhwang0730@arizona.edu

Hydrologic models must reconcile predictive flexibility with physical interpretability, yet the complexity required to reproduce catchment dynamics remains uncertain. Here we compare conceptual hydrologic models and neural networks built from mass-conserving computational units across 513 catchments in the contiguous United States, varying state dimensionality and cross-state information sharing. Under likelihood-based selection, median testing-period Kling–Gupta efficiency skill score ($KGE_{ss}$) increases from 0.83 with one allowable state to 0.89 with two, but reaches only 0.90 with five. KGE-based selection shows similarly rapid saturation. Cross-state information sharing provides an additional source of predictive skill: sharing-enabled candidate spaces restricted to two states outperform non-sharing spaces allowing up to five states, with lower median errors in five of six hydrologic diagnostics. Although likelihood- and KGE-based criteria favor different architectures, both preserve the same structural relationships. Conceptual and neural formulations likewise select different specific architectures but retain the same state dimensionality in 75% of catchments. These findings indicate that effective hydrologic representations are low-dimensional but interaction-dependent, while predictive success alone does not uniquely identify physical storage structures.

Catchment-scale rainfall–runoff models use simplified representations of catchment dynamics to describe how precipitation is stored, transformed and released as streamflow[1–4]. Classical conceptual models typically use fewer than 20 parameters and a small number of state variables, as exemplified by HyMOD[5], GR4J[6] and SAC-SMA[7]. Such parsimonious structures require an appropriate representation of hydrologic complexity across contrasting catchments and locations[8]. Although these models retain substantial conceptual and physical interpretability, the simplified nature of their representations can limit predictive accuracy relative to the now widely used Long Short-Term Memory (LSTM) network[9], which has achieved strong performance for hydrologic prediction across diverse large-sample applications[10–14].

The success of LSTM networks may be partly attributable to their conceptual similarity and functional isomorphism to physical dynamical systems, particularly their ability to maintain and update internal state variables through time[15]. This capability has supported a dominant large-sample modelling paradigm in which heterogeneous information across diverse hydro-geo-climatic conditions is integrated into a single model trained across hundreds of catchments, with the broader ambition of developing models that generalize across continental or even global domains[16–19].

Although this paradigm has substantially advanced practical hydrologic prediction and flood forecasting, the internal representations learned by these networks remain only partially physically-interpretable, motivating increasing efforts to understand their internal states and functions[20]. For example, regional LSTM benchmarks developed using the CAMELS-US dataset[21,22] commonly employ architectures with tens to hundreds of internal nodes trained jointly across hundreds of catchments[23–25]. Even when physical constraints are incorporated into LSTM-based architectures[26–28] or differentiable conceptual models such as HBV[29], a shared multi-state architecture is generally prescribed across catchments. Such designs are therefore not intended to directly answer a more fundamental question: how many internal state variables are actually required to represent the rainfall–runoff dynamics of individual catchments spanning diverse hydro-geo-climatic conditions?

The recently developed mass-conserving perceptron (MCP) modelling paradigm provides a conceptually and physically-interpretable framework for investigating this question at the catchment scale. The MCP can operate as a single computational unit (**Fig. 1a**)[30], be assembled into traditional hydrologic model structures (**Fig. 1b**, left)[31], or be connected into neural-network architectures (**Fig. 1b–e**)[15]. In the MCP, the time-invariant parameters governing fluxes from a finite-capacity, linear-reservoir-type storage are replaced by LSTM-like, time-varying gating functions, while explicitly preserving mass conservation and interpretable internal states. Previous applications further showed that compact one- to two-state representations, when augmented with snow-informed units, can approach the predictive performance of basin-wise LSTM benchmarks[15]. Building on this model–network duality, the present study uses the MCP framework as a controlled basis for investigating how much state-space complexity is actually required to represent catchment-scale rainfall–runoff dynamics.

However, state dimensionality represents only one aspect of hydrologic representational complexity[4,8]. Complexity can also arise from the information exchanged among internal states and from the number of trainable parameters used to describe their dynamics. Increasing any of these dimensions can enhance model flexibility, but additional flexibility need not provide additional independent predictive information[8,32]. In particular, increasing the number of uncoupled states expands the dimensionality of the internal representation, whereas allowing information exchange among states changes how those representations jointly regulate hydrologic fluxes[15]. These forms of complexity are therefore not necessarily interchangeable. Distinguishing their respective contributions is central to determining which aspects of model architecture are necessary for predictive accuracy, which encode additional hydrologic information, and which primarily increase flexibility without materially improving the representation of catchment dynamics.

Snow-affected catchments provide a particularly useful testbed for examining these questions because seasonal snow accumulation and melt can decouple precipitation from runoff and introduce additional storage and interactions among hydrologic states[33–35]. Here, we exploit the MCP model–network duality to investigate the effective complexity of catchment-scale hydrologic representations. We progressively relax architectural constraints from prescribed snow–soil pathways to mass-conserving neural networks built from the same interpretable units, evaluating complexity in terms of state dimensionality, cross-state connectivity and parameterization. We further compare MCP-based representations with basin-wise LSTM architectures trained separately for each catchment, providing a controlled benchmark in which both model families are optimized using the same catchment-specific information. This comparison isolates differences associated with architectural constraints: the MCP imposes mass conservation and physically-interpretable internal states, whereas the LSTM serves as a flexible, non-mass-conserving machine-learning representation that was **not originally designed to provide fully physically-interpretable internal states for geoscientific applications**.

Using 513 CAMELS-US catchments spanning broad gradients in elevation, aridity, snow influence, forest cover and climate regime (**Extended Data Fig. 1**), we ask: (1) how does predictive performance change with state dimensionality; (2) can additional uncoupled states substitute for cross-state information exchange; (3) how do these requirements vary across hydroclimatic regimes; and (4) do the inferred relationships depend on the architecture-selection objective. We test these questions using nested one- to five-state MCP and LSTM representations, alternative connectivity structures, and likelihood- and KGE-based selection.

## Physically-interpretable networks bridge conceptual and neural hydrologic representations

We first examine whether explicitly coupling snow and soil-moisture representations improves streamflow prediction relative to single-state alternatives. The two prescribed HYDROMCP structures represent different hypotheses for how snowmelt reaches streamflow: the routing formulation passes snowmelt through the soil-moisture-like state, whereas the bypass formulation allows snowmelt to contribute directly to streamflow (**Fig. 1b**). Across 513 catchments, selecting the better-performing two-state HYDROMCP formulation improves testing $KGE_{ss}$ relative to single-state HMCP in 279 basins (**Fig. 2a(i)**). Among improving catchments, the mean $KGE_{ss}$ gain is 0.16 across CONUS and 0.36 in the Rocky Mountains. Relative to the SWE-pretrained single-state SNOWMCP, HYDROMCP improves testing $KGE_{ss}$ in 265 basins (**Fig. 2a(ii)**), although gains are generally smaller in strongly snow-influenced mountainous regions. This contrast is consistent with the snow-informed baseline already capturing much of the first-order seasonal snowmelt response, while additional coupling with a rainfall–runoff storage representation can provide further predictive value.

Relaxing the prescribed snow–soil routing into a parallel mass-conserving network increases architectural flexibility without requiring higher state dimensionality. Two-state HYDROMCP networks provide the highest predictive skill in 437 of 513 catchments, with 375 favoring sharing-enabled architectures. Importantly, this additional flexibility also changes how the internal states should be interpreted. Because both nominally snow- and soil-oriented units are fine-tuned against streamflow, their learned functions need not retain strict process identities. The resulting states are therefore better interpreted as physically constrained interacting hydrologic representations than as literal snow and soil stores.

Allowing cross-state information sharing provides additional predictive benefits without increasing state dimensionality. For fixed two-state HYDROMCP representations, the baseline configuration achieves a median testing $KGE_{ss}$ of 0.82, whereas sharing variants reach 0.83–0.84 (**Supplementary Fig. S1**). This suggests that cross-state connectivity can improve predictive performance independently of the number of represented states, raising the question of whether its contribution can instead be recovered by adding additional uncoupled states.

Despite differences in detailed architecture, the hydrologic-model and neural-network formulations converge on similar state dimensionality. The same specific architecture is selected in only 173 of 513 catchments (34%), whereas approximately 75% retain the same one- or two-state dimensionality under both formulations (**Extended Data Fig. 2a,b**). Among the 338 catchments favoring two-state hydrologic-model architectures, only 14 shift to a one-state representation under the network formulation (**Extended Data Fig. 2c**). Thus, relaxing prescribed process pathways changes how hydrologic information is represented while leaving the selected state dimensionality largely unchanged. Conceptual and neural formulations can therefore differ substantially in architecture while converging on similarly low-dimensional representations.

## Two states recover most aggregate predictive skill

To examine the influence of state dimensionality, we progressively expand nested candidate spaces from one to five allowable states. A1–A5 denote likelihood-based architecture selection, whereas KA1–KA5 denote the corresponding KGE-based selections. Under likelihood-based selection, median testing $KGE_{ss}$ increases from 0.83 with one allowable state (A1) to 0.89 with two (A2), but reaches only 0.90 with five (A5). Under KGE-based selection, median testing $KGE_{ss}$ similarly increases from 0.83 (KA1) to 0.89 (KA2) and remains approximately unchanged through KA5. Consequently, the largest reduction in the remaining performance gap ($1 - \mathrm{KGEss}$) occurs when moving from one to two states, with substantially smaller improvements thereafter (**Fig. 3a**). Thus, allowing more than two states provides little additional aggregate predictive skill under either selection objective.

The same rapid saturation appears across individual MCP formulations. For non-sharing SOIL-MCP networks, median testing $KGE_{ss}$ increases from 0.74 with one state to 0.77 with two, with little improvement thereafter. Snow-informed networks show a similar pattern, increasing from 0.78 with one state to 0.82 with two and only 0.83 with five. Multi-node mass-relaxation networks likewise show limited additional improvement beyond approximately three states (**Extended Data Fig. 3**). Thus, diminishing returns with increasing dimensionality are not specific to a single MCP architecture.

Selected architectural complexity nevertheless continues to increase even after most aggregate predictive gains have been realized. From A2 to A5, the mean number of trainable parameters increases from 35.53 to 117.34, while the corresponding increase under KGE-based selection (KA2 to KA5) is from 37.34 to 123.70 (**Fig. 3b**). Over the same range, the mean selected number of states increases from 1.97 to 4.42 under likelihood-based selection and from 2.00 to 4.47 under KGE-based selection (**Supplementary Tables S2, S5, S7 and S10**). Despite these substantial increases in model complexity, median testing $KGE_{ss}$ changes little beyond two states. Thus, increasing state dimensionality considerably expands model complexity without producing commensurate gains in aggregate predictive performance.

To assess whether aggregate predictive saturation obscures improvements in specific streamflow characteristics, we further evaluate six process-oriented hydrologic diagnostics covering flow-duration curves, flow quantiles, and high- and low-flow errors. Increasing the maximum allowable dimensionality from one to two states reduces median errors across all six diagnostics, whereas increasing it from two to five yields smaller but continued improvements in several diagnostics, particularly in log-transformed flow-duration-curve and flow-quantile errors (**Supplementary Table S13**). KGE-based selection shows the same overall pattern. Thus, although two states recover most aggregate predictive skill, allowing additional states can still refine specific aspects of the flow distribution.

## State connectivity provides information that additional uncoupled states cannot replace

To examine whether additional uncoupled states can substitute for cross-state information sharing, we compare sharing-enabled MCP candidate spaces (B/KB) with corresponding no-sharing spaces (C/KC), where K-prefixed spaces use KGE-based rather than likelihood-based architecture selection. Under likelihood-based selection, the sharing-enabled candidate space restricted to at most two states (B2) achieves a median testing $KGE_{ss}$ of 0.87, exceeding the 0.85 obtained from the no-sharing candidate space even when up to five states are allowed (C5). KGE-based selection yields the same relationship, with median $KGE_{ss}$ of 0.88 for KB2 versus 0.85 for KC5 (**Fig. 4a,b; Extended Data Tables 1 and 2; Supplementary Tables S16–S31**). This advantage extends beyond aggregate predictive skill, with B2 and KB2 outperforming C5 and KC5, respectively, in five of the six median hydrologic diagnostics (**Fig. 4c; Supplementary Table S13**). For example, FDC log-RMSE decreases from 1.057 for C5 to 0.799 for B2. Thus, increasing state dimensionality alone does not reproduce the predictive contribution of cross-state information sharing within the tested candidate spaces.

The predictive effect of connectivity also depends on the direction and extent of information sharing among states. Among snowy catchments, snow-to–rainfall–runoff state sharing is preferred more frequently than the reverse direction, whereas bidirectional sharing achieves higher median $KGE_{ss}$ than either one-way configuration (0.88 versus 0.86). However, effective connectivity does not necessarily require unrestricted all-to-all exchange. In larger SOIL-MCP networks, restricting cross-state information sharing to neighboring states produces performance comparable to full sharing (**Extended Data Fig. 4**; **Supplementary Fig. S2**). These results suggest that predictive benefits depend on how relevant information is exchanged among states rather than simply on the density of network connections.

The predictive benefit of cross-state information sharing is also observed across different MCP formulations. Fully connected SOIL-MCP networks achieve higher predictive skill, with particularly pronounced improvements in snow-affected mountainous catchments. Snow-informed networks provide a more direct comparison between connectivity and dimensionality: a two-state SNOWMCP with sharing achieves a median testing $KGE_{ss}$ of 0.84, slightly exceeding the 0.83 obtained with five uncoupled states. Mass-relaxation networks likewise benefit from information sharing. Together, these results indicate that the contribution of connectivity is not specific to a single MCP formulation.

## Hydrologic representation requirements vary across hydroclimatic regimes

Hydrologic representation requirements vary systematically across hydroclimatic settings. The spatial distributions of positive testing-$KGE_{ss}$ improvements relative to HMCP and SWE-pretrained SNOWMCP reveal distinct baseline-dependent patterns (**Fig. 2a(i,ii)**). Improvements over HMCP are particularly pronounced in snow-affected mountainous regions, whereas gains over SNOWMCP are generally smaller in these regions, indicating that the pretrained snow-informed representation already captures substantial predictive information. The corresponding elevation-stratified comparisons show that median testing $KGE_{ss}$ decreases from 0.67 to 0.31 for HMCP between low- and high-elevation catchments, whereas SNOWMCP increases from 0.67 to 0.86 and HYDROMCP from 0.65 to 0.89 (**Fig. 2b**). A similar contrast emerges across snowpack regimes: from catchments with maximum annual SWE below 100 mm to those above 600 mm, median testing $KGE_{ss}$ changes from 0.68 to 0.43 for HMCP, from 0.65 to 0.84 for SNOWMCP, and from 0.63 to 0.87 for HYDROMCP (**Fig. 2c**). Together, these results indicate that the predictive value of snow-informed representations increases with snow influence, while the incremental benefit of coupling an additional rainfall–runoff component depends on the baseline representation.

Additional state structure can provide further predictive gains under strongly snow-influenced conditions. For catchments with maximum annual SWE of 400–600 mm, one- to two-state HYDROMCP architectures achieve a median testing $KGE_{ss}$ of 0.90, compared with 0.93 for a three-state sharing-enabled SNOWMCP network (**Supplementary Table S32**). This suggests that catchments with substantial seasonal snow storage may benefit from additional representational flexibility. Thus, the low-dimensional saturation observed across CONUS does not imply a universal state requirement; more demanding hydroclimatic regimes can benefit from somewhat richer representations.

Hydroclimatic dependence is also evident along the aridity gradient (**Fig. 2d**). In relatively humid catchments (PET/P < 0.5), SNOWMCP achieves the highest median testing $KGE_{ss}$ (0.82), compared with approximately 0.77 for HMCP and HYDROMCP. In contrast, under more water-limited conditions (PET/P = 1.5–2), HYDROMCP achieves a median $KGE_{ss}$ of 0.68, compared with 0.49 and 0.51 for HMCP and SNOWMCP, respectively. Even in the driest catchments (PET/P > 2), where predictive skill declines across all three models, HYDROMCP retains a higher median $KGE_{ss}$ (0.54 versus 0.40 and 0.37 for HMCP and SNOWMCP, respectively). Together, these results demonstrate that the relative predictive benefits of different hydrologic representations vary across hydroclimatic regimes, rather than following a universal hierarchy determined by model complexity.

## Different selection objectives identify different architectures but preserve the same structural relationships

Architecture-selection objectives influence model-family preferences, although aggregate out-of-sample performance remains similar. Within the full five-state candidate spaces, likelihood-based selection (A5) favors LSTM networks in 312 of 513 catchments, compared with 201 MCP-based architectures. Under KGE-based selection (KA5), the corresponding frequencies are 350 and 163 catchments, respectively (**Fig. 5a**). The selected architectures have similar mean state dimensionalities (4.42 versus 4.47) and parameter counts (117.34 versus 123.70), with median testing $KGE_{ss}$ values of 0.90 and 0.89, respectively (**Fig. 5b,c; Extended Data Table 3**). Thus, changing the selection objective alters model-family preferences without substantially affecting aggregate predictive performance or selected architectural complexity.

The broader predictive relationships are more stable than the identities of the selected architectures. Low-dimensional saturation is also reproduced within the LSTM family: median testing $KGE_{ss}$ increases from 0.81 with one state to approximately 0.88 with two and changes little thereafter under either selection framework, despite continued increases in dimensionality and parameter count (**Extended Data Fig. 5**; **Supplementary Tables S33–S42**). Rapid saturation is therefore not specific to mass-conserving architectures. In a separate comparison of the best MCP-based and LSTM representations, both achieve a median testing $KGE_{ss}$ of 0.90, although MCP networks use substantially fewer trainable parameters on average (69.45 versus 115.86; **Supplementary Tables S43–S45**). These results indicate that similarly high predictive performance can be achieved through different architectural representations with substantially different parameter requirements.

Most importantly, the principal relationships among state dimensionality, connectivity and predictive performance remain consistent across selection objectives. Across the full candidate spaces (A/KA), sharing-enabled MCP spaces (B/KB), no-sharing MCP spaces (C/KC) and LSTM-only spaces (D/KD), both likelihood- and KGE-based selection recover the dominant predictive gains from one to two states, followed by diminishing returns at higher dimensionality (**Fig. 3**; **Supplementary Tables S1–S10 and S33–S42**). The predictive advantage of cross-state connectivity is likewise preserved: B2 outperforms C5, and KB2 outperforms KC5, in five of the six median hydrologic diagnostics (**Fig. 4**; **Supplementary Table S13**). Thus, although the selection objective influences which architectures are preferred, it does not materially change the inferred structural relationships governing effective hydrologic complexity.

## Discussion

Our results indicate that catchment-scale rainfall–runoff dynamics can often be represented within a compact state space, but that effective hydrologic complexity cannot be characterized by state dimensionality alone. Across 513 CAMELS-US catchments, most of the gains in aggregate $KGE_{ss}$ occur when moving from one to two states, whereas additional states provide progressively smaller overall gains while continuing to refine selected aspects of the flow distribution. The recurrence of this pattern across both MCP and LSTM architectures suggests that rapid low-dimensional saturation is not simply imposed by mass conservation or by a particular MCP parameterization. Instead, different model families can achieve comparable predictive performance using relatively few internal states.

Importantly, however, state number alone does not determine the information available to a model. Sharing-enabled MCP candidate spaces restricted to two states outperform no-sharing spaces allowing up to five states in median testing $KGE_{ss}$ under both likelihood- and KGE-based selection. This advantage extends to five of the six median hydrologic diagnostics, while the 5th-percentile flow error is indistinguishable at the reported precision (**Supplementary Table S13**). Moreover, the comparable performance of partially and fully shared SOIL-MCP networks suggests that useful cross-state interactions do not necessarily require unrestricted information exchange among all states. Effective complexity therefore reflects not only how many states are represented, but also how information is exchanged among them. In this sense, connectivity constitutes a distinct dimension of representational capacity that cannot be characterized by state dimensionality or parameter count alone.

The effective representation is also strongly regime dependent. Snow-informed and state-sharing architectures provide their clearest benefits in high-elevation and strongly snow-influenced catchments, where seasonal accumulation and delayed melt introduce memory and interactions that are poorly captured by simple non-snow-informed states. Although two-state representations recover most aggregate predictive skill across CONUS, strongly snow-influenced catchments can retain measurable gains from somewhat richer representations. Notably, two-state HYDROMCP networks are also preferred in 116 of 153 non-snowy catchments, suggesting that the benefits of multiple interacting states cannot be attributed exclusively to explicit snow processes. Such flexibility may instead help to represent multiple runoff pathways or other forms of hydrologic memory and interaction. These patterns argue against a single universally appropriate level of state complexity. Rather, the useful form of complexity depends on the hydrologic functions that must be represented within each catchment. Differences in predictive skill across broad land-cover classes further suggest that identifying these requirements will require attributes that more directly characterize hydrologic function than coarse categorical descriptors alone.

At the same time, our results suggest that aggregate predictive performance and the principal relationships among dimensionality and connectivity are more stable than the specific architectures selected to achieve them. Likelihood- and KGE-based selection alter model-family preferences across catchments, yet produce similar mean selected state dimensionalities, parameter counts and aggregate testing $KGE_{ss}$ within the full candidate spaces. This behavior is consistent with structural equifinality in streamflow prediction: multiple internal representations can reproduce similar observable hydrologic responses. Consequently, selecting a more complex architecture should not by itself be interpreted as evidence that the catchment contains a correspondingly larger number of independently identifiable physical stores or pathways. The continued increase in selected dimensionality after most aggregate predictive gains have been realized is particularly important in this regard. Additional states may provide useful functional refinement without being uniquely identifiable as distinct physical components of the catchment. Predictive adequacy and physical identifiability are therefore related but separate questions.

This distinction between functional usefulness and physical identification is central to interpreting MCP states. Mass conservation at the computational-unit level and explicit storage–flux relationships make MCP representations structurally- and conceptually-interpretable in ways that generic recurrent-network states are not specifically designed to be. However, these constraints do not establish that each learned state corresponds uniquely to an observable snowpack, soil-water store, groundwater reservoir or other physical quantity. Streamflow-only fine-tuning can alter the functional roles of nominally snow- or soil-oriented units, while cross-state information sharing allows the output and loss gates of individual units to respond to information stored in other states. Importantly, such information exchange should not be confused with a physical transfer of water between storage components. Although these interactions increase representational flexibility, they do not independently establish the physical identities of the learned states. Stronger physical attribution will therefore require additional observational constraints, potentially including SWE, soil moisture, groundwater storage, evapotranspiration and other internal states or fluxes. Joint training and evaluation against multiple hydrologic responses would provide a means of testing predictive adequacy and physical state correspondence separately rather than inferring the latter from streamflow performance alone.

More broadly, these results support viewing conceptual hydrologic models and neural networks not as categorically separate modelling paradigms, but as different points along a continuum of structural constraint. Prescribed conceptual architectures encode strong hypotheses about states and physical process pathways, whereas neural-network formulations relax some of those hypotheses and permit greater functional adaptation. The MCP framework makes this continuum explicit because the same locally mass-conserving computational units can be organized as traditional hydrologic structures or as more flexible networks. The hydrologic-model and network formulations select the same specific architecture in only 173 of 513 catchments, yet retain the same one- or two-state dimensionality in 386 catchments. This contrast demonstrates that relaxing structural constraints can substantially alter the preferred architecture without proportionally changing the effective state dimensionality or aggregate predictive performance. The resulting perspective suggests a route toward hydrologic modelling frameworks in which physical principles constrain admissible representations, while data help determine the dimensionality, connectivity and functional flexibility needed at individual catchments. Rather than seeking either a universally fixed conceptual structure or an unconstrained increase in network complexity, the central challenge becomes identifying the simplest interacting representation that retains the hydrologic information needed for the prediction and scientific question of interest.

## Methods

### Mass-conserving computational units

The Mass-Conserving Perceptron (MCP) is a recurrent computational unit that represents system memory through an internal storage state while enforcing nodal mass conservation. Context-dependent input, output, loss and retention fluxes are represented through learnable gating functions, combining reservoir-like physical constraints with data-driven flexibility. All MCP-based architectures are implemented using standard differentiable machine-learning operations; detailed state-update equations, gate parameterizations and trainable parameters are provided in **Supplementary Notes 1–3**. We consider three MCP unit types. HMCP represents a soil-moisture-like catchment storage that receives precipitation and partitions stored water among streamflow, evapotranspirative loss and retained storage. Its output gate depends on internal storage, whereas its loss gate depends on potential evapotranspiration and storage state. HMCP provides the basic rainfall–runoff unit used throughout the MCP network experiments.

SNOWMCP represents snow accumulation and melt through a SWE-like storage state. Precipitation is partitioned into rainfall and snowfall according to air temperature; snowfall enters the storage, whereas rainfall bypasses it. Stored snow is partitioned among snowmelt, sublimation–evapotranspiration loss and retained storage through non-negative gates constrained to sum to one. We use the simplest SNOWMCP configuration, in which rain–snow partitioning depends on mean air temperature and the melt and loss gates depend on temperature and internal SWE. The third unit augments HMCP with a mass-relaxation mechanism that permits bidirectional exchange with an unresolved surrounding environment. Exchange depends on the internal state relative to a learned reference level, while the positive mass-relaxation gate is limited by the fraction of storage remaining after the output and loss fluxes. This formulation provides an alternative mechanism for representing unresolved hydrologic exchange without explicitly adding additional storage states (**Supplementary Note 3**).

## Coupled snow–soil hydrologic representations

We construct two two-state HYDROMCP hydrologic models by coupling SNOWMCP with HMCP. In the routing formulation (**Fig. 1b**, top left), rainfall and SNOWMCP-generated snowmelt enter the downstream HMCP storage, and streamflow is generated from the HMCP output. In the bypass formulation (**Fig. 1b**, bottom left), rainfall enters HMCP, whereas snowmelt bypasses the soil-moisture-like storage and contributes directly to streamflow. These alternatives represent different hypotheses about whether snowmelt is routed through catchment storage before reaching the outlet. To relax these prescribed routing hypotheses, we additionally formulate a two-state HYDROMCP neural network (**Fig. 1b**, right), in which HMCP and SNOWMCP operate as parallel mass-conserving units and their streamflow contributions are combined through a learned Softmax output layer. This retains interpretable internal storage states while replacing prescribed snow-to-soil routing with a more flexible learned representation.

## State connectivity and distributed-state networks

Cross-state connectivity is introduced by allowing the gating functions of one MCP unit to access the internal states of other units. In the two-state HYDROMCP representations, we evaluate one-way sharing from SNOWMCP to HMCP, one-way sharing from HMCP to SNOWMCP and bidirectional sharing. We further construct one- to five-state distributed MCP networks (**Fig. 1d,e**) using homogeneous HMCP, SNOWMCP or mass-relaxation units. Each node receives the same forcing, and node outputs are combined through a Softmax layer. Three connectivity structures are evaluated: no sharing, full sharing and partial sharing, in which each node accesses only neighboring states (**Fig. 1c**). Additional SNOWMCP output-partitioning and mass-relaxation variants, together with the complete set of MCP-based model and network configurations evaluated in this study, are summarized in **Supplementary Table S46**.

For comparison with a flexible non-mass-conserving architecture, we train **single-layer, basin-wise LSTM networks** separately for each catchment using the same data allocation and evaluation framework. LSTM architectures with one to five hidden-state dimensions are evaluated. For each state dimensionality, we test batch sizes of 256 and 512 and sequence lengths of 270, 300, 330, 360 and 390 days, with each hyperparameter combination trained from 10 random initializations. **The best-performing realization across all hyperparameter combinations and random initializations is retained based on the highest $KGE$ on the independent architecture-** **selection subset.** The selected LSTM is then evaluated on the same held-out testing subset as the MCP-based models. The LSTM implementation follows the basin-wise benchmark framework described in our previous study[15], with code adapted from Kratzert et al.[11].

Together, these architectures provide controlled variations in three dimensions of representational complexity: state dimensionality, defined by the number of internal storage or recurrent states; state connectivity, defined by the information exchanged among states; and parameterization, quantified by the number of trainable parameters. The complete architecture set is summarized in **Supplementary Table S46**, while the scientific questions associated with each model and network class are summarized in **Supplementary Table S47**.

## Data and hydroclimatic classification

We use 513 catchments from the CAMELS-US dataset with continuous streamflow records over water years 1982–2008. Meteorological inputs include precipitation and potential evapotranspiration (PET). Precipitation is obtained from the Maurer forcing dataset, while potential evapotranspiration (PET) is estimated using the Priestley–Taylor formulation with a default coefficient of 1.26, following the implementation provided by Gnann[36]. Catchment-averaged daily SWE is obtained from the University of Arizona 4-km SWE product. Catchments are characterized by geographic region, elevation, aridity, snowpack magnitude, forest cover and climate classification (**Extended Data Fig. 1**). A basin is classified as snowy when its CAMELS-US snowfall fraction exceeds 5% and median annual SWE exceeds 3 mm. April 1 SWE, snow-season duration and the timing of annual maximum SWE are used to further characterize snow regimes (**Supplementary Fig. S3**).

## Training and independent data allocation

Models are trained separately for each basin. Daily streamflow and SWE observations are independently partitioned into training, architecture-selection and testing subsets using a distribution-preserving data-allocation procedure[37]. This procedure maintains similar response distributions among subsets while reserving the testing data exclusively for final out-of-sample evaluation. For streamflow, the 9,862 daily observations over WY1982–2008 are divided into 4,930 training timesteps and 2,466 timesteps each for architecture selection and testing. SWE is partitioned independently using the same procedure, such that SWE and streamflow subsets do not necessarily contain identical dates.

MCP-based models are optimized using the gradient-based Adam algorithm[38], with a three-year spin-up period used to initialize recurrent states. HYDROMCP models and networks are initialized from previously trained HMCP and SNOWMCP components and optimized using stagewise training. We compare progressive initialization, in which newly introduced sharing architectures inherit parameters from simpler trained configurations, with baseline initialization, in which sharing variants inherit common parameters directly from the baseline two-state network, except for the reinitialized Softmax output weights. For each basin and HYDROMCP sharing architecture, both initialization strategies are evaluated across the same ten random seeds using identical optimization settings. Among the resulting model realizations, the one achieving the highest KGE on the independent architecture-selection subset is retained for subsequent evaluation. The spatial distribution of the preferred initialization strategy across HYDROMCP architectures is shown in **Supplementary Fig. S4**. Detailed parameter-initialization and learning-rate procedures are provided in **Supplementary Note 4**.

## Performance evaluation

All models are trained by maximizing Kling–Gupta efficiency[39] ($KGE$) on the training subset. $KGE$ is also used as one of the two architecture-selection objectives described below. Final out-of-sample predictive performance is reported using the Kling–Gupta efficiency skill score[40] ($KGE_{ss}$), defined as $KGEss = 1 - (1 - KGE)/\sqrt{2}$. This transformation uses $KGE = 1 - \sqrt{2}$ as the reference benchmark, such that $KGE_{ss}$ = 1 represents perfect agreement and $KGE_{ss}$ = 0 corresponds to the benchmark performance. We additionally report the correlation, variability and bias components of $KGE$ (**Supplementary Tables S48–S50**). To diagnose aspects of streamflow behavior that may be obscured by an aggregate performance metric, we evaluate six process-oriented hydrologic diagnostics following the diagnostic modelling perspective[41]: normalized flow-duration-curve (FDC) RMSE, log-transformed FDC RMSE, normalized absolute errors in the 95th- and 5th-percentile flows, and normalized RMSE over the highest and lowest 10% of observed flows. All six are error metrics for which lower values indicate better performance.

Basin-wise distributions of the six diagnostic errors are summarized using performance-oriented percentiles, for which higher percentile ranks always denote better-performing catchments. Because lower values indicate better performance for these errors, the 95th performance percentile corresponds to the numerical 5th percentile, the 75th to the numerical 25th percentile, the median to the numerical median, the 25th to the numerical 75th percentile and the 5th to the numerical 95th percentile (**Supplementary Tables S11–S15**).

## Architecture selection and complexity analysis

We use two complementary criteria for basin-specific architecture selection, both applied exclusively to the independent architecture-selection subset; the testing subset is never used for architecture selection. For likelihood-based selection, residuals from each candidate architecture are evaluated under six alternative probability distributions: Normal, Gumbel, Gamma, Lognormal, lower-bounded Lognormal and Pearson Type III. For each catchment and residual distribution, distribution parameters are estimated by maximum likelihood, and candidate architectures are compared using their maximized selection-period log-likelihoods. **Architectures are compared only within the same residual distribution, and the architecture with the highest log-likelihood is selected independently for each of the six distributions. The final architecture is determined by the modal selection across the six distributions; when two or more architectures are selected equally often, the simpler architecture is retained.** Log-likelihood values from different residual distributions are therefore not compared directly, and no explicit parameter-count penalty is included in this selection criterion.

For KGE-based selection, the architecture with the highest $KGE$ on the independent architecture-selection subset is retained. Model complexity is evaluated separately through the selected number of states and trainable parameters and is not incorporated explicitly into either selection objective. We define four nested candidate spaces. **A/KA** contains all MCP-based, HYDROMCP and LSTM candidate architectures; **B/KB** excludes LSTM architectures but retains MCP networks with cross-state sharing; **C/KC** additionally excludes cross-state sharing; and **D/KD** contains only LSTM architectures. The first letter denotes likelihood-based selection, whereas the corresponding K-prefixed space denotes KGE-based selection. A numerical suffix from 1 to 5 specifies the maximum allowable state dimensionality; for example, **A2** and **KA2** contain candidates with at most two states under likelihood- and KGE-based selection, respectively.

Architecture selection is repeated over nested candidate spaces allowing maximum state dimensionalities from one to five. This design allows testing performance, hydrologic diagnostics, selected state dimensionality and parameter count to be evaluated as representational capacity increases, while the comparison between sharing-enabled and no-sharing candidate spaces isolates the contribution of cross-state connectivity. Partial-sharing networks are analyzed separately because their restricted topology is not directly comparable with the fully connected sharing-enabled candidate spaces.

## Reporting summary

Further information on research design is available in the Nature Portfolio Reporting Summary linked to this article.

## Data availability

The datasets used in this study include daily meteorological forcing, observed streamflow and snow water equivalent (SWE) data for 513 CAMELS-US basins. The input datasets, basin-specific data partitioning information, normalization statistics and supporting data required to reproduce the main figures and analyses are available via Zenodo[42]. A README accompanying the repository documents the data organization and provides instructions for reproducibility.

## Code availability

The source code and analysis scripts used in this study are publicly available via Zenodo[42]. The repository includes implementations of HMCP, SNOWMCP, HYDROMCP and LSTM benchmark models, together with scripts for model training and evaluation. The paper_figures/ directory contains the files required to reproduce all main figures, while paper_tables/ provides MATLAB scripts for calculating six streamflow signatures, performing model selection using maximum-likelihood and Kling–Gupta efficiency (KGE) procedures, and conducting regionalization-based post-processing of the model-selection results. The accompanying README describes the repository structure, software dependencies and instructions for reproducing the analyses.

## Acknowledgements

The authors acknowledge the high-performance computing resources provided by the University of Arizona.

## Author contributions

Y.-H.W. conceived and designed the study, developed the models, conducted the computational experiments, performed the analyses and visualizations, and wrote the original manuscript. H.V.G. contributed to the study conceptualization, interpretation of results, and review and editing of the manuscript. Both authors reviewed and approved the final manuscript.

## Funding

This research received no specific funding.

## Competing interests

The authors declare no competing interests.

## Additional information

**Correspondence and requests for materials** should be addressed to Yuan-Heng Wang.

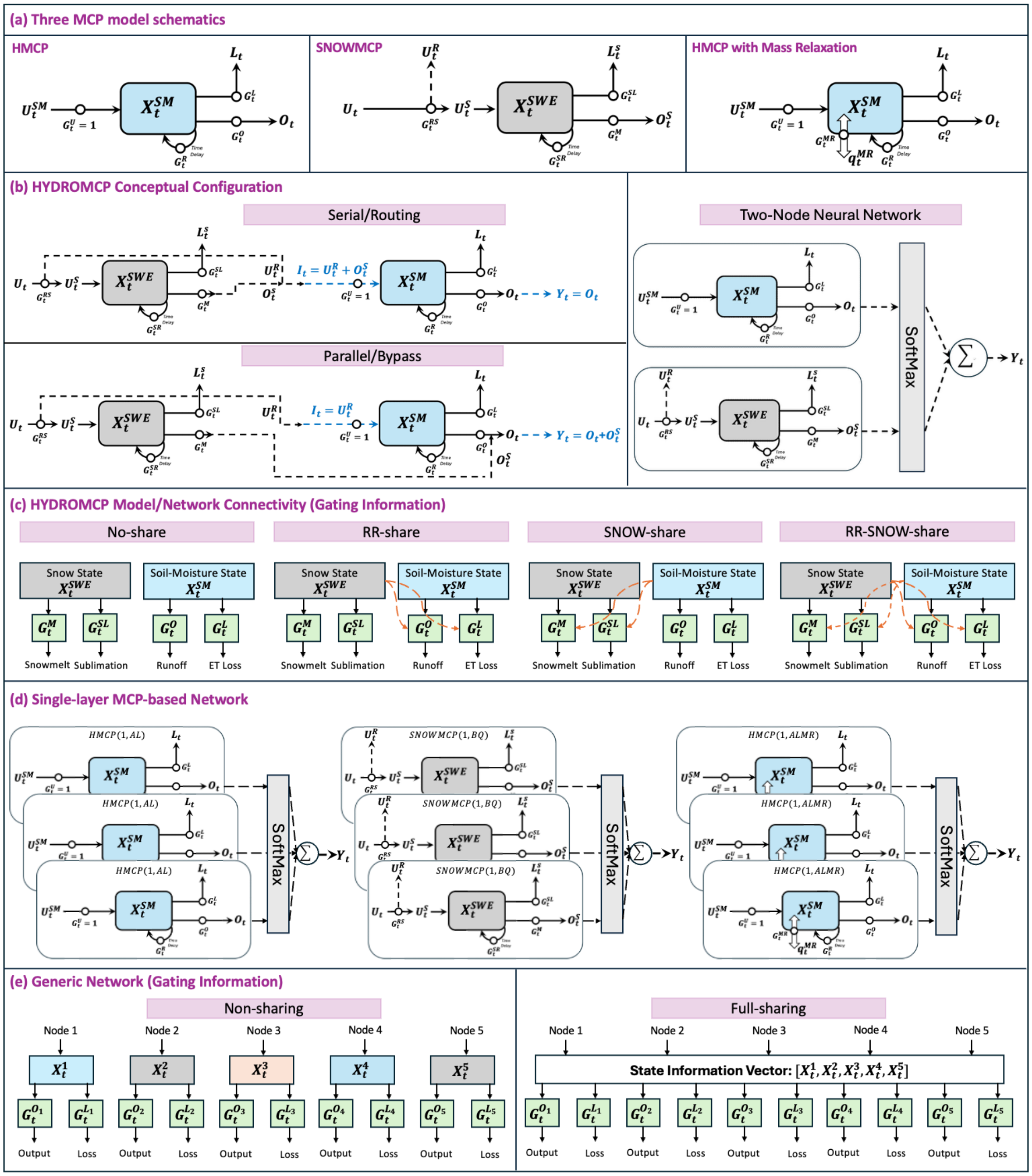

**Fig. 1 | From mass-conserving perceptron units to conceptual hydrologic models and information-sharing networks. a,** Three mass-conserving perceptron (MCP) computational units: the rainfall–runoff HMCP, the snow-informed SNOWMCP and HMCP with mass relaxation. In HMCP, $X_t^{SM}$ denotes a soil-moisture-like storage state, $U_t^{SM}$ its input, $O_t$ runoff and $L_t$ losses; $G_t^U = 1$, $G_t^Q$, $G_t^L$ and $G_t^R$ denote the input, output, loss and retention (remember) gates, respectively. In SNOWMCP, $G_t^{RS}$ partitions precipitation $U_t$ into snowfall $U_t^S$ and rainfall $U_t^R$; $X_t^{SWE}$ is snow-water-equivalent storage, $G_t^M$ and $G_t^{SL}$ control snowmelt output $O_t^S$ and snow-related loss $L_t^S$, and $G_t^{SR}$ is the snow-retention gate. The mass-relaxation HMCP additionally permits signed exchange $q_t^{MR}$ with unresolved system components, regulated by $G_t^{MR}$. **b,** Two prescribed HYDROMCP configurations and a parallel two-node neural network. In serial/routing, rainfall and snowmelt enter the rainfall–runoff state ($I_t = U_t^R + O_t^S$) and streamflow is $Y_t = O_t$; in parallel/bypass, only rainfall enters that state ($I_t = U_t^R$), while snowmelt contributes directly to streamflow ($Y_t = O_t + O_t^S$). The two-node network combines separate HMCP and SNOWMCP outputs through Softmax-based aggregation. **c,** Four configurations of cross-state gating information: no-share, RR-share, SNOW-share and bidirectional RR–SNOW-share. RR-share supplies $X_t^{SWE}$ to both the output and loss gates of the rainfall–runoff node; SNOW-share supplies $X_t^{SM}$ to both the melt and loss gates of the snow node. Orange dashed arrows indicate additional information supplied to gates, not physical mass transfer. **d,** Three examples of single-layer, three-node MCP-based networks assembled from HMCP(1,AL), SNOWMCP(1,BQ) or mass-relaxation HMCP(1,ALMR) units; their node outputs are combined through Softmax-based aggregation. **e,** Generic five-node non-sharing and fully sharing networks. In non-sharing networks, each state $X_t^i$ informs its own output and loss gates, $G_t^{O,i}$ and $G_t^{L,i}$; under full sharing, the state vector $[X_t^1, X_t^2, X_t^3, X_t^4, X_t^5]$ informs both gates of every node. Subscript $t$ denotes the time step.

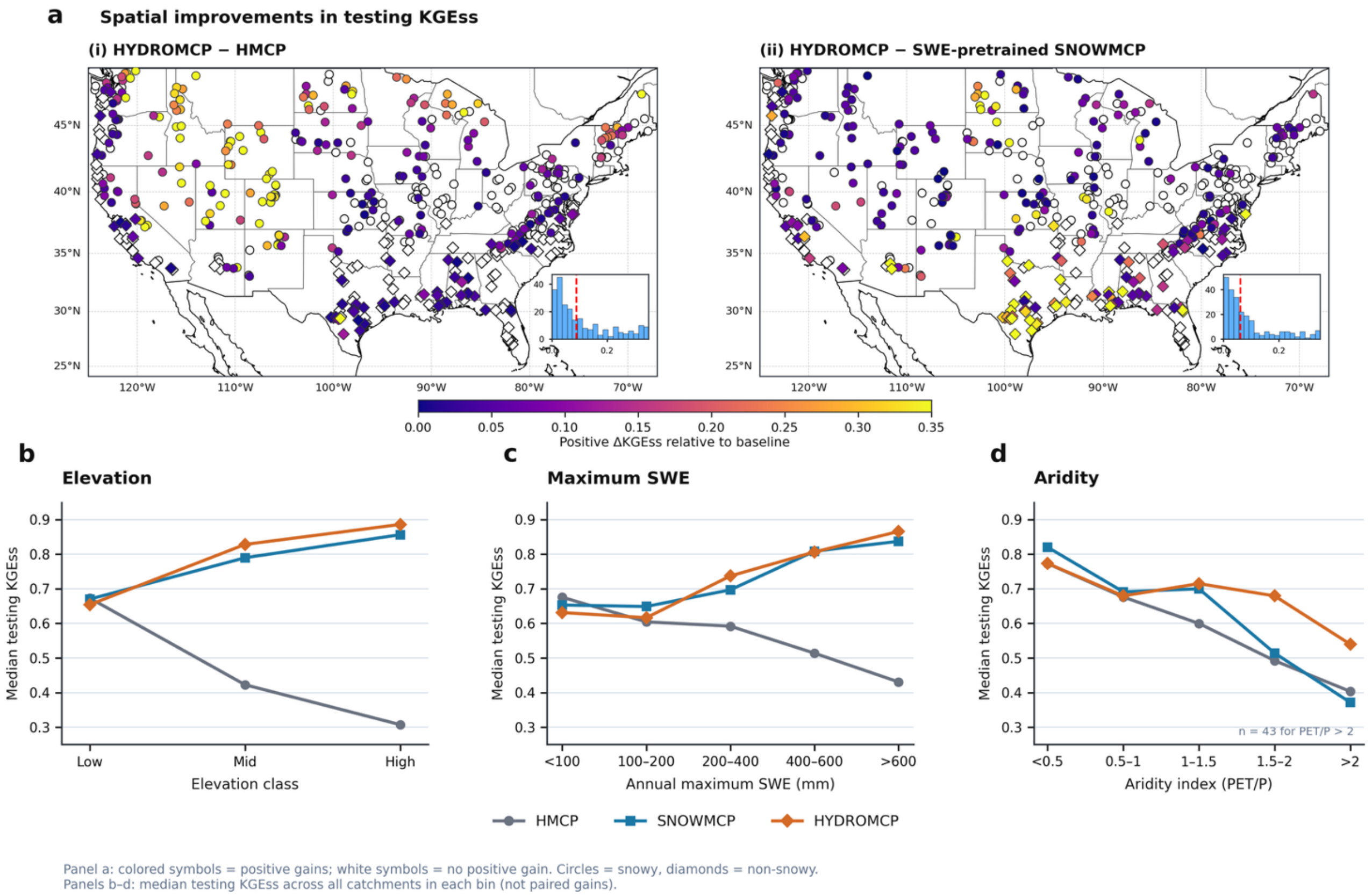

**Fig. 2 | Hydroclimatic dependence of hydrologic representation performance. a,** Spatial distributions of changes in testing-period $KGE_{ss}$ achieved by the basin-wise better-performing two-state HYDROMCP routing or bypass configuration relative to (i) single-state HMCP and (ii) SWE-pretrained single-state SNOWMCP across 513 CAMELS-US catchments. Colored circles (snowy catchments) and diamonds (non-snowy catchments) indicate positive improvements; unfilled symbols indicate no positive improvement. The two maps use a common color scale. Inset histograms show the distributions of positive improvements, with red dashed lines marking their medians. **b–d,** Median testing-period $KGE_{ss}$ for HMCP, SNOWMCP and HYDROMCP grouped by elevation (**b**), maximum annual snow water equivalent (SWE; **c**) and aridity index (PET/P, the ratio of potential evapotranspiration to precipitation; **d**). Unlike panel **a**, panels **b–d** include all catchments in each hydroclimatic category and present absolute performance rather than paired improvements.

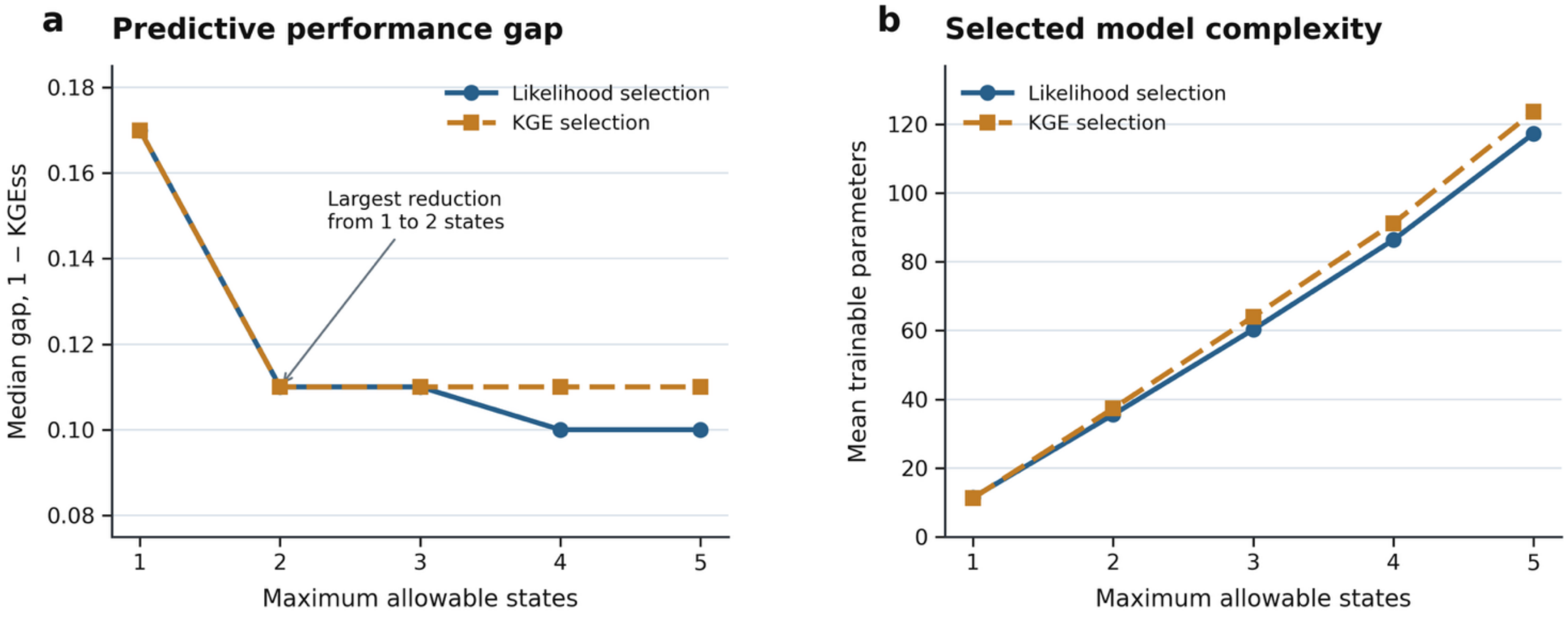

**Fig. 3 | Predictive skill saturates while selected model complexity grows. a,** Median testing-period predictive-performance gap, defined as one minus the cross-catchment median testing $KGE_{ss}$, as the maximum allowable number of states increases from one to five across nested candidate spaces. Blue and orange lines represent likelihood-based (A1–A5) and KGE-based (KA1–KA5) architecture selection, respectively. The largest reduction in the performance gap occurs between one and two allowable states. **b,** Mean number of trainable parameters in the architectures selected from the same candidate spaces. Results are aggregated across 513 CAMELS-US catchments. $KGE_{ss}$ denotes the Kling–Gupta efficiency skill score; smaller gaps indicate better predictive performance.

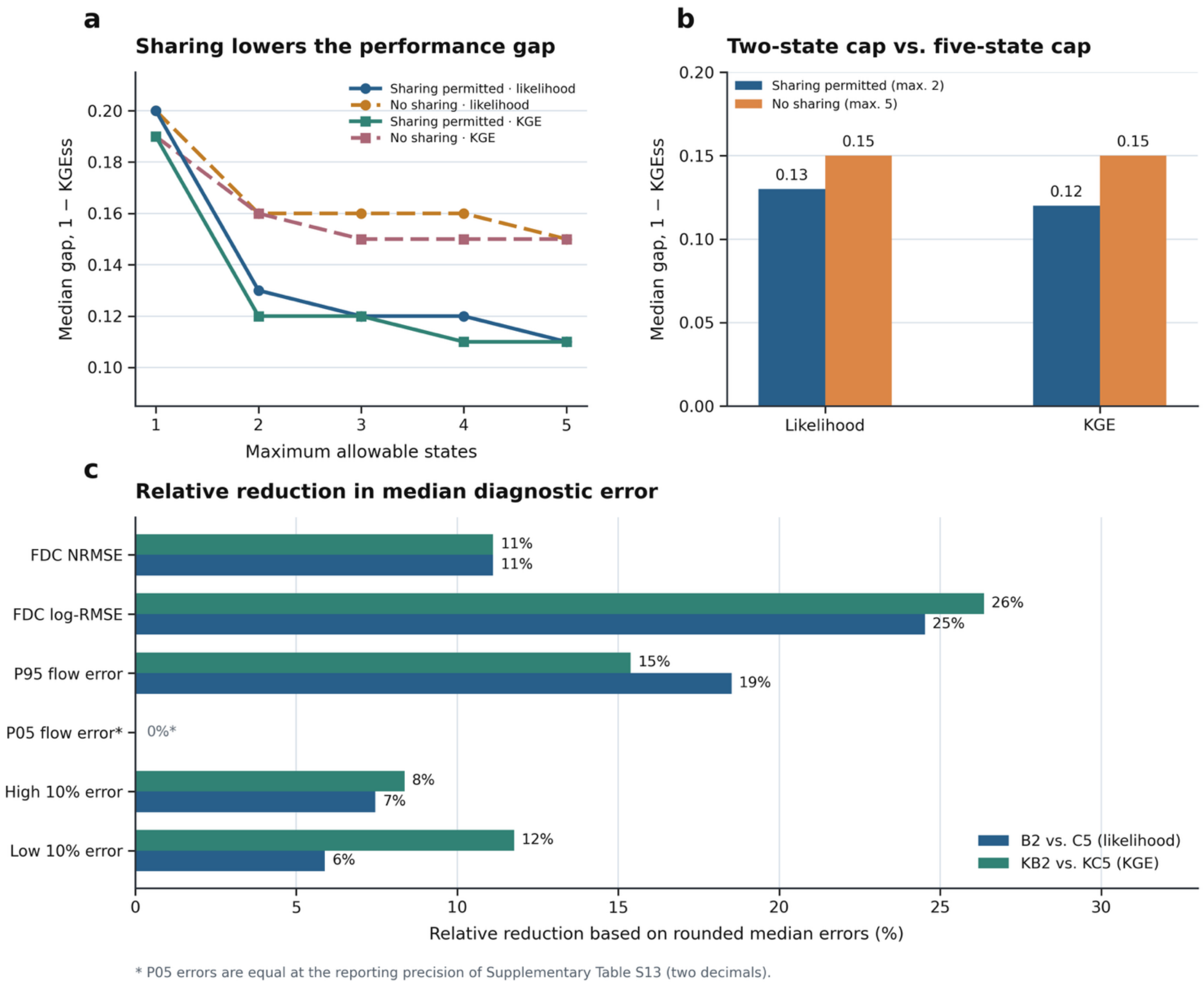

**Fig. 4 | Cross-state interaction provides a distinct predictive benefit. a,** Median testing-period performance gap (one minus median testing $KGE_{ss}$), across 513 catchments for MCP candidate spaces permitting cross-state information sharing (B/KB) or prohibiting sharing (C/KC), with one to five maximum allowable states. Likelihood-based and KGE-based selection results are shown separately. **b,** Direct comparison of sharing-permitted candidate spaces capped at two states (B2 and KB2) with no-sharing spaces permitting up to five states (C5 and KC5), under the two selection objectives. **c,** Relative reductions in cross-catchment median diagnostic errors for B2 versus C5 and KB2 versus KC5, expressed as a percentage of the corresponding no-sharing error. Diagnostics include normalized flow-duration-curve (FDC) $RMSE$, log-FDC $RMSE$, normalized absolute errors in the 95th- and 5th-percentile flows (P95 and P05), and normalized $RMSE$ for the highest and lowest 10% of flows. Percentage reductions are calculated from medians rounded to two decimal places (Supplementary Table S13); P05 errors are equal at this reporting precision. Lower performance gaps and diagnostic errors indicate better predictions.

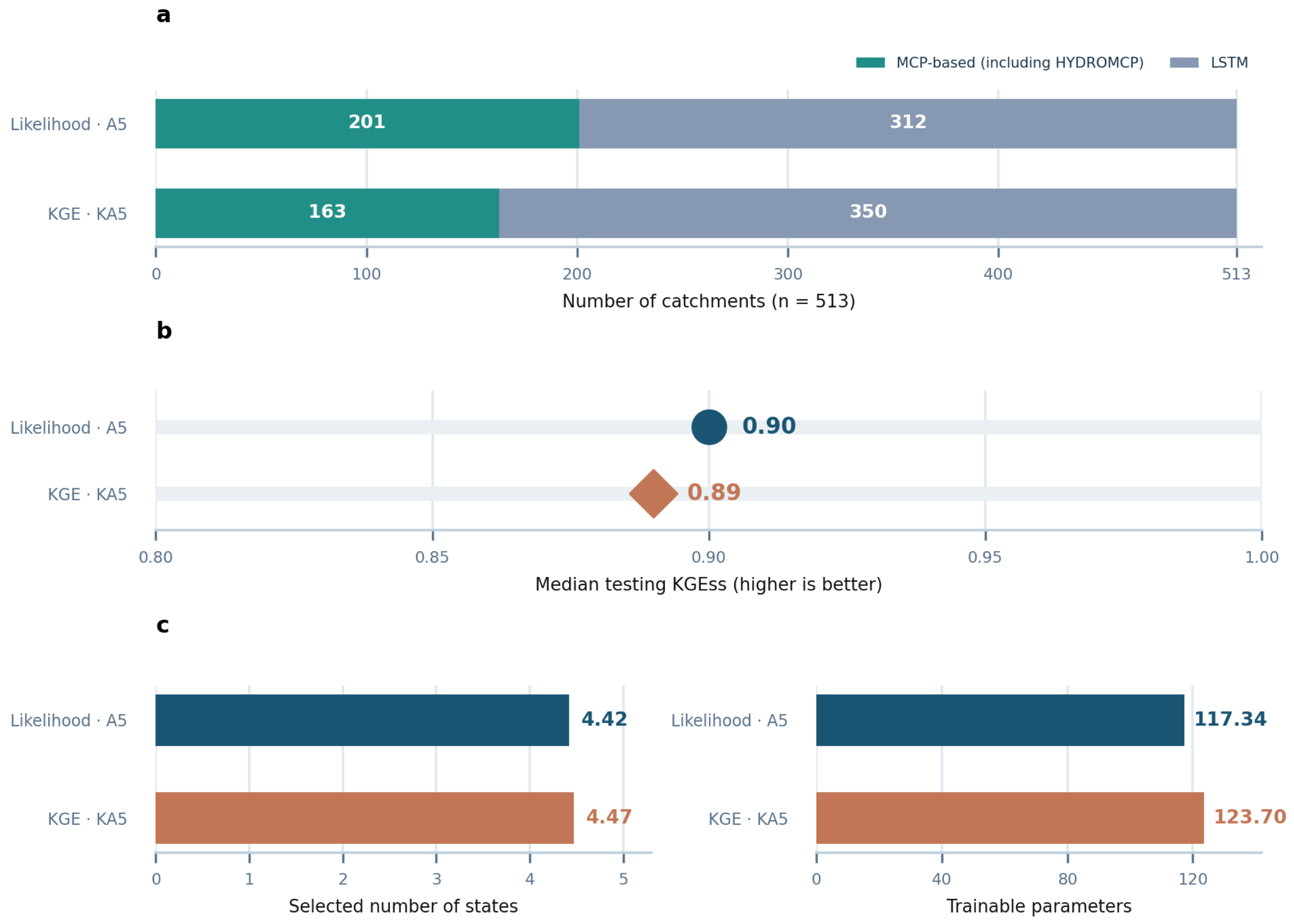

**Fig. 5 | Selection objectives favor different architectures, with similar testing skill. a,** Numbers of catchments selecting an MCP-based architecture (including HYDROMCP) or a long short-term memory (LSTM) network from the full candidate space permitting up to five states, under likelihood-based (A5) and KGE-based (KA5) architecture selection. **b,** Median testing-period $KGE_{ss}$ of the basin-wise architectures selected using the respective objectives. **c,** Mean state dimensionality (left) and mean number of trainable parameters (right) for the corresponding selected architectures. Results cover 513 CAMELS-US catchments; A5 and KA5 are defined by the selection objective, not by a requirement that every selected architecture contain five states (Supplementary Tables S5 and S10).

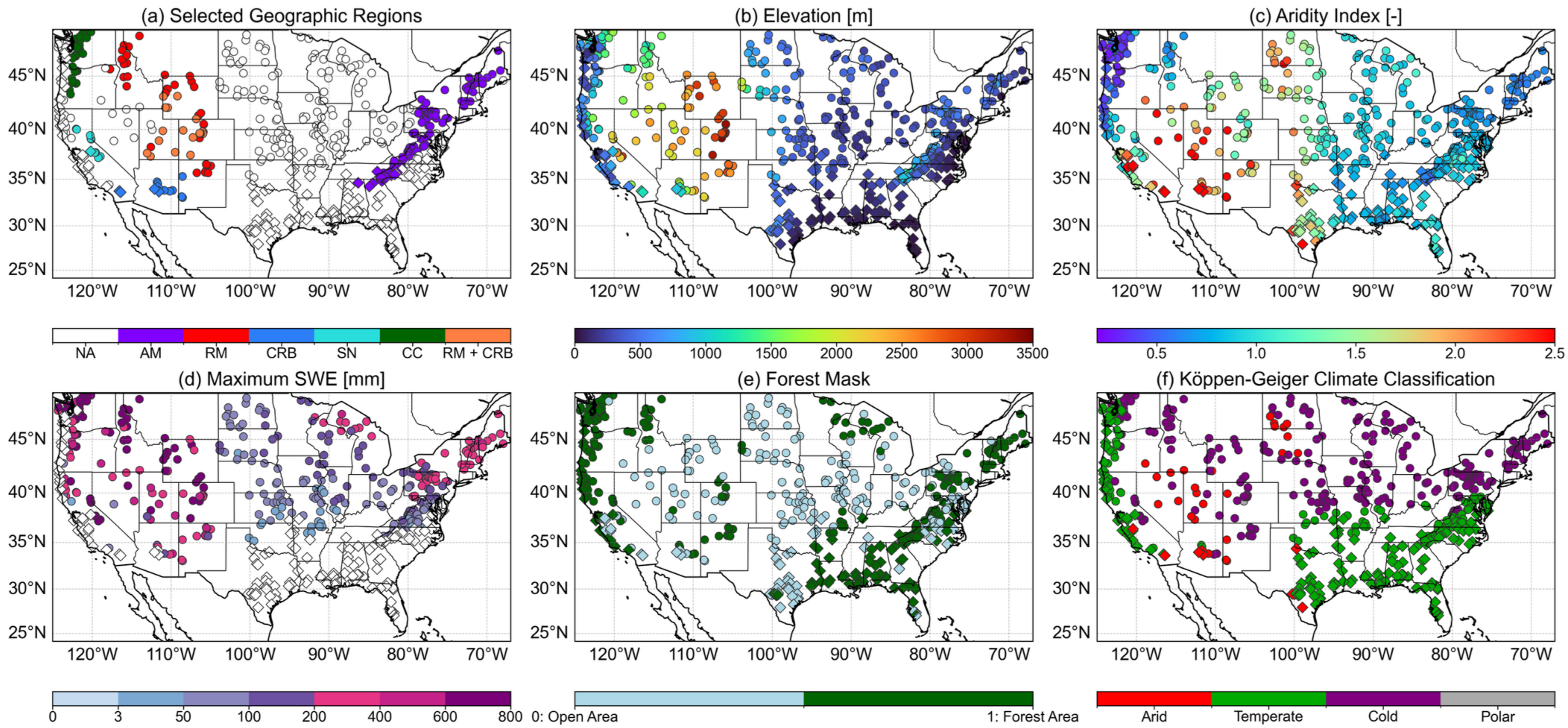

**Extended Data Fig. 1 | Hydroclimatic and physiographic characteristics of the study catchments.**

**a–f,** Spatial characteristics of the 513 CAMELS-US catchments used in this study, showing selected geographic regions (**a**), elevation (**b**), aridity index (**c**), maximum snow water equivalent (SWE; **d**), forest versus open classification (**e**) and Köppen–Geiger climate classification (**f**). Elevation and aridity are obtained from CAMELS-US, and SWE is derived from the University of Arizona 4-km SWE product over water years 1982–2008. Circles denote snowy catchments and diamonds denote non-snowy catchments. Catchments are classified as snowy when median annual SWE exceeds 3 mm and the CAMELS-US snowfall fraction exceeds 0.05. In **a**, AM denotes the Appalachian Mountains, RM the Rocky Mountains, CRB the Columbia River Basin, SN the Sierra Nevada, CC the California Coast and RM + CRB catchments influenced by both the Rocky Mountains and Columbia River Basin.

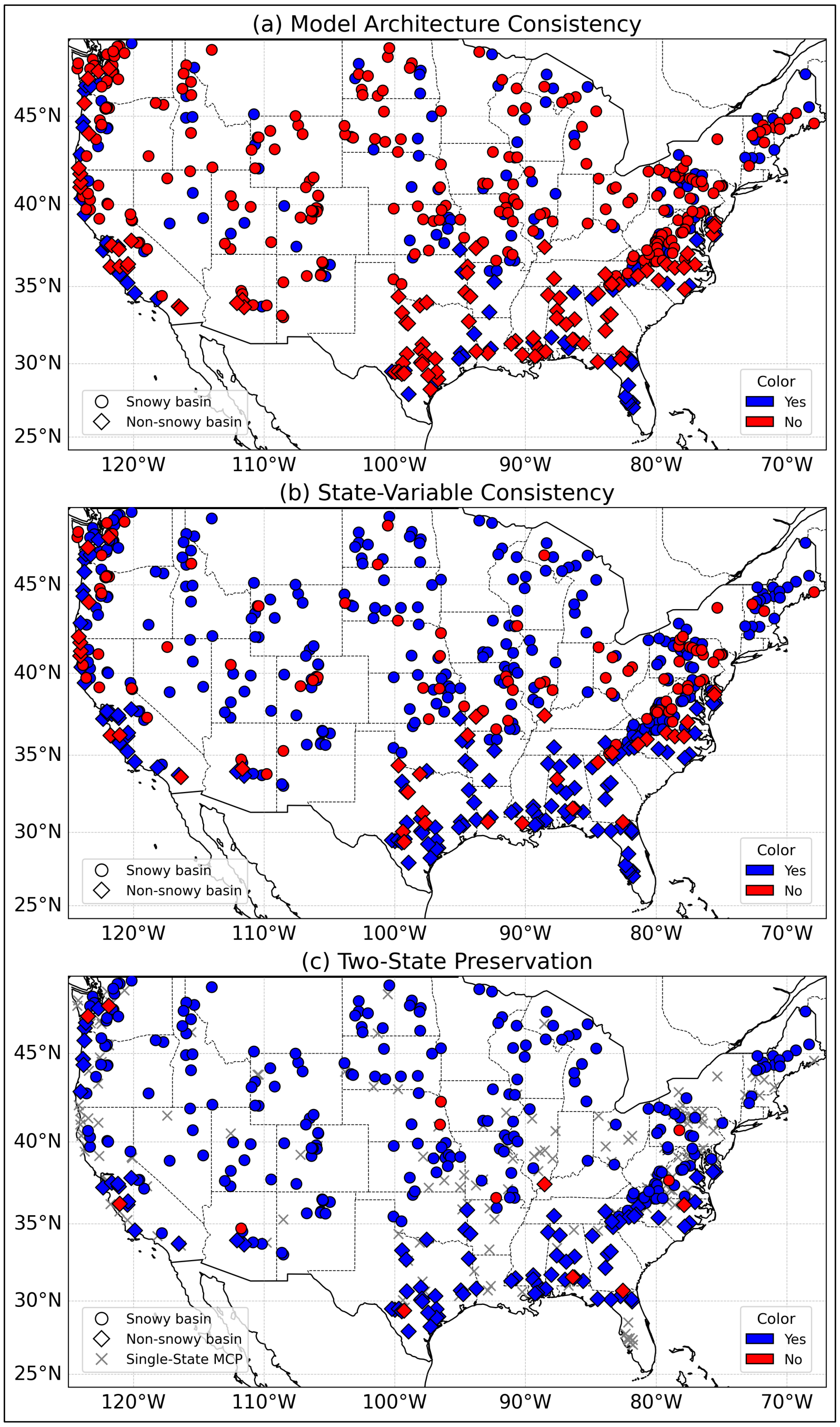

**Extended Data Fig. 2 | Architectural and state-dimensionality correspondence between conceptual hydrologic models and mass-conserving neural networks.**

**a–c,** Spatial comparison of the best-performing HYDROMCP and HYDROMCP-Net representations across 513 CAMELS-US catchments. Both formulations are evaluated using six candidate configurations comprising two single-state MCP models and four two-state architectures with different levels of cross-state information sharing. **a,** Agreement in the selected model architecture between the conceptual hydrologic-model and neural-network formulations. The same architecture is selected in 173 of 513 catchments (34%). **b,** Agreement in the selected number of state variables, irrespective of the specific architecture or information-sharing configuration. The same state dimensionality is retained in 386 catchments (75%). **c,** Preservation of two-state representations when transitioning from the conceptual hydrologic-model formulation to the neural-network formulation, irrespective of the information-sharing configuration. Of the 338 catchments selecting two-state conceptual architectures, 324 retain two states under the network formulation. Blue and red symbols indicate agreement (Yes) and disagreement (No), respectively. Gray crosses in c denote catchments selecting single-state conceptual architectures. Circles represent snowy catchments, whereas diamonds represent non-snowy catchments.

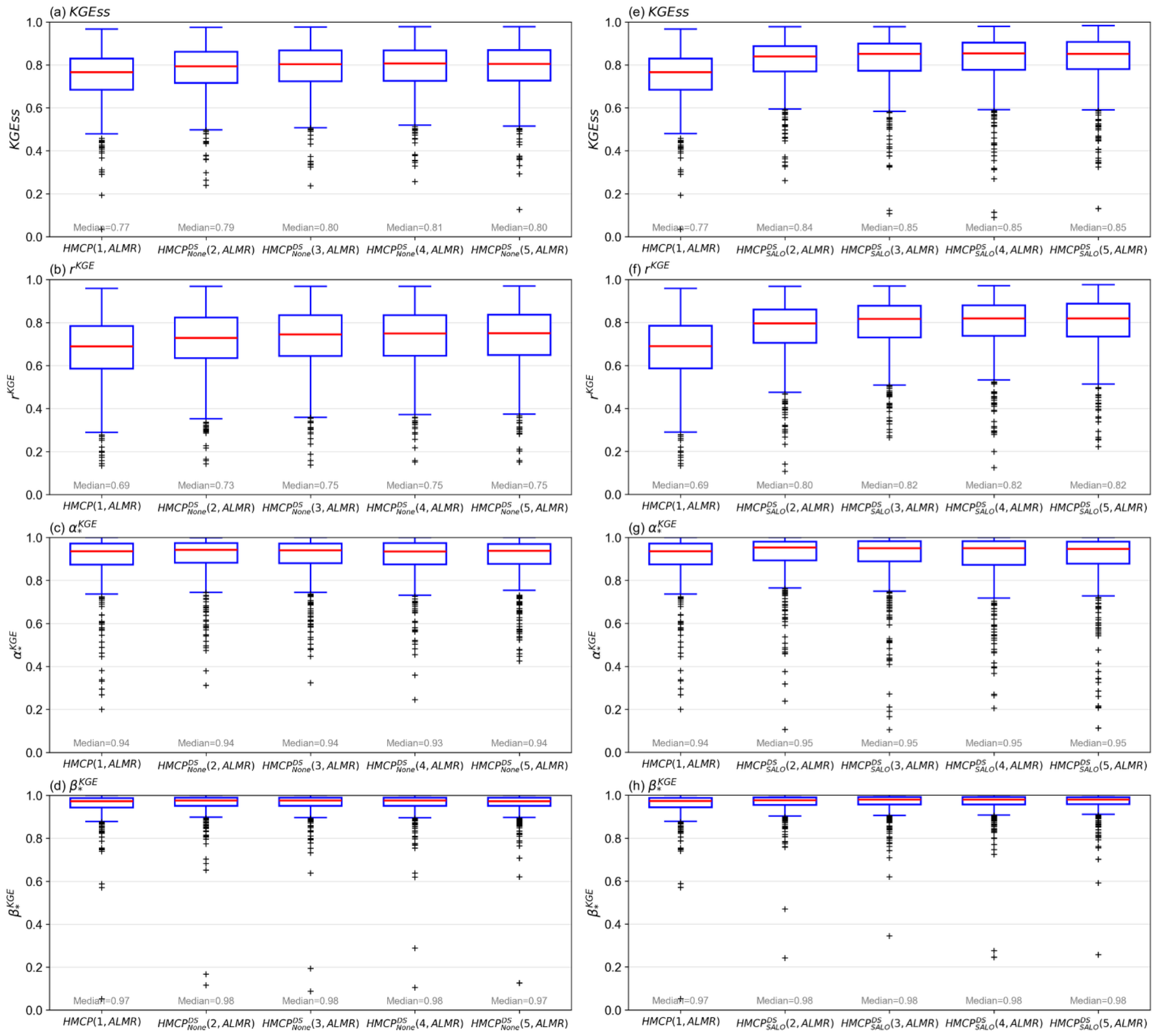

**Extended Data Fig. 3 | Predictive performance of mass-relaxation MCP networks across state dimensionality and connectivity.**

Box-and-whisker distributions of testing-period $KGE_{ss}$, $r^{KGE}$, $\alpha_*^{KGE}$ and $\beta_*^{KGE}$ across 513 CONUS catchments for mass-relaxation HMCP networks. **a–d,** Non-sharing distributed-state configurations with state dimensionality increasing from one to five. **e–h,** Corresponding sharing-enabled configurations across the same state-dimensionality range. Gray annotations report cross-catchment median values. Boxes indicate the interquartile range, red lines denote medians, whiskers extend to 1.5 times the interquartile range and crosses denote values beyond the whiskers.

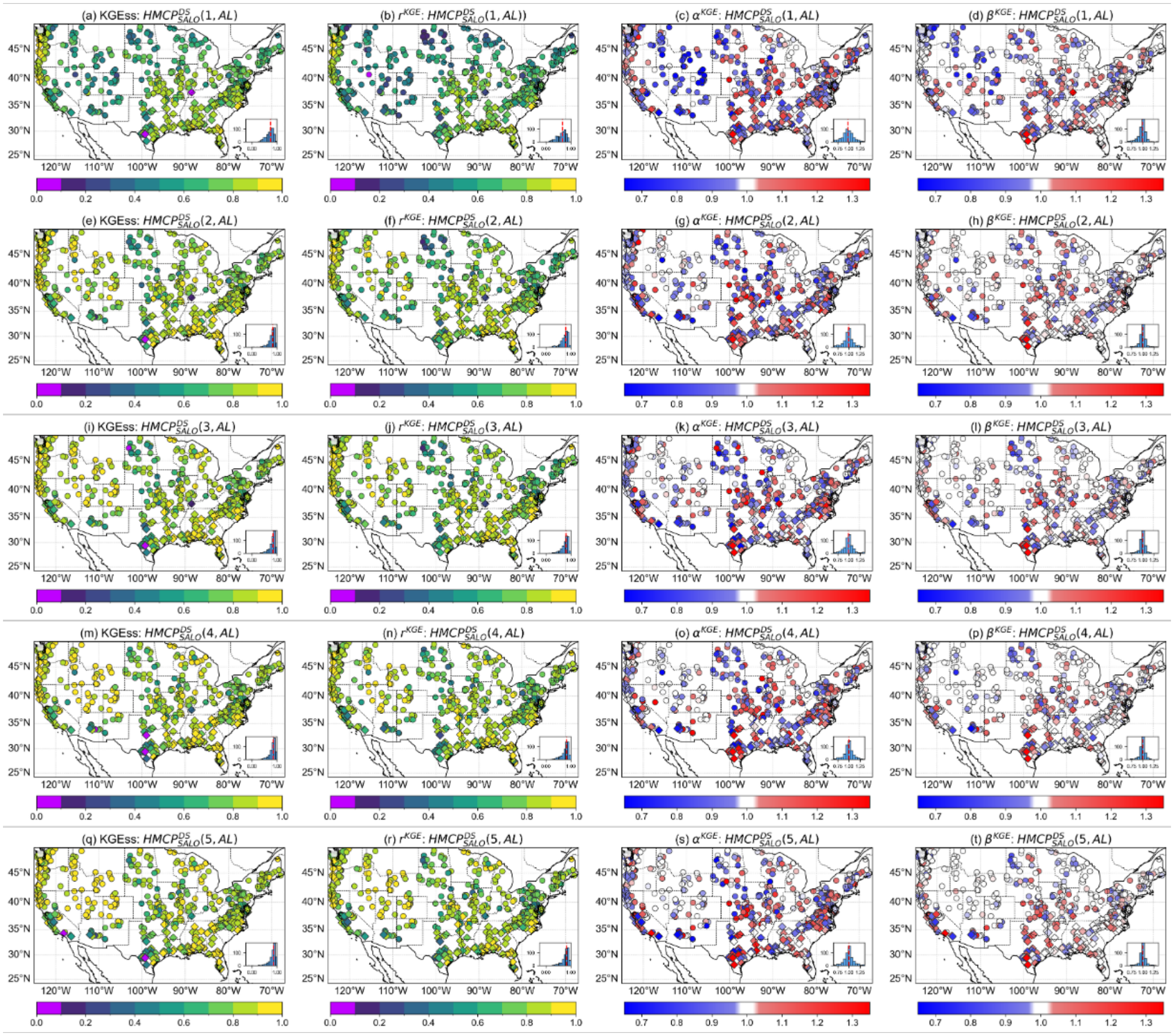

**Extended Data Fig. 4 | Spatial performance of fully connected HMCP distributed-state networks across state dimensionality.**

Testing-period performance of fully connected $HMCP_{SALO}^{DS}$ networks across 513 CONUS catchments. **a–d,** One-state model; **e–h,** two-state network; **i–l,** three-state network; **m–p,** four-state network; and **q–t,** five-state network. For each dimensionality, panels show $KGE_{ss}$, $r^{KGE}$, $\alpha^{KGE}$ and $\beta^{KGE}$, respectively. Circles denote snowy catchments and diamonds denote non-snowy catchments.

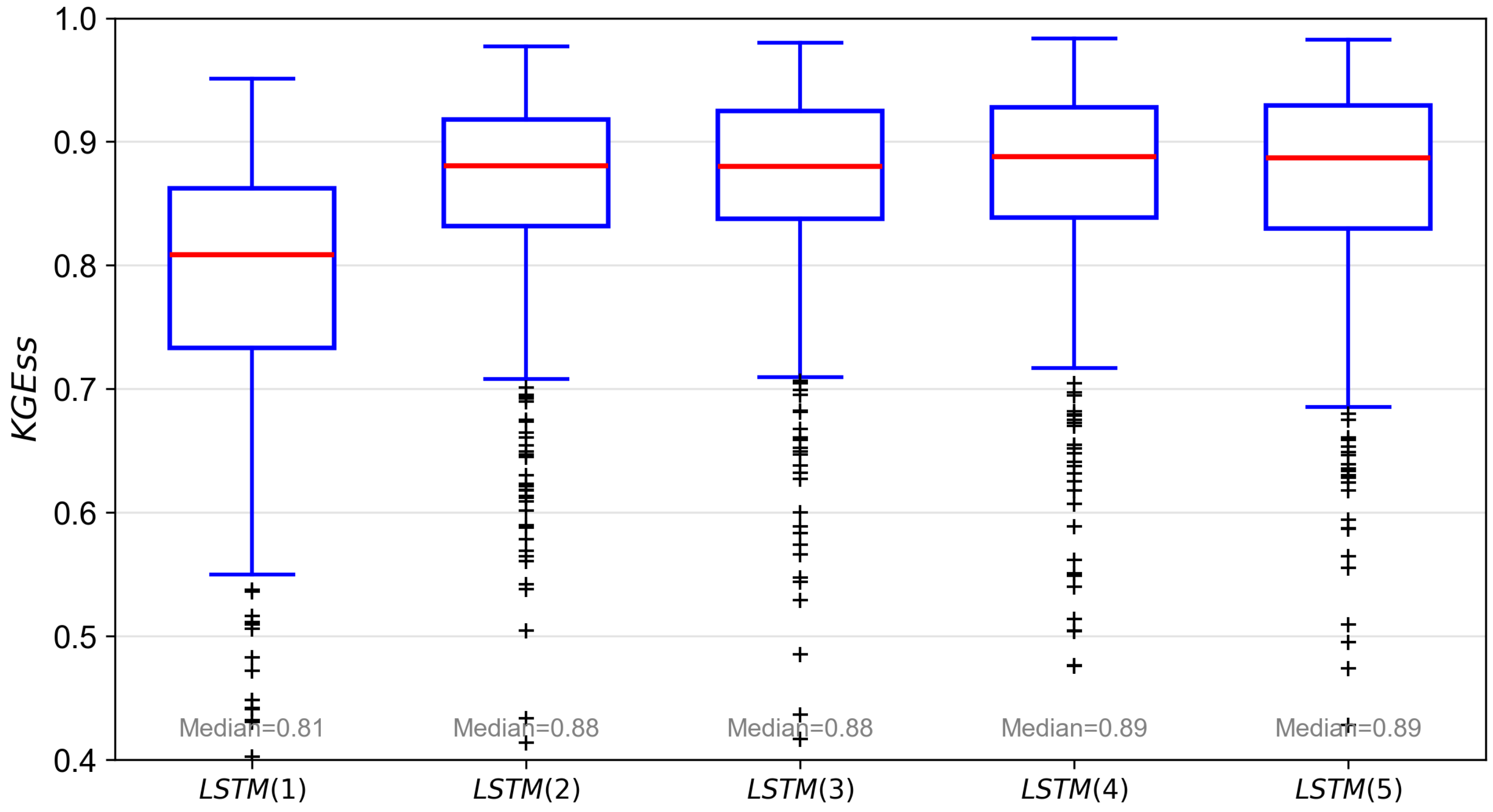

**Extended Data Fig. 5 | Predictive performance of LSTM networks across state dimensionality.**

Box-and-whisker distributions of testing-period $KGE_{ss}$, across 513 CAMELS-US catchments for basin-wise, single-layer LSTM networks with one to five hidden-state dimensions. Gray annotations report cross-catchment median $KGE_{ss}$. Boxes indicate the interquartile range, red lines denote medians, whiskers extend to 1.5 times the interquartile range and crosses denote values beyond the whiskers

**Extended Data Table 1 | KGE-based architecture selection within the two-state sharing-enabled MCP candidate space (KB2).**
For each hydroclimatic basin grouping, the table reports the number of catchments selecting each MCP model family (HMCP, SNOWMCP, HMCPMR and HYDROMCP-Net), together with the median testing-period Kling–Gupta efficiency skill score (KGEss), mean selected state dimensionality and mean number of trainable parameters. The candidate space permits cross-state information sharing and includes architectures with up to two states.

| Regional Categorization | KGE-based model selection | | | | | | |
|---|---|---|---|---|---|---|---|
| | HMCP | SNOWMCP | HMCPMR | HYDROMCP-Net | $KGE_{ss}^{50\%}$ | Ave State no | Ave Par no |
| CONUS | 72 | 216 | 101 | 124 | 0.88 | 2.00 | 26.15 |
| Eastern CONUS | 58 | 106 | 79 | 83 | 0.86 | 1.99 | 25.63 |
| Western CONUS | 14 | 110 | 22 | 41 | 0.91 | 2.00 | 27.06 |
| AM | 7 | 65 | 9 | 22 | 0.87 | 2.00 | 27.05 |
| RM | 1 | 50 | 0 | 3 | 0.95 | 2.00 | 29.02 |
| CRB | 1 | 22 | 5 | 6 | 0.91 | 2.00 | 27.35 |
| SN | 0 | 7 | 1 | 2 | 0.91 | 2.00 | 28.00 |
| CC | 0 | 24 | 1 | 8 | 0.91 | 2.00 | 27.70 |
| $EL < 1500$ m (low elevation) | 70 | 152 | 98 | 113 | 0.87 | 2.00 | 25.75 |
| $1500\ m <= EL <= 2500$ m (mid elevation) | 2 | 34 | 3 | 9 | 0.92 | 2.00 | 27.90 |
| $EL > 2500$ m (high elevation) | 0 | 30 | 0 | 2 | 0.95 | 2.00 | 29.00 |
| Aridity Index<1 (Energy Limited) | 39 | 149 | 63 | 75 | 0.88 | 2.00 | 26.41 |
| Aridity Index>1 (Water Limited) | 33 | 67 | 38 | 49 | 0.84 | 1.99 | 25.70 |
| Snowy | 33 | 200 | 44 | 83 | 0.88 | 2.00 | 26.84 |
| Non-Snowy | 39 | 16 | 57 | 41 | 0.86 | 1.99 | 24.55 |
| $Max\ SWE \leq$ 100mm | 66 | 45 | 90 | 80 | 0.84 | 1.99 | 24.78 |
| 100mm $<\ Max\ SWE\ \leq$ 200mm | 4 | 33 | 5 | 17 | 0.84 | 2.00 | 26.59 |
| 200mm $<\ Max\ SWE\ \leq$ 400mm | 2 | 60 | 4 | 14 | 0.89 | 2.00 | 28.03 |
| 400mm $<\ Max\ SWE\ \leq$ 600mm | 0 | 22 | 1 | 4 | 0.92 | 2.00 | 28.52 |
| $Max\ SWE$ > 600mm | 0 | 56 | 1 | 9 | 0.94 | 2.00 | 28.38 |
| Forest | 25 | 145 | 53 | 59 | 0.89 | 2.00 | 26.81 |
| Open | 47 | 71 | 48 | 65 | 0.84 | 2.00 | 25.35 |
| Climate Zone: Arid | 3 | 10 | 8 | 14 | 0.77 | 2.00 | 25.71 |
| Climate Zone: Temperate | 53 | 59 | 67 | 63 | 0.87 | 1.99 | 25.12 |
| Climate Zone: Cold | 16 | 145 | 26 | 47 | 0.89 | 2.00 | 27.27 |
| Climate Zone: Polar | 0 | 2 | 0 | 0 | 0.97 | 2.00 | 29.00 |

**Extended Data Table 2 | KGE-based architecture selection within the five-state no-sharing MCP candidate space (KC5).**
For each hydroclimatic basin grouping, the table reports the number of catchments selecting each MCP model family (HMCP, SNOWMCP, HMCPMR and HYDROMCP-Net), together with the median testing-period KGEss, mean selected state dimensionality and mean number of trainable parameters. The candidate space prohibits cross-state information sharing and includes architectures with up to five states.

| Regional Categorization | KGE-based model selection | | | | | | |
|---|---|---|---|---|---|---|---|
| | HMCP | SNOWMCP | HMCPMR | HYDROMCP-Net | $KGE_{ss}^{50\%}$ | Ave State no | Ave Par no |
| CONUS | 78 | 168 | 191 | 76 | 0.85 | 3.72 | 42.61 |
| Eastern CONUS | 65 | 73 | 145 | 43 | 0.83 | 3.70 | 42.02 |
| Western CONUS | 13 | 95 | 46 | 33 | 0.89 | 3.74 | 43.64 |
| AM | 17 | 48 | 27 | 11 | 0.83 | 3.93 | 45.15 |
| RM | 0 | 39 | 1 | 14 | 0.91 | 3.83 | 45.22 |
| CRB | 1 | 13 | 8 | 12 | 0.86 | 3.32 | 38.47 |
| SN | 0 | 6 | 2 | 2 | 0.87 | 3.70 | 43.80 |
| CC | 0 | 26 | 5 | 2 | 0.89 | 3.94 | 47.09 |
| $EL < 1500$ m (low elevation) | 76 | 113 | 182 | 62 | 0.84 | 3.69 | 42.07 |
| $1500\ m <= EL <= 2500$ m (mid elevation) | 2 | 27 | 8 | 11 | 0.87 | 3.65 | 42.63 |
| $EL > 2500$ m (high elevation) | 0 | 28 | 1 | 3 | 0.91 | 4.19 | 49.97 |
| Aridity Index<1 (Energy Limited) | 48 | 118 | 121 | 39 | 0.86 | 3.83 | 44.11 |
| Aridity Index>1 (Water Limited) | 30 | 50 | 70 | 37 | 0.81 | 3.51 | 40.01 |
| Snowy | 43 | 149 | 104 | 64 | 0.85 | 3.75 | 43.19 |
| Non-Snowy | 35 | 19 | 87 | 12 | 0.86 | 3.64 | 41.25 |
| $Max\ SWE \leq$ 100mm | 70 | 29 | 150 | 32 | 0.83 | 3.62 | 40.54 |
| 100mm $< Max\ SWE \leq$ 200mm | 6 | 16 | 23 | 14 | 0.80 | 3.73 | 43.02 |
| 200mm $< Max\ SWE \leq$ 400mm | 2 | 50 | 13 | 15 | 0.84 | 3.89 | 45.83 |
| 400mm $< Max\ SWE \leq$ 600mm | 0 | 17 | 2 | 8 | 0.89 | 3.44 | 40.44 |
| $Max\ SWE >$ 600mm | 0 | 56 | 3 | 7 | 0.90 | 4.03 | 48.05 |
| Forest | 35 | 112 | 101 | 34 | 0.86 | 3.92 | 45.33 |
| Open | 43 | 56 | 90 | 42 | 0.82 | 3.47 | 39.29 |
| Climate Zone: Arid | 3 | 7 | 12 | 13 | 0.73 | 3.14 | 35.83 |
| Climate Zone: Temperate | 54 | 50 | 120 | 18 | 0.86 | 3.74 | 42.33 |
| Climate Zone: Cold | 21 | 110 | 59 | 44 | 0.85 | 3.79 | 43.98 |
| Climate Zone: Polar | 0 | 1 | 0 | 1 | 0.94 | 3.00 | 34.50 |

**Extended Data Table 3 | Likelihood-based (A5) and KGE-based (KA5) architecture selection across hydroclimatic basin groupings.**
For each basin grouping, the table reports the number of catchments selecting MCP-based or long short-term memory (LSTM) architectures, together with the median testing-period KGEss, mean selected state dimensionality and mean number of trainable parameters. Results are based on the full candidate spaces permitting up to five states (A5 and KA5). Values outside parentheses correspond to likelihood-based selection (A5), whereas values in parentheses correspond to KGE-based selection (KA5). HYDROMCP architectures are included within the MCP-based category rather than reported separately.

| Regional Categorization | Likelihood-based (A5) and KGE-based (KA5) architecture selection | | | | |
|---|---|---|---|---|---|
| | MCP-Based Net | LSTM Net | $KGE_{ss}^{50\%}$ | Ave State no | Ave Par no |
| CONUS | 201(163) | 312(350) | 0.90(0.89) | 4.42(4.47) | 117.34(123.70) |
| Eastern CONUS | 117(91) | 209(235) | 0.88(0.88) | 4.33(4.41) | 115.16(123.46) |
| Western CONUS | 84(72) | 103(115) | 0.93(0.93) | 4.58(4.58) | 121.14(124.12) |
| AM | 50(46) | 53(57) | 0.88(0.88) | 4.46(4.53) | 114.31(120.60) |
| RM | 30(29) | 24(25) | 0.97(0.97) | 4.83(4.83) | 126.87(126.69) |
| CRB | 11(11) | 23(23) | 0.95(0.95) | 4.76(4.79) | 137.97(137.12) |
| SN | 4(3) | 6(7) | 0.93(0.92) | 4.80(4.70) | 130.30(133.60) |
| CC | 27(20) | 6(13) | 0.92(0.92) | 4.48(4.52) | 100.15(110.15) |
| $EL < 1500$ m (low elevation) | 165(129) | 268(304) | 0.88(0.88) | 4.35(4.41) | 114.76(122.32) |
| $1500\ m <= EL <= 2500$ m (mid elevation) | 17(16) | 31(32) | 0.95(0.95) | 4.79(4.73) | 137.54(134.96) |
| $EL > 2500$ m (high elevation) | 19(18) | 13(14) | 0.97(0.97) | 4.81(4.88) | 121.91(125.44) |
| Aridity Index<1 (Energy Limited) | 141(112) | 185(214) | 0.90(0.90) | 4.39(4.45) | 114.11(121.75) |
| Aridity Index>1 (Water Limited) | 60(51) | 127(136) | 0.87(0.87) | 4.47(4.50) | 122.96(127.10) |
| Snowy | 161(139) | 199(221) | 0.90(0.90) | 4.51(4.56) | 118.71(123.91) |
| Non-Snowy | 40(24) | 113(129) | 0.88(0.87) | 4.20(4.25) | 114.11(123.20) |
| $Max\ SWE \leq$ 100mm | 70(44) | 211(237) | 0.87(0.87) | 4.28(4.36) | 117.84(127.03) |
| 100mm $< Max\ SWE \leq$ 200mm | 19(14) | 40(45) | 0.87(0.87) | 4.44(4.53) | 122.81(130.92) |
| 200mm $< Max\ SWE \leq$ 400mm | 50(50) | 30(30) | 0.93(0.92) | 4.58(4.59) | 112.86(112.96) |
| 400mm $< Max\ SWE \leq$ 600mm | 13(10) | 14(17) | 0.94(0.94) | 4.48(4.63) | 121.37(130.59) |
| $Max\ SWE >$ 600mm | 49(45) | 17(21) | 0.96(0.96) | 4.80(4.70) | 114.09(113.24) |
| Forest | 130(107) | 152(175) | 0.91(0.91) | 4.47(4.51) | 115.49(122.54) |
| Open | 71(56) | 160(175) | 0.87(0.87) | 4.36(4.42) | 119.59(125.11) |
| Climate Zone: Arid | 7(8) | 28(27) | 0.87(0.87) | 4.57(4.69) | 135.23(137.60) |
| Climate Zone: Temperate | 70(43) | 172(199) | 0.88(0.88) | 4.24(4.32) | 114.87(124.80) |
| Climate Zone: Cold | 122(110) | 112(124) | 0.91(0.91) | 4.59(4.59) | 117.59(120.68) |
| Climate Zone: Polar | 2(2) | 0(0) | 0.98(0.98) | 4.00(5.00) | 74.00(100.00) |

# Supplementary Note 1

## HMCP state update and gate parameterization

The single-node HMCP represents rainfall-runoff dynamics through the discrete-time mass-balance equation (S1)

$$X_{t+1} = X_t - O_t - L_t + U_t \tag{S1}$$

where $X_t$ is the conceptual soil-moisture storage state, $U_t$ is precipitation input, $O_t$ is streamflow, and $L_t$ is evapotranspirative loss.

The output and loss fluxes are computed as

$$O_t = G_t^O X_t \tag{S2}$$

$$L_t = G_t^L X_t \tag{S3}$$

where $G_t^O$ and $G_t^L$ are time-varying output and loss conductivity gates. The state update can therefore be expressed as

$$X_{t+1} = G_t^R X_t + U_t \tag{S4}$$

where $G_t^R$ is the remember gate. All three gates are constrained to remain between zero and one, with

$$G_t^R + G_t^O + G_t^L = 1 \tag{S5}$$

Accordingly, the remember gate is calculated as

$$G_t^R = 1 - G_t^O - G_t^L \tag{S6}$$

When system losses are omitted, as in the surface-channel and groundwater-routing applications of Wang and Gupta (2024b), the update reduces to $X_{t+1} = X_t - O_t + U_t$, and the conservation constraint becomes $G_t^R + G_t^O = 1$.

Following Wang and Gupta (2024a,b), the output and loss gates are parameterized as

$$G_t^O = \kappa_O \sigma(a_O + b_O X_t) \tag{S7}$$

$$G_t^L = \kappa_L \sigma(a_L + b_L PET_t) \tag{S8}$$

PET refers to potential evapotranspiration. In Wang and Gupta (2025), the loss gate further incorporates the cell state as an independent variable and is parameterized as

$$G_t^L = \kappa_L \sigma(a_L + b_L PET_t + b_L^* X_t) \tag{S9}$$

where $\sigma(\cdot)$ is the sigmoid activation function. Because the slope parameters $b_O$ , $b_L$ and $b_L^*$ are unconstrained, the gate responses may be either increasing or decreasing with respect to their contextual inputs.

The scaling coefficients are obtained using a Softmax normalization:

$$\kappa_O = \frac{\exp(c_O)}{\Psi} \tag{S10}$$

$$\kappa_L = \frac{\exp(c_L)}{\Psi} \tag{S11}$$

with

$$\Psi = \exp(c_O) + \exp(c_L) + \exp(c_R) \tag{S12}$$

This formulation ensures that the output, loss, and remember gates remain nonnegative and jointly satisfy the mass-conservation constraint. The eight trainable parameters are $\{a_O, b_O, c_O, a_L, b_L, b_L^*, c_L, c_R\}$.

## Supplementary Note 2

### SNOWMCP state update and gate parameterization

The single-state SNOWMCP computational unit introduced by Wang et al. (2025) represents snowpack accumulation, melt, and snow-related loss using a conceptual snow-water-equivalent storage state. Following the nodal mass-conservation principle used in HMCP, the snow-storage state is updated as

$$SWE_{t+1} = SWE_t - O_t^s - L_t^s + U_t^s \tag{S13}$$

where $SWE_t$ is the conceptual snow-water-equivalent (SWE) storage at time step $t$, $U_t^s$ is snowfall input, $O_t^s$ is snowmelt output, and $L_t^s$ represents snow-related mass loss through processes such as sublimation and evaporation during the interval from $t$ to $t+1$.

Total precipitation input $U_t$ is partitioned into snowfall and rainfall using a time-varying rain–snow partition gate:

$$U_t^s = G_t^{RS} U_t \tag{S14}$$

$$U_t^R = (1 - G_t^{RS}) U_t = U_t - U_t^s \tag{S15}$$

where $G_t^{RS}$ is bounded between zero and one. Accordingly, $U_t^s$ enters the snow-storage node, whereas the rainfall component $U_t^R$ bypasses the snow storage.

Snowmelt and snow-related loss are computed as fractions of the current snow-storage state:

$$O_t^s = G_t^M SWE_t \tag{S16}$$

$$L_t^s = G_t^{SL} SWE_t \tag{S17}$$

where $G_t^M$ and $G_t^{SL}$ are the time-varying melt and snow-loss gates, respectively. Substitution into Equation (S13) gives

$$SWE_{t+1} = SWE_t - G_t^M SWE_t - G_t^{SL} SWE_t + U_t^s \tag{S18}$$

which can be written as

$$SWE_{t+1} = G_t^{SR} SWE_t + U_t^s \tag{S19}$$

where $G_t^{SR}$ is the snow remember gate and represents the fraction of $SWE_t$ retained until the next time step. Nodal mass conservation is enforced through

$$G_t^{SR} + G_t^M + G_t^{SL} = 1 \tag{S20}$$

such that

$$G_t^{SR} = 1 - G_t^M - G_t^{SL} \tag{S21}$$

The rain–snow partition, melt, and snow-loss gates are parameterized using sigmoid functions. Their contextual scores are defined as

$$S_t^{RS} = a_{RS} + b_{RS}T_t \tag{S22}$$

$$S_t^{M} = a_M + b_M T_t + b_M^* SWE_t \tag{S23}$$

$$S_t^{SL} = a_{SL} + b_{SL}T_t + b_{SL}^* SWE_t \tag{S24}$$

where $T_t$ is mean air temperature at time step $t$. The corresponding gates are

$$G_t^{RS} = \sigma(S_t^{RS}) \tag{S25}$$

$$G_t^{M} = \kappa_M \sigma(S_t^{M}) \tag{S26}$$

$$G_t^{SL} = \kappa_{SL} \sigma(S_t^{SL}) \tag{S27}$$

where $\sigma(\cdot)$ is the sigmoid activation function. The slope parameters $b_{RS}$, $b_M$, $b_M^*$, $b_{SL}$, and $b_{SL}^*$ are unconstrained and allowed to be any real values.

To ensure that the melt, snow-loss, and remember gates jointly satisfy the mass-conservation constraint, the scaling coefficients are obtained using a Softmax-based normalization:

$$\kappa_M = \frac{\exp(c_M)}{\Psi} \tag{S28}$$

$$\kappa_{SL} = \frac{\exp(c_{SL})}{\Psi} \tag{S29}$$

with

$$\Psi = \exp(c_M) + \exp(c_{SL}) + \exp(c_{SR}) \tag{S30}$$

Because the denominator includes the additional term $\exp(c_{SR})$, the sum of $\kappa_M$ and $\kappa_{SL}$ does not exceed one. Together with the sigmoid bounds, this formulation ensures that $G_t^M$ and $G_t^{SL}$ remain nonnegative and that the resulting remember gate $G_t^{SR}$ remains nonnegative. The formulation contains 11 trainable parameters: $\{a_{RS}, b_{RS}, a_M, b_M, b_M^*, c_M, a_{SL}, b_{SL}, b_{SL}^*, c_{SL}, c_{SR}\}$ all of which are unconstrained real-valued parameters. The coefficients $\kappa_M$ and $\kappa_{SL}$ are derived from $c_M$, $c_{SL}$, and $c_{SR}$ and are therefore not independently trained.

Wang et al. (2025) considered several SNOWMCP variants that incorporated different combinations of contextual variables into the gate functions. Because the differences in predictive performance among these variants were limited, the present study uses the simplest configuration, in which the rain–snow partition gate depends on mean temperature, while the melt and snow-loss gates depend on mean temperature and the internal SWE state.

## Supplementary Note 3

### HMCP mass-relaxation formulation

To represent bidirectional mass exchange with unresolved components of the hydrologic system, the mass-relaxation mechanism proposed by Wang and Gupta (2024a) is incorporated into the single-node HMCP architecture. The mechanism determines the direction and magnitude of mass exchange according to the current storage state relative to a learned reference level.

The storage state is first scaled as

$$\tilde{X}_t = \frac{X_t}{s_{MR}}, s_{MR} = 500 \quad \text{(S31)}$$

where $s_{MR}$ is a fixed scaling factor. The reference level in the scaled domain, $\tilde{c}_{MR}$, is an unconstrained trainable parameter, and its corresponding value in the original storage domain is

$$c_{MR} = s_{MR}\tilde{c}_{MR} \quad \text{(S32)}$$

The unadjusted mass-relaxation function is defined as

$$f_t^{MR} = \kappa_{MR}\tanh\left[a_{MR}\left(\tilde{X}_t - \tilde{c}_{MR}\right)\right] \quad \text{(S33)}$$

where $\kappa_{MR}$ controls the maximum absolute magnitude of the mass-relaxation response and $a_{MR}$ controls the steepness of the transition around the learned reference level. In the implementation, these quantities are parameterized as

$$\kappa_{MR} = \sigma(w_{MR}^r), a_{MR} = \exp(w_{MR}^s) \quad \text{(S34)}$$

where $w_{MR}^r$ and $w_{MR}^s$ are unconstrained trainable parameters. Consequently,

$$0 < \kappa_{MR} < 1, a_{MR} > 0, \tilde{c}_{MR} \in \mathbb{R} \quad \text{(S35)}$$

Because the hyperbolic tangent function is bounded between $-1$ and $1$,

$$-\kappa_{MR} < f_t^{MR} < \kappa_{MR} \quad \text{(S36)}$$

When $X_t > c_{MR}$, $f_t^{MR} > 0$, indicating outward mass exchange from the node to the unresolved environment. Conversely, when $X_t < c_{MR}$, $f_t^{MR} < 0$, indicating inward mass exchange from the unresolved environment to the node. No mass exchange occurs when $X_t = c_{MR}$.

Because the learned reference level $\tilde{c}_{MR}$ is unconstrained, it may lie outside the attained range of the scaled storage state. When $\tilde{c}_{MR}$ lies below the attained storage range, mass exchange is exclusively outward. Conversely, when $\tilde{c}_{MR}$ lies above the attained storage range, mass exchange is exclusively inward. When $\tilde{c}_{MR}$ lies within the attained storage range, the direction of exchange may vary over time.

Before mass relaxation is applied, the fraction of storage remaining after the output and loss fluxes is

$$\bar{G}_t^R = 1 - G_t^O - G_t^L \quad \text{(S37)}$$

where $G_t^O$ and $G_t^L$ are the output and loss gates, respectively. To cap the positive mass-relaxation gate by the retained fraction, the adjusted gate is computed as

$$G_t^{MR} = f_t^{MR} - \text{ReLU}(f_t^{MR} - \bar{G}_t^R) \quad \text{(S38)}$$

Equivalently,

$$G_t^{MR} = \min(f_t^{MR}, \bar{G}_t^R) \tag{S39}$$

This adjustment caps positive values of the mass-relaxation gate by the fraction retained after the ordinary output and loss fluxes. Negative values are not truncated because they represent mass entering the node. Accordingly,

$$-\kappa_{MR} < G_t^{MR} \leq \min(\kappa_{MR}, \bar{G}_t^R) \tag{S40}$$

The mass exchanged with the unresolved environment at time step $t$is

$$q_t^{MR} = G_t^{MR} \mid X_t - c_{MR} \mid \tag{S41}$$

Under this sign convention, $q_t^{MR} > 0$represents mass leaving the node, whereas $q_t^{MR} < 0$represents mass entering the node.

The HMCP state update with mass relaxation is therefore

$$X_{t+1} = X_t - O_t - L_t - q_t^{MR} + U_t \tag{S42}$$

or, equivalently,

$$X_{t+1} = \bar{G}_t^R X_t + U_t - q_t^{MR} \tag{S43}$$

where $U_t$ , $O_t$ , and $L_t$ denote precipitation input, streamflow output, and evapotranspirative loss, respectively. Because inward exchange is represented by a negative value of $q_t^{MR}$, subtracting $q_t^{MR}$ adds the incoming mass to the storage state.

The mass-relaxation formulation introduces three unconstrained trainable parameters:

$$\{\tilde{c}_{MR}, w_{MR}^r, w_{MR}^s\} \tag{S44}$$

The quantities $c_{MR}$, $\kappa_{MR}$, and $a_{MR}$ are derived from these parameters using the fixed scaling factor, sigmoid transformation, and exponential transformation, respectively.

The mass-relaxation gating function used here is one of several formulations introduced by Wang and Gupta (2024a); readers are referred to that study for further details. Parameters were initialized from pretrained model components to provide hydrologically informed starting points and encourage physically plausible state trajectories, although this initialization does not guarantee state non-negativity. Given the focus of this study on the duality between hydrologic models and neural networks, we retain the mass-relaxation formulation without imposing an additional non-negativity constraint on the internal state. The formulation preserves the algebraic balance among inputs, outputs, losses, and exchange, but the gate bound in Equation (S38) does not necessarily restrict outward exchange to the available storage when the learned reference level is negative. If negative states occur, they should therefore be interpreted as signed internal representations of the neural network rather than as literal physical water storage.

## Supplementary Note 4.

### Model training, parameter initialization, and learning-rate schedules

All MCP-based models were trained using the Adam optimizer, following the optimization procedure adopted in our previous MCP study (Wang et al., 2025). Ten random seeds were used: 2925, 9998, 2025, 2525, 3410, 9899, 5555, 2520, 2828, and 3140. Model parameters were optimized by maximizing the Kling–Gupta efficiency (KGE; Gupta et al., 2009) over the training subset. To minimize sensitivity to the initialization of recurrent storage states, the first water year was repeated three times before the simulation period to provide a three-year spin-up. Unless otherwise specified, models were trained for up to 3,000 epochs with an initial learning rate of 0.025, which was reduced to 0.0125 after 500 epochs.

Model development followed a stagewise parameter initialization procedure. The single-state HMCP and SNOWMCP components were trained independently before being integrated into the coupled HYDROMCP architectures. Parameters associated with previously trained components were inherited from their corresponding pretrained models, whereas parameters introduced by additional model components or cross-state information-sharing connections were randomly initialized from a uniform distribution over the interval [−1, 1].

The baseline two-state HYDROMCP architecture was initialized using the corresponding pretrained HMCP and SNOWMCP components. For architectures incorporating cross-state information sharing, two initialization strategies were evaluated: **progressive initialization** and **baseline initialization**.

Under progressive initialization, each one-way-sharing architecture inherited its existing parameters from the trained baseline HYDROMCP architecture, while the newly introduced cross-state-sharing parameters were randomly initialized from a uniform distribution over [−1, 1]. The bidirectional-sharing architecture subsequently inherited its common parameters, including the Softmax output weights, from the baseline HYDROMCP architecture, while the two sets of cross-state-sharing parameters were transferred from their corresponding pretrained one-way-sharing architectures.

Under baseline initialization, each information-sharing architecture was initialized directly from the trained baseline HYDROMCP architecture, with all common parameters inherited without modification, except for the Softmax output weights. These weights were reinitialized while maintaining the sum-to-one constraint, allowing the relative contributions of the HMCP and SNOWMCP components to be relearned during optimization.

For each basin and HYDROMCP architecture, both initialization strategies were evaluated across the same ten random seeds using identical optimization settings. Among the resulting model realizations, the one achieving the highest KGE on the independent architecture-selection subset was retained for subsequent evaluation.

## Supplementary References

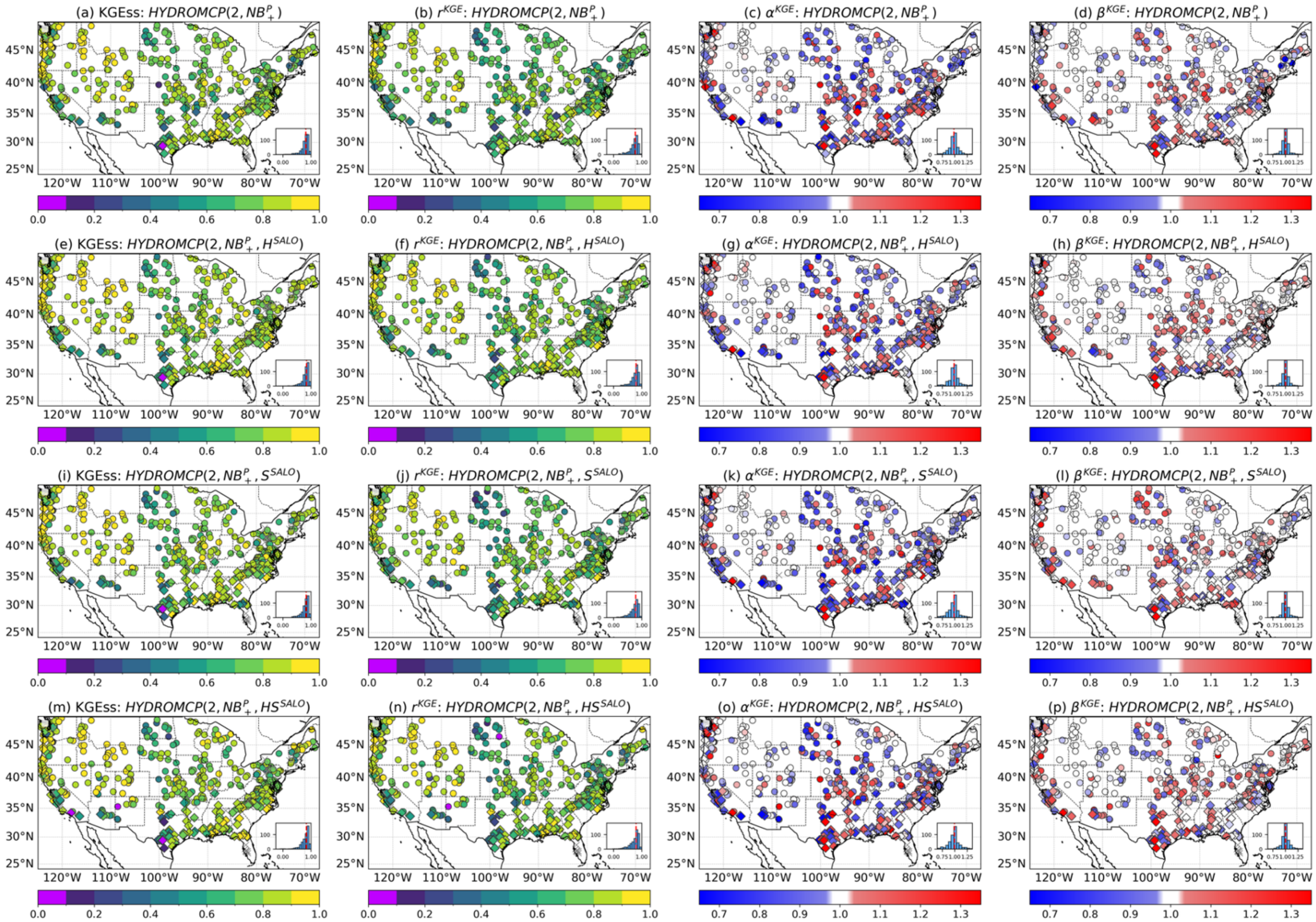

**Fig. S1 | Spatial performance of two-state HYDROMCP architectures with alternative cross-state information-sharing configurations.**

**a–d,** Testing-period performance of the baseline two-state HYDROMCP architecture without cross-state information sharing. **e–h,** Corresponding performance when the HMCP gating functions additionally receive information from the SNOWMCP state (RR-share). **i–l,** Corresponding performance when the SNOWMCP gating functions additionally receive information from the HMCP state (SNOW-share). **m–p,** Corresponding performance with bidirectional cross-state information sharing (RR–SNOW-share). Within each architecture, panels report Kling–Gupta efficiency skill score ($KGE_{ss}$), linear correlation ($r^{KGE}$), flow-variability ratio ($\alpha^{KGE}$) and mass-balance ratio ($\beta^{KGE}$). Circles denote snowy catchments and diamonds denote non-snowy catchments. Results are shown for 513 CONUS catchments.

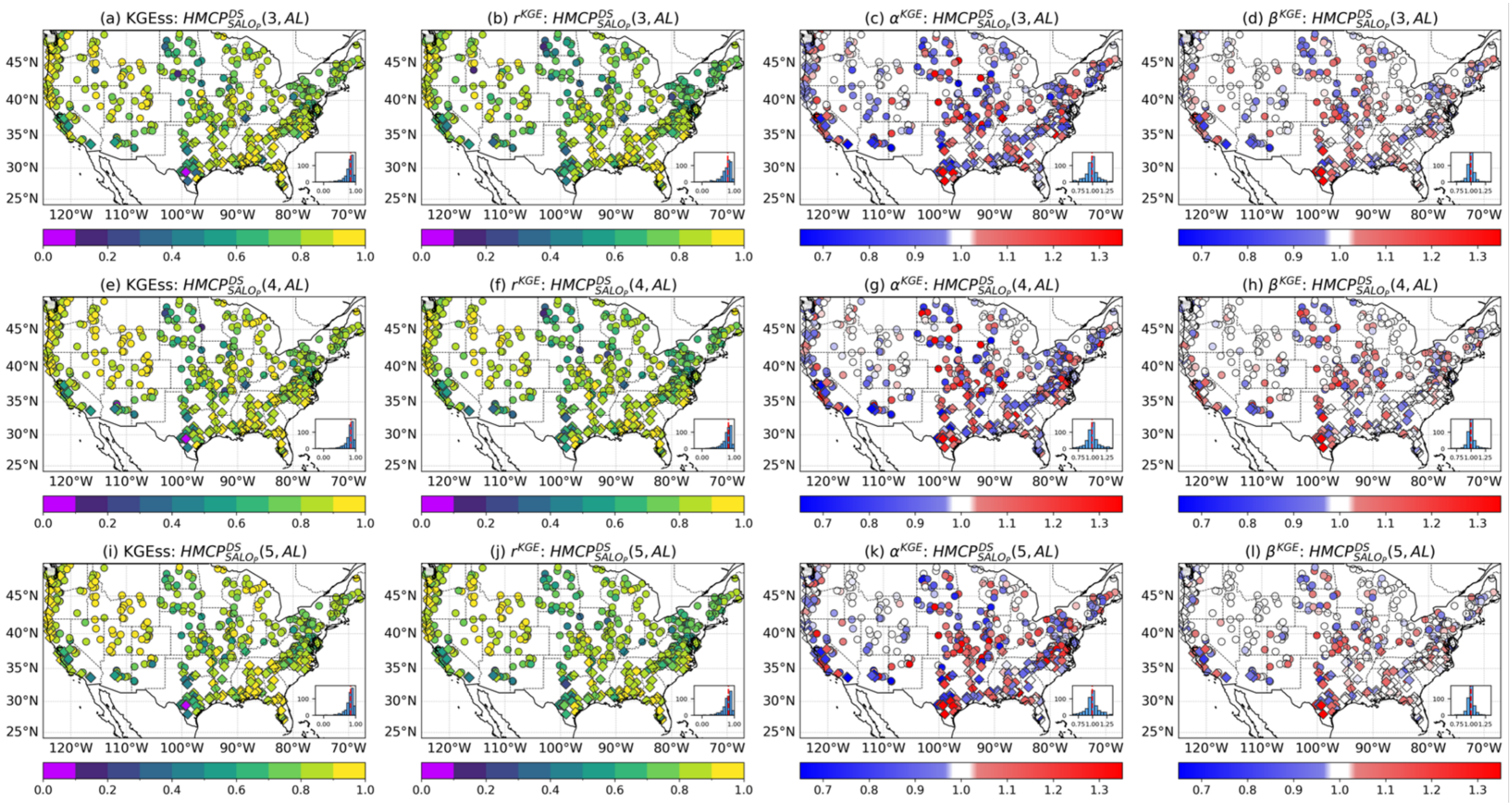

**Fig. S2 | Spatial performance of partially connected HMCP distributed-state networks across state dimensionality.**

Testing-period performance of partially connected $HMCP^{DS}_{SALO_p}$ networks across 513 CONUS catchments. **a–d,** Three-state network; **e–h,** four-state network; and **i–l,** five-state network. For each dimensionality, panels show $KGE_{ss}$, $r^{KGE}$, $\alpha^{KGE}$ and $\beta^{KGE}$, respectively. Circles denote snowy catchments and diamonds denote non-snowy catchments.

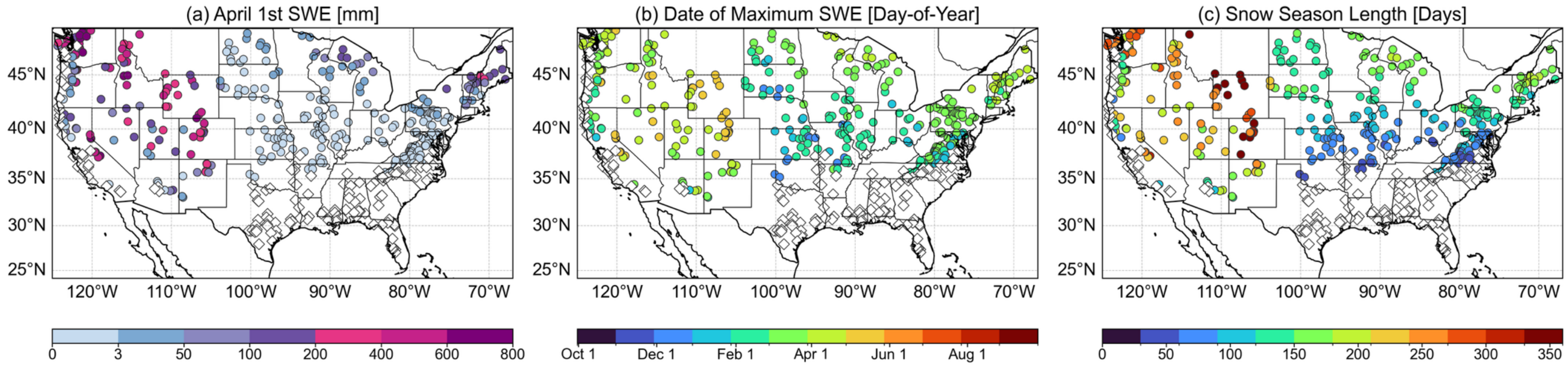

**Fig. S3 | Spatial characteristics of seasonal snow regimes across the study catchments.**

**a–c,** Snow-regime characteristics across the 513 CAMELS-US catchments, showing April 1 SWE (**a**), timing of annual maximum SWE (**b**) and snow-season length (**c**). Circles denote snowy catchments and diamonds denote non-snowy catchments. Catchments are classified as snowy when median annual SWE exceeds 3 mm and the CAMELS-US snowfall fraction exceeds 0.05. Snow metrics are summarized using the annual median over the simulation period.

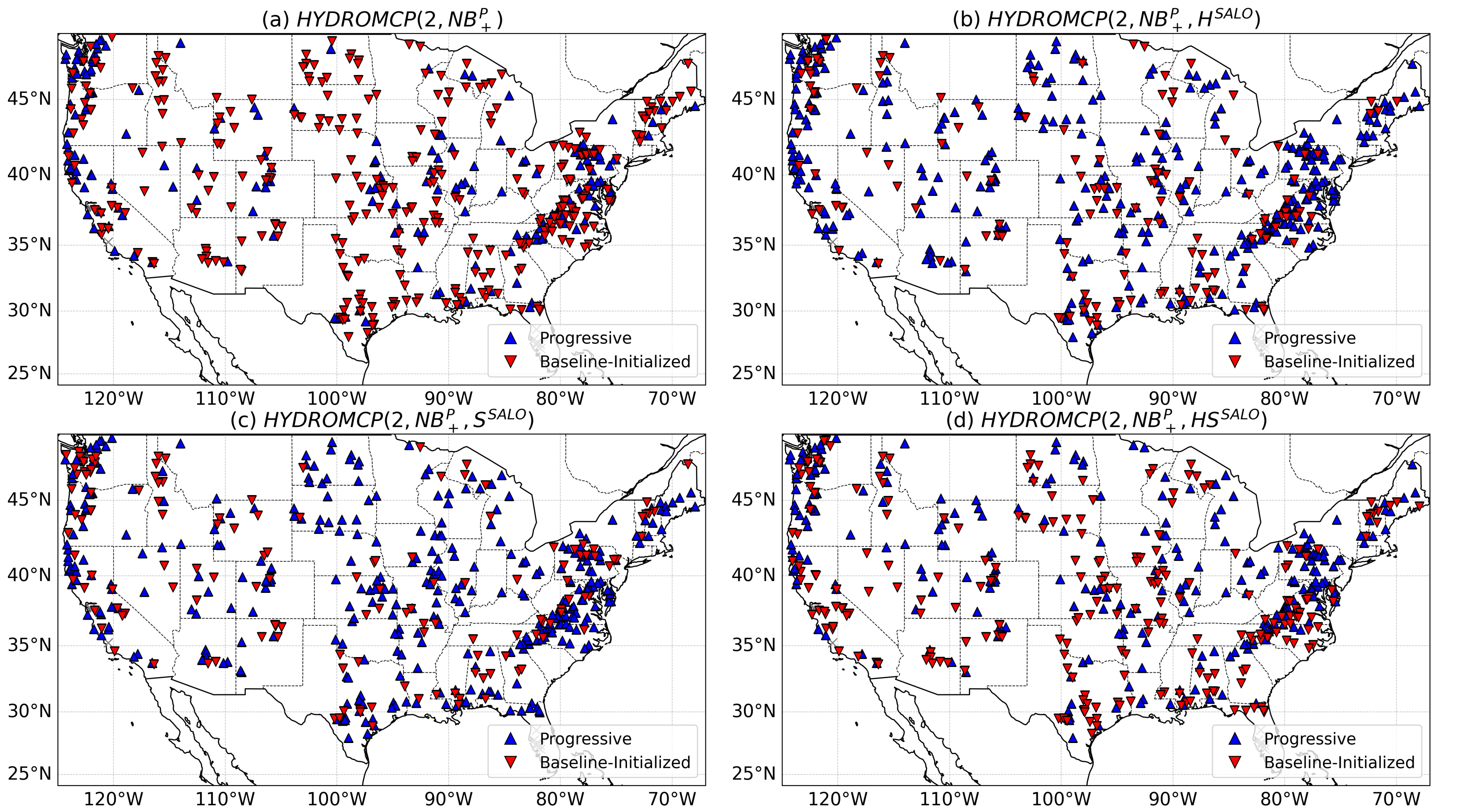

**Fig. S4 | Spatial comparison of progressive and baseline initialization strategies for HYDROMCP architectures.**

Preferred initialization strategy for each of 513 CONUS catchments for the baseline two-state HYDROMCP architecture (**a**), HYDROMCP with RR-share (**b**), HYDROMCP with SNOW-share (**c**) and HYDROMCP with bidirectional RR–SNOW-share (**d**). Blue upward triangles indicate catchments for which progressive initialization yields the higher KGE on the independent architecture-selection subset, whereas red downward triangles indicate catchments for which baseline initialization performs better. For each catchment and architecture, the initialization yielding the higher architecture-selection KGE is retained for subsequent testing-period evaluation.

**Table S1 | Likelihood-based architecture selection across the full candidate space with a maximum of one state.** For each basin grouping, the table reports the number of catchments assigned to each model family, median testing-period $KGE_{ss}$, mean selected state dimensionality and mean number of trainable parameters. The candidate space includes all eligible MCP-based and LSTM architectures containing at most one state.

| Regional Categorization | Likelihood-based model selection | | | | |
|---|---|---|---|---|---|
| | MCP-Based Net | LSTM Net | $KGE_{ss}^{50\%}$ | Ave State no | Ave Par no |
| CONUS | 217 | 296 | 0.83 | 1.00 | 11.42 |
| Eastern CONUS | 145 | 181 | 0.81 | 1.00 | 11.32 |
| Western CONUS | 72 | 115 | 0.87 | 1.00 | 11.58 |
| AM | 39 | 64 | 0.80 | 1.00 | 11.53 |
| RM | 34 | 20 | 0.90 | 1.00 | 11.37 |
| CRB | 19 | 15 | 0.86 | 1.00 | 11.35 |
| SN | 5 | 5 | 0.90 | 1.00 | 11.50 |
| CC | 10 | 23 | 0.88 | 1.00 | 11.70 |
| $EL < 1500$ m (low elevation) | 170 | 263 | 0.82 | 1.00 | 11.41 |
| $1500\ m <= EL <= 2500$ m (mid elevation) | 20 | 28 | 0.87 | 1.00 | 11.60 |
| $EL > 2500$ m (high elevation) | 27 | 5 | 0.90 | 1.00 | 11.16 |
| Aridity Index<1 (Energy Limited) | 137 | 189 | 0.84 | 1.00 | 11.44 |
| Aridity Index>1 (Water Limited) | 80 | 107 | 0.80 | 1.00 | 11.37 |
| Snowy | 157 | 203 | 0.83 | 1.00 | 11.47 |
| Non-Snowy | 60 | 93 | 0.83 | 1.00 | 11.29 |
| $Max\ SWE \leq$ 100mm | 129 | 152 | 0.82 | 1.00 | 11.27 |
| 100mm < $Max\ SWE \leq$ 200mm | 15 | 44 | 0.78 | 1.00 | 11.69 |
| 200mm < $Max\ SWE \leq$ 400mm | 23 | 57 | 0.81 | 1.00 | 11.64 |
| 400mm < $Max\ SWE \leq$ 600mm | 16 | 11 | 0.89 | 1.00 | 11.44 |
| $Max\ SWE$ > 600mm | 34 | 32 | 0.89 | 1.00 | 11.48 |
| Forest | 111 | 171 | 0.84 | 1.00 | 11.46 |
| Open | 106 | 125 | 0.80 | 1.00 | 11.36 |
| Climate Zone: Arid | 7 | 28 | 0.75 | 1.00 | 11.69 |
| Climate Zone: Temperate | 100 | 142 | 0.84 | 1.00 | 11.36 |
| Climate Zone: Cold | 108 | 126 | 0.82 | 1.00 | 11.44 |
| Climate Zone: Polar | 2 | 0 | 0.93 | 1.00 | 11.00 |

**Table S2 | Likelihood-based architecture selection across the full candidate space with a maximum of 2 states.** For each basin grouping, the table reports the number of catchments assigned to each model family, median testing-period $KGE_{ss}$, mean selected state dimensionality and mean number of trainable parameters. The candidate space includes all eligible MCP-based and LSTM architectures containing at most 2 states.

| Regional Categorization | Likelihood-based model selection | | | | |
|---|---|---|---|---|---|
| | MCP-Based Net | LSTM Net | $KGE_{ss}^{50\%}$ | Ave State no | Ave Par no |
| CONUS | 219 | 294 | 0.89 | 1.97 | 35.53 |
| Eastern CONUS | 132 | 194 | 0.87 | 1.98 | 35.83 |
| Western CONUS | 87 | 100 | 0.92 | 1.96 | 34.99 |
| AM | 57 | 46 | 0.87 | 1.99 | 34.48 |
| RM | 34 | 20 | 0.95 | 2.00 | 34.24 |
| CRB | 15 | 19 | 0.93 | 1.97 | 35.44 |
| SN | 4 | 6 | 0.90 | 2.00 | 37.20 |
| CC | 26 | 7 | 0.91 | 2.00 | 31.73 |
| $EL < 1500$ m (low elevation) | 175 | 258 | 0.88 | 1.97 | 35.63 |
| $1500\ m\ <= EL <=\ 2500$ m (mid elevation) | 23 | 25 | 0.93 | 1.98 | 35.75 |
| $EL > 2500$ m (high elevation) | 21 | 11 | 0.95 | 2.00 | 33.78 |
| Aridity Index<1 (Energy Limited) | 155 | 171 | 0.89 | 1.99 | 35.51 |
| Aridity Index>1 (Water Limited) | 64 | 123 | 0.86 | 1.94 | 35.56 |
| Snowy | 184 | 176 | 0.89 | 1.98 | 34.86 |
| Non-Snowy | 35 | 118 | 0.87 | 1.95 | 37.09 |
| $Max\ SWE \leq$ 100mm | 77 | 204 | 0.87 | 1.95 | 36.79 |
| 100mm $<\ Max\ SWE\ \leq$ 200mm | 22 | 37 | 0.87 | 1.97 | 36.68 |
| 200mm $<\ Max\ SWE\ \leq$ 400mm | 55 | 25 | 0.90 | 2.00 | 33.23 |
| 400mm $<\ Max\ SWE\ \leq$ 600mm | 12 | 15 | 0.92 | 2.00 | 36.26 |
| $Max\ SWE >$ 600mm | 53 | 13 | 0.94 | 2.00 | 31.62 |
| Forest | 141 | 141 | 0.90 | 1.99 | 35.12 |
| Open | 78 | 153 | 0.87 | 1.95 | 36.03 |
| Climate Zone: Arid | 7 | 28 | 0.83 | 1.86 | 35.34 |
| Climate Zone: Temperate | 81 | 161 | 0.88 | 1.97 | 36.45 |
| Climate Zone: Cold | 129 | 105 | 0.89 | 1.99 | 34.67 |
| Climate Zone: Polar | 2 | 0 | 0.97 | 2.00 | 28.00 |

**Table S3 | Likelihood-based architecture selection across the full candidate space with a maximum of 3 states.** For each basin grouping, the table reports the number of catchments assigned to each model family, median testing-period $KGE_{ss}$, mean selected state dimensionality and mean number of trainable parameters. The candidate space includes all eligible MCP-based and LSTM architectures containing at most 3 states.

| Regional Categorization | Likelihood-based model selection | | | | |
|---|---|---|---|---|---|
| | MCP-Based Net | LSTM Net | $KGE_{ss}^{50\%}$ | Ave State no | Ave Par no |
| CONUS | 201 | 312 | 0.89 | 2.83 | 60.32 |
| Eastern CONUS | 117 | 209 | 0.88 | 2.82 | 60.35 |
| Western CONUS | 84 | 103 | 0.92 | 2.86 | 60.26 |
| AM | 55 | 48 | 0.88 | 2.83 | 57.43 |
| RM | 32 | 22 | 0.96 | 2.96 | 60.07 |
| CRB | 13 | 21 | 0.94 | 2.88 | 62.56 |
| SN | 1 | 9 | 0.93 | 2.80 | 66.90 |
| CC | 27 | 6 | 0.92 | 2.94 | 54.00 |
| $EL < 1500$ m (low elevation) | 166 | 267 | 0.88 | 2.81 | 59.78 |
| $1500\ m <= EL <= 2500$ m (mid elevation) | 19 | 29 | 0.94 | 2.90 | 63.29 |
| $EL > 2500$ m (high elevation) | 16 | 16 | 0.96 | 3.00 | 63.13 |
| Aridity Index<1 (Energy Limited) | 147 | 179 | 0.90 | 2.85 | 59.42 |
| Aridity Index>1 (Water Limited) | 54 | 133 | 0.87 | 2.80 | 61.88 |
| Snowy | 166 | 194 | 0.90 | 2.84 | 59.01 |
| Non-Snowy | 35 | 118 | 0.88 | 2.81 | 63.39 |
| $Max\ SWE \leq$ 100mm | 62 | 219 | 0.87 | 2.78 | 62.46 |
| 100mm < $Max\ SWE \leq$ 200mm | 21 | 38 | 0.88 | 2.83 | 61.22 |
| 200mm < $Max\ SWE \leq$ 400mm | 59 | 21 | 0.92 | 2.88 | 54.20 |
| 400mm < $Max\ SWE \leq$ 600mm | 11 | 16 | 0.93 | 3.00 | 65.37 |
| $Max\ SWE$ > 600mm | 48 | 18 | 0.95 | 2.94 | 55.73 |
| Forest | 132 | 150 | 0.91 | 2.85 | 59.53 |
| Open | 69 | 162 | 0.87 | 2.81 | 61.28 |
| Climate Zone: Arid | 6 | 29 | 0.83 | 2.80 | 64.37 |
| Climate Zone: Temperate | 69 | 173 | 0.88 | 2.81 | 62.05 |
| Climate Zone: Cold | 124 | 110 | 0.90 | 2.86 | 58.02 |
| Climate Zone: Polar | 2 | 0 | 0.98 | 3.00 | 49.50 |

**Table S4 | Likelihood-based architecture selection across the full candidate space with a maximum of** 4 **states.** For each basin grouping, the table reports the number of catchments assigned to each model family, median testing-period $KGE_{ss}$, mean selected state dimensionality and mean number of trainable parameters. The candidate space includes all eligible MCP-based and LSTM architectures containing at most 4 states.

| Regional Categorization | Likelihood-based model selection | | | | |
|---|---|---|---|---|---|
| | MCP-Based Net | LSTM Net | $KGE_{ss}^{50\%}$ | Ave State no | Ave Par no |
| CONUS | 202 | 311 | 0.90 | 3.63 | 86.40 |
| Eastern CONUS | 117 | 209 | 0.88 | 3.56 | 85.11 |
| Western CONUS | 85 | 102 | 0.93 | 3.75 | 88.65 |
| AM | 51 | 52 | 0.88 | 3.66 | 83.92 |
| RM | 34 | 20 | 0.97 | 3.94 | 88.41 |
| CRB | 13 | 21 | 0.94 | 3.76 | 91.85 |
| SN | 2 | 8 | 0.93 | 3.70 | 95.70 |
| CC | 28 | 5 | 0.91 | 3.64 | 70.91 |
| $EL < 1500$ m (low elevation) | 163 | 270 | 0.89 | 3.58 | 85.30 |
| $1500\ m\ <= EL <=\ 2500$ m (mid elevation) | 22 | 26 | 0.95 | 3.85 | 92.58 |
| $EL > 2500$ m (high elevation) | 17 | 15 | 0.97 | 3.94 | 92.06 |
| Aridity Index<1 (Energy Limited) | 142 | 184 | 0.90 | 3.61 | 84.10 |
| Aridity Index>1 (Water Limited) | 60 | 127 | 0.87 | 3.66 | 90.41 |
| Snowy | 168 | 192 | 0.90 | 3.66 | 84.72 |
| Non-Snowy | 34 | 119 | 0.88 | 3.56 | 90.37 |
| $Max\ SWE \leq$ 100mm | 60 | 221 | 0.87 | 3.53 | 89.18 |
| 100mm $<\ Max\ SWE\ \leq$ 200mm | 23 | 36 | 0.87 | 3.71 | 89.86 |
| 200mm $<\ Max\ SWE\ \leq$ 400mm | 56 | 24 | 0.92 | 3.69 | 78.24 |
| 400mm $<\ Max\ SWE\ \leq$ 600mm | 13 | 14 | 0.94 | 3.74 | 88.37 |
| $Max\ SWE$ > 600mm | 50 | 16 | 0.95 | 3.85 | 80.56 |
| Forest | 128 | 154 | 0.90 | 3.66 | 85.37 |
| Open | 74 | 157 | 0.87 | 3.58 | 87.66 |
| Climate Zone: Arid | 6 | 29 | 0.84 | 3.66 | 97.00 |
| Climate Zone: Temperate | 65 | 177 | 0.88 | 3.55 | 87.89 |
| Climate Zone: Cold | 129 | 105 | 0.90 | 3.71 | 83.50 |
| Climate Zone: Polar | 2 | 0 | 0.98 | 3.50 | 60.00 |

**Table S5 | Likelihood-based architecture selection across the full candidate space with a maximum of** 5 **states.** For each basin grouping, the table reports the number of catchments assigned to each model family, median testing-period $KGE_{ss}$, mean selected state dimensionality and mean number of trainable parameters. The candidate space includes all eligible MCP-based and LSTM architectures containing at most 5 states.

| Regional Categorization | Likelihood-based model selection | | | | |
|---|---|---|---|---|---|
| | MCP-Based Net | LSTM Net | $KGE_{ss}^{50\%}$ | Ave State no | Ave Par no |
| CONUS | 201 | 312 | 0.90 | 4.42 | 117.34 |
| Eastern CONUS | 117 | 209 | 0.88 | 4.33 | 115.16 |
| Western CONUS | 84 | 103 | 0.93 | 4.58 | 121.14 |
| AM | 50 | 53 | 0.88 | 4.46 | 114.31 |
| RM | 30 | 24 | 0.97 | 4.83 | 126.87 |
| CRB | 11 | 23 | 0.95 | 4.76 | 137.97 |
| SN | 4 | 6 | 0.93 | 4.80 | 130.30 |
| CC | 27 | 6 | 0.92 | 4.48 | 100.15 |
| $EL < 1500$ m (low elevation) | 165 | 268 | 0.88 | 4.35 | 114.76 |
| $1500\ m\ <= EL <=\ 2500$ m (mid elevation) | 17 | 31 | 0.95 | 4.79 | 137.54 |
| $EL > 2500$ m (high elevation) | 19 | 13 | 0.97 | 4.81 | 121.91 |
| Aridity Index<1 (Energy Limited) | 141 | 185 | 0.90 | 4.39 | 114.11 |
| Aridity Index>1 (Water Limited) | 60 | 127 | 0.87 | 4.47 | 122.96 |
| Snowy | 161 | 199 | 0.90 | 4.51 | 118.71 |
| Non-Snowy | 40 | 113 | 0.88 | 4.20 | 114.11 |
| $Max\ SWE \leq$ 100mm | 70 | 211 | 0.87 | 4.28 | 117.84 |
| 100mm $<\ Max\ SWE\ \leq$ 200mm | 19 | 40 | 0.87 | 4.44 | 122.81 |
| 200mm $<\ Max\ SWE\ \leq$ 400mm | 50 | 30 | 0.93 | 4.58 | 112.86 |
| 400mm $<\ Max\ SWE\ \leq$ 600mm | 13 | 14 | 0.94 | 4.48 | 121.37 |
| $Max\ SWE$ > 600mm | 49 | 17 | 0.96 | 4.80 | 114.09 |
| Forest | 130 | 152 | 0.91 | 4.47 | 115.49 |
| Open | 71 | 160 | 0.87 | 4.36 | 119.59 |
| Climate Zone: Arid | 7 | 28 | 0.87 | 4.57 | 135.23 |
| Climate Zone: Temperate | 70 | 172 | 0.88 | 4.24 | 114.87 |
| Climate Zone: Cold | 122 | 112 | 0.91 | 4.59 | 117.59 |
| Climate Zone: Polar | 2 | 0 | 0.98 | 4.00 | 74.00 |

**Table S6 | KGE-based architecture selection across the full candidate space with a maximum of one state.** For each basin grouping, the table reports the number of catchments assigned to each model family, median testing-period $KGE_{ss}$, mean selected state dimensionality and mean number of trainable parameters. The candidate space includes all eligible MCP-based and LSTM architectures containing at most one state.

| Regional Categorization | KGE-based model selection | | | | |
|---|---|---|---|---|---|
| | MCP-Based Net | LSTM Net | $KGE_{ss}^{50\%}$ | Ave State no | Ave Par no |
| CONUS | 269 | 244 | 0.83 | 1.00 | 11.24 |
| Eastern CONUS | 187 | 139 | 0.81 | 1.00 | 11.10 |
| Western CONUS | 82 | 105 | 0.87 | 1.00 | 11.48 |
| AM | 48 | 55 | 0.80 | 1.00 | 11.45 |
| RM | 34 | 20 | 0.90 | 1.00 | 11.37 |
| CRB | 23 | 11 | 0.86 | 1.00 | 11.15 |
| SN | 7 | 3 | 0.90 | 1.00 | 11.30 |
| CC | 8 | 25 | 0.88 | 1.00 | 11.76 |
| $EL$ < 1500 m (low elevation) | 219 | 214 | 0.82 | 1.00 | 11.22 |
| 1500 $m$ <= $EL$ <= 2500 m (mid elevation) | 22 | 26 | 0.87 | 1.00 | 11.48 |
| $EL$ > 2500 m (high elevation) | 28 | 4 | 0.90 | 1.00 | 11.13 |
| Aridity Index<1 (Energy Limited) | 156 | 170 | 0.84 | 1.00 | 11.32 |
| Aridity Index>1 (Water Limited) | 113 | 74 | 0.81 | 1.00 | 11.09 |
| Snowy | 190 | 170 | 0.83 | 1.00 | 11.29 |
| Non-Snowy | 79 | 74 | 0.83 | 1.00 | 11.11 |
| $Max\ SWE \leq$ 100mm | 163 | 118 | 0.82 | 1.00 | 11.01 |
| 100mm < $Max\ SWE$ ≤ 200mm | 30 | 29 | 0.78 | 1.00 | 11.49 |
| 200mm < $Max\ SWE$ ≤ 400mm | 29 | 51 | 0.81 | 1.00 | 11.53 |
| 400mm < $Max\ SWE$ ≤ 600mm | 16 | 11 | 0.89 | 1.00 | 11.41 |
| $Max\ SWE$ > 600mm | 31 | 35 | 0.89 | 1.00 | 11.53 |
| Forest | 132 | 150 | 0.85 | 1.00 | 11.39 |
| Open | 137 | 94 | 0.81 | 1.00 | 11.04 |
| Climate Zone: Arid | 16 | 19 | 0.75 | 1.00 | 11.29 |
| Climate Zone: Temperate | 119 | 123 | 0.84 | 1.00 | 11.21 |
| Climate Zone: Cold | 132 | 102 | 0.82 | 1.00 | 11.26 |
| Climate Zone: Polar | 2 | 0 | 0.93 | 1.00 | 11.00 |

**Table S7 | KGE-based architecture selection across the full candidate space with a maximum of 2 states.** For each basin grouping, the table reports the number of catchments assigned to each model family, median testing-period $KGE_{ss}$, mean selected state dimensionality and mean number of trainable parameters. The candidate space includes all eligible MCP-based and LSTM architectures containing at most 2 states.

| Regional Categorization | KGE-based model selection | | | | |
|---|---|---|---|---|---|
| | MCP-Based Net | LSTM Net | $KGE_{ss}^{50\%}$ | Ave State no | Ave Par no |
| CONUS | 180 | 333 | 0.89 | 2.00 | 37.34 |
| Eastern CONUS | 115 | 211 | 0.87 | 2.00 | 37.10 |
| Western CONUS | 65 | 122 | 0.91 | 2.00 | 37.75 |
| AM | 50 | 53 | 0.87 | 2.00 | 35.79 |
| RM | 27 | 27 | 0.95 | 2.00 | 36.15 |
| CRB | 11 | 23 | 0.93 | 2.00 | 38.44 |
| SN | 2 | 8 | 0.90 | 2.00 | 40.00 |
| CC | 17 | 16 | 0.91 | 2.00 | 34.88 |
| $EL < 1500$ m (low elevation) | 149 | 284 | 0.88 | 2.00 | 37.31 |
| $1500\ m <= EL <= 2500$ m (mid elevation) | 16 | 32 | 0.93 | 2.00 | 38.10 |
| $EL > 2500$ m (high elevation) | 15 | 17 | 0.95 | 2.00 | 36.53 |
| Aridity Index<1 (Energy Limited) | 124 | 202 | 0.89 | 2.00 | 37.01 |
| Aridity Index>1 (Water Limited) | 56 | 131 | 0.87 | 2.00 | 37.90 |
| Snowy | 149 | 211 | 0.89 | 2.00 | 36.54 |
| Non-Snowy | 31 | 122 | 0.88 | 2.00 | 39.22 |
| $Max\ SWE \leq$ 100mm | 63 | 218 | 0.87 | 2.00 | 38.81 |
| 100mm $<\ Max\ SWE\ \leq$ 200mm | 20 | 39 | 0.86 | 2.00 | 37.73 |
| 200mm $<\ Max\ SWE\ \leq$ 400mm | 48 | 32 | 0.90 | 2.00 | 34.35 |
| 400mm $<\ Max\ SWE\ \leq$ 600mm | 9 | 18 | 0.92 | 2.00 | 38.11 |
| $Max\ SWE$ > 600mm | 40 | 26 | 0.94 | 2.00 | 34.00 |
| Forest | 110 | 172 | 0.90 | 2.00 | 37.00 |
| Open | 70 | 161 | 0.87 | 2.00 | 37.74 |
| Climate Zone: Arid | 8 | 27 | 0.86 | 2.00 | 38.89 |
| Climate Zone: Temperate | 55 | 187 | 0.88 | 2.00 | 38.89 |
| Climate Zone: Cold | 115 | 119 | 0.89 | 2.00 | 35.56 |
| Climate Zone: Polar | 2 | 0 | 0.97 | 2.00 | 29.00 |

**Table S8 | KGE-based architecture selection across the full candidate space with a maximum of 3 states.** For each basin grouping, the table reports the number of catchments assigned to each model family, median testing-period $KGE_{ss}$, mean selected state dimensionality and mean number of trainable parameters. The candidate space includes all eligible MCP-based and LSTM architectures containing at most 3 states.

| Regional Categorization | KGE-based model selection | | | | |
|---|---|---|---|---|---|
| | MCP-Based Net | LSTM Net | $KGE_{ss}^{50\%}$ | Ave State no | Ave Par no |
| CONUS | 172 | 341 | 0.89 | 2.91 | 64.06 |
| Eastern CONUS | 101 | 225 | 0.88 | 2.90 | 64.15 |
| Western CONUS | 71 | 116 | 0.92 | 2.94 | 63.91 |
| AM | 50 | 53 | 0.88 | 2.90 | 60.25 |
| RM | 31 | 23 | 0.96 | 2.98 | 61.04 |
| CRB | 12 | 22 | 0.94 | 2.97 | 65.76 |
| SN | 2 | 8 | 0.93 | 3.00 | 70.40 |
| CC | 21 | 12 | 0.92 | 3.00 | 59.27 |
| $EL < 1500$ m (low elevation) | 137 | 296 | 0.88 | 2.90 | 64.05 |
| $1500\ m\ <= EL <= \ 2500$ m (mid elevation) | 18 | 30 | 0.94 | 2.98 | 65.44 |
| $EL > 2500$ m (high elevation) | 17 | 15 | 0.96 | 3.00 | 62.25 |
| Aridity Index<1 (Energy Limited) | 121 | 205 | 0.90 | 2.92 | 63.54 |
| Aridity Index>1 (Water Limited) | 51 | 136 | 0.87 | 2.89 | 64.98 |
| Snowy | 150 | 210 | 0.90 | 2.93 | 62.55 |
| Non-Snowy | 22 | 131 | 0.88 | 2.88 | 67.62 |
| $Max\ SWE \leq$ 100mm | 45 | 236 | 0.87 | 2.86 | 66.87 |
| 100mm $<\ Max\ SWE\ \leq$ 200mm | 19 | 40 | 0.87 | 2.92 | 64.86 |
| 200mm $<\ Max\ SWE\ \leq$ 400mm | 54 | 26 | 0.92 | 2.98 | 57.63 |
| 400mm $<\ Max\ SWE\ \leq$ 600mm | 10 | 17 | 0.93 | 3.00 | 66.19 |
| $Max\ SWE >$ 600mm | 44 | 22 | 0.95 | 2.98 | 58.33 |
| Forest | 111 | 171 | 0.91 | 2.94 | 63.51 |
| Open | 61 | 170 | 0.87 | 2.88 | 64.74 |
| Climate Zone: Arid | 7 | 28 | 0.86 | 2.91 | 67.91 |
| Climate Zone: Temperate | 46 | 196 | 0.88 | 2.89 | 66.98 |
| Climate Zone: Cold | 117 | 117 | 0.90 | 2.93 | 60.60 |
| Climate Zone: Polar | 2 | 0 | 0.98 | 3.00 | 49.50 |

**Table S9 | KGE-based architecture selection across the full candidate space with a maximum of 4 states.** For each basin grouping, the table reports the number of catchments assigned to each model family, median testing-period $KGE_{ss}$, mean selected state dimensionality and mean number of trainable parameters. The candidate space includes all eligible MCP-based and LSTM architectures containing at most 4 states.

| Regional Categorization | KGE-based model selection | | | | |
|---|---|---|---|---|---|
| | MCP-Based Net | LSTM Net | $KGE_{ss}^{50\%}$ | Ave State no | Ave Par no |
| CONUS | 170 | 343 | 0.89 | 3.68 | 91.24 |
| Eastern CONUS | 96 | 230 | 0.88 | 3.63 | 90.82 |
| Western CONUS | 74 | 113 | 0.93 | 3.79 | 91.96 |
| AM | 49 | 54 | 0.88 | 3.68 | 85.64 |
| RM | 30 | 24 | 0.97 | 3.96 | 92.26 |
| CRB | 10 | 24 | 0.94 | 3.82 | 97.35 |
| SN | 4 | 6 | 0.93 | 3.70 | 87.10 |
| CC | 20 | 13 | 0.91 | 3.88 | 87.36 |
| $EL < 1500$ m (low elevation) | 134 | 299 | 0.88 | 3.65 | 90.94 |
| $1500\ m\ <= EL <=\ 2500$ m (mid elevation) | 17 | 31 | 0.95 | 3.81 | 95.21 |
| $EL > 2500$ m (high elevation) | 19 | 13 | 0.97 | 3.94 | 89.34 |
| Aridity Index<1 (Energy Limited) | 118 | 208 | 0.90 | 3.68 | 89.95 |
| Aridity Index>1 (Water Limited) | 52 | 135 | 0.87 | 3.68 | 93.48 |
| Snowy | 148 | 212 | 0.90 | 3.72 | 89.31 |
| Non-Snowy | 22 | 131 | 0.88 | 3.61 | 95.78 |
| $Max\ SWE \leq$ 100mm | 42 | 239 | 0.87 | 3.60 | 95.23 |
| 100mm $<\ Max\ SWE\ \leq$ 200mm | 22 | 37 | 0.87 | 3.64 | 89.14 |
| 200mm $<\ Max\ SWE\ \leq$ 400mm | 48 | 32 | 0.91 | 3.76 | 83.93 |
| 400mm $<\ Max\ SWE\ \leq$ 600mm | 13 | 14 | 0.94 | 3.85 | 91.70 |
| $Max\ SWE$ > 600mm | 45 | 21 | 0.95 | 3.89 | 84.79 |
| Forest | 109 | 173 | 0.90 | 3.72 | 90.62 |
| Open | 61 | 170 | 0.87 | 3.64 | 91.99 |
| Climate Zone: Arid | 8 | 27 | 0.86 | 3.71 | 97.31 |
| Climate Zone: Temperate | 41 | 201 | 0.88 | 3.63 | 95.40 |
| Climate Zone: Cold | 119 | 115 | 0.90 | 3.73 | 86.19 |
| Climate Zone: Polar | 2 | 0 | 0.97 | 4.00 | 72.00 |

**Table S10 | KGE-based architecture selection across the full candidate space with a maximum of 5 states.** For each basin grouping, the table reports the number of catchments assigned to each model family, median testing-period $KGE_{ss}$, mean selected state dimensionality and mean number of trainable parameters. The candidate space includes all eligible MCP-based and LSTM architectures containing at most 5 states.

| Regional Categorization | KGE-based model selection | | | | |
|---|---|---|---|---|---|
| | MCP-Based Net | LSTM Net | $KGE_{ss}^{50\%}$ | Ave State no | Ave Par no |
| CONUS | 163 | 350 | 0.89 | 4.47 | 123.70 |
| Eastern CONUS | 91 | 235 | 0.88 | 4.41 | 123.46 |
| Western CONUS | 72 | 115 | 0.93 | 4.58 | 124.12 |
| AM | 46 | 57 | 0.88 | 4.53 | 120.60 |
| RM | 29 | 25 | 0.97 | 4.83 | 126.69 |
| CRB | 11 | 23 | 0.95 | 4.79 | 137.12 |
| SN | 3 | 7 | 0.92 | 4.70 | 133.60 |
| CC | 20 | 13 | 0.92 | 4.52 | 110.15 |
| $EL < 1500$ m (low elevation) | 129 | 304 | 0.88 | 4.41 | 122.32 |
| $1500\ m\ <= EL <=\ 2500$ m (mid elevation) | 16 | 32 | 0.95 | 4.73 | 134.96 |
| $EL > 2500$ m (high elevation) | 18 | 14 | 0.97 | 4.88 | 125.44 |
| Aridity Index<1 (Energy Limited) | 112 | 214 | 0.90 | 4.45 | 121.75 |
| Aridity Index>1 (Water Limited) | 51 | 136 | 0.87 | 4.50 | 127.10 |
| Snowy | 139 | 221 | 0.90 | 4.56 | 123.91 |
| Non-Snowy | 24 | 129 | 0.87 | 4.25 | 123.20 |
| $Max\ SWE \leq$ 100mm | 44 | 237 | 0.87 | 4.36 | 127.03 |
| 100mm $< Max\ SWE \leq$ 200mm | 14 | 45 | 0.87 | 4.53 | 130.92 |
| 200mm $< Max\ SWE \leq$ 400mm | 50 | 30 | 0.92 | 4.59 | 112.96 |
| 400mm $< Max\ SWE \leq$ 600mm | 10 | 17 | 0.94 | 4.63 | 130.59 |
| $Max\ SWE >$ 600mm | 45 | 21 | 0.96 | 4.70 | 113.24 |
| Forest | 107 | 175 | 0.91 | 4.51 | 122.54 |
| Open | 56 | 175 | 0.87 | 4.42 | 125.11 |
| Climate Zone: Arid | 8 | 27 | 0.87 | 4.69 | 137.60 |
| Climate Zone: Temperate | 43 | 199 | 0.88 | 4.32 | 124.80 |
| Climate Zone: Cold | 110 | 124 | 0.91 | 4.59 | 120.68 |
| Climate Zone: Polar | 2 | 0 | 0.98 | 5.00 | 100.00 |

**Table S11 |** Hydrologic diagnostic errors at the 5th performance percentile across architecture-selection experiments and state dimensionalities.

| Experiment | State no. | 5th Performance Percentile | | | | | |
|---|---|---|---|---|---|---|---|
| | | FDC_NRMSE_MEANQ | FDC_LOG_RMSE | NABSERR_P95_MEANQ | NABSERR_P05_MEANQ | HIGH10_NRMSE_MEANQ | LOW10_NRMSE_MEANQ |
| ML (ALL) | 1 | 0.78 | 6.98 | 1.65 | 0.22 | 13.81 | 1.15 |
| ML (ALL) | 2 | 0.58 | 8.01 | 1.30 | 0.16 | 12.62 | 0.57 |
| ML (ALL) | 3 | 0.54 | 7.61 | 1.16 | 0.16 | 12.30 | 0.55 |
| ML (ALL) | 4 | 0.54 | 7.48 | 1.02 | 0.15 | 12.29 | 0.53 |
| ML (ALL) | 5 | 0.48 | 7.18 | 0.94 | 0.14 | 12.37 | 0.55 |
| ML (w.o LSTM) | 1 | 0.96 | 5.62 | 2.19 | 0.23 | 16.18 | 1.29 |
| ML (w.o LSTM) | 2 | 0.72 | 5.13 | 1.63 | 0.18 | 13.92 | 0.97 |
| ML (w.o LSTM) | 3 | 0.70 | 4.76 | 1.60 | 0.17 | 14.42 | 0.77 |
| ML (w.o LSTM) | 4 | 0.71 | 4.69 | 1.43 | 0.18 | 14.47 | 0.85 |
| ML (w.o LSTM) | 5 | 0.70 | 4.73 | 1.38 | 0.17 | 15.37 | 0.73 |
| ML (w.o LSTM & w.o sharing) | 1 | 0.96 | 5.62 | 2.19 | 0.23 | 16.18 | 1.29 |
| ML (w.o LSTM & w.o sharing) | 2 | 0.87 | 5.15 | 1.96 | 0.21 | 14.42 | 1.06 |
| ML (w.o LSTM & w.o sharing) | 3 | 0.82 | 5.21 | 1.77 | 0.19 | 14.81 | 1.16 |
| ML (w.o LSTM & w.o sharing) | 4 | 0.81 | 5.04 | 1.77 | 0.19 | 14.81 | 1.18 |
| ML (w.o LSTM & w.o sharing) | 5 | 0.85 | 4.85 | 1.80 | 0.18 | 14.80 | 1.18 |
| ML (LSTM only) | 1 | 0.93 | 8.67 | 1.72 | 0.21 | 12.77 | 1.29 |
| ML (LSTM only) | 2 | 0.55 | 8.12 | 1.31 | 0.18 | 13.04 | 0.60 |
| ML (LSTM only) | 3 | 0.54 | 7.87 | 1.16 | 0.17 | 12.76 | 0.56 |
| ML (LSTM only) | 4 | 0.53 | 7.74 | 1.02 | 0.15 | 12.34 | 0.51 |
| ML (LSTM only) | 5 | 0.45 | 7.62 | 0.94 | 0.14 | 12.29 | 0.55 |
| KGE (ALL) | 1 | 0.70 | 6.65 | 1.40 | 0.22 | 14.23 | 1.02 |
| KGE (ALL) | 2 | 0.55 | 7.56 | 1.08 | 0.16 | 13.19 | 0.57 |
| KGE (ALL) | 3 | 0.55 | 7.57 | 1.10 | 0.16 | 12.33 | 0.54 |
| KGE (ALL) | 4 | 0.55 | 7.77 | 0.94 | 0.16 | 12.33 | 0.51 |
| KGE (ALL) | 5 | 0.45 | 7.59 | 0.91 | 0.15 | 12.46 | 0.53 |
| KGE (w.o LSTM) | 1 | 0.77 | 5.59 | 1.49 | 0.23 | 16.73 | 1.10 |
| KGE (w.o LSTM) | 2 | 0.67 | 4.65 | 1.31 | 0.18 | 14.24 | 0.75 |
| KGE (w.o LSTM) | 3 | 0.70 | 4.39 | 1.44 | 0.18 | 15.45 | 0.74 |
| KGE (w.o LSTM) | 4 | 0.70 | 4.41 | 1.41 | 0.18 | 15.39 | 0.72 |
| KGE (w.o LSTM) | 5 | 0.63 | 4.62 | 1.19 | 0.18 | 15.83 | 0.69 |
| KGE (w.o LSTM & w.o sharing) | 1 | 0.77 | 5.59 | 1.49 | 0.23 | 16.73 | 1.10 |
| KGE (w.o LSTM & w.o sharing) | 2 | 0.78 | 5.11 | 1.35 | 0.21 | 14.57 | 0.93 |
| KGE (w.o LSTM & w.o sharing) | 3 | 0.79 | 5.28 | 1.33 | 0.19 | 14.81 | 0.96 |
| KGE (w.o LSTM & w.o sharing) | 4 | 0.79 | 5.06 | 1.31 | 0.18 | 14.80 | 0.96 |
| KGE (w.o LSTM & w.o sharing) | 5 | 0.79 | 5.04 | 1.34 | 0.18 | 14.86 | 0.96 |
| KGE (LSTM only) | 1 | 0.93 | 8.67 | 1.72 | 0.21 | 12.77 | 1.29 |
| KGE (LSTM only) | 2 | 0.53 | 8.34 | 1.08 | 0.17 | 13.08 | 0.58 |
| KGE (LSTM only) | 3 | 0.51 | 7.66 | 1.09 | 0.16 | 12.39 | 0.54 |
| KGE (LSTM only) | 4 | 0.55 | 7.84 | 0.96 | 0.15 | 12.34 | 0.48 |
| KGE (LSTM only) | 5 | 0.47 | 7.70 | 0.93 | 0.14 | 12.56 | 0.48 |

**Table S12 |** Hydrologic diagnostic errors at the 25th performance percentile across architecture-selection experiments and state dimensionalities.

| Experiment | State no. | 25th Performance Percentile | | | | | |
|---|---|---|---|---|---|---|---|
| | | FDC_NRMSE_MEANQ | FDC_LOG_RMSE | NABSERR_P95_MEANQ | NABSERR_P05_MEANQ | HIGH10_NRMSE_MEANQ | LOW10_NRMSE_MEANQ |
| ML (ALL) | 1 | 0.31 | 5.06 | 0.54 | 0.11 | 5.56 | 0.39 |
| ML (ALL) | 2 | 0.24 | 4.81 | 0.43 | 0.09 | 4.90 | 0.22 |
| ML (ALL) | 3 | 0.22 | 4.66 | 0.37 | 0.08 | 4.87 | 0.21 |
| ML (ALL) | 4 | 0.21 | 4.86 | 0.36 | 0.08 | 4.73 | 0.21 |
| ML (ALL) | 5 | 0.21 | 4.46 | 0.34 | 0.08 | 4.62 | 0.22 |
| ML (w.o LSTM) | 1 | 0.36 | 2.69 | 0.65 | 0.12 | 6.23 | 0.42 |
| ML (w.o LSTM) | 2 | 0.29 | 1.73 | 0.50 | 0.10 | 5.51 | 0.29 |
| ML (w.o LSTM) | 3 | 0.26 | 1.73 | 0.47 | 0.09 | 5.28 | 0.26 |
| ML (w.o LSTM) | 4 | 0.26 | 1.41 | 0.45 | 0.08 | 5.36 | 0.26 |
| ML (w.o LSTM) | 5 | 0.25 | 1.53 | 0.41 | 0.07 | 5.36 | 0.26 |
| ML (w.o LSTM & w.o sharing) | 1 | 0.36 | 2.69 | 0.65 | 0.12 | 6.23 | 0.42 |
| ML (w.o LSTM & w.o sharing) | 2 | 0.32 | 2.47 | 0.60 | 0.11 | 5.86 | 0.35 |
| ML (w.o LSTM & w.o sharing) | 3 | 0.31 | 2.32 | 0.58 | 0.11 | 5.78 | 0.32 |
| ML (w.o LSTM & w.o sharing) | 4 | 0.30 | 2.27 | 0.56 | 0.11 | 5.63 | 0.31 |
| ML (w.o LSTM & w.o sharing) | 5 | 0.30 | 2.25 | 0.56 | 0.11 | 5.67 | 0.30 |
| ML (LSTM only) | 1 | 0.40 | 6.31 | 0.62 | 0.11 | 5.50 | 0.53 |
| ML (LSTM only) | 2 | 0.25 | 5.91 | 0.44 | 0.10 | 4.76 | 0.22 |
| ML (LSTM only) | 3 | 0.24 | 5.66 | 0.36 | 0.09 | 4.87 | 0.21 |
| ML (LSTM only) | 4 | 0.22 | 5.56 | 0.36 | 0.08 | 4.72 | 0.20 |
| ML (LSTM only) | 5 | 0.21 | 5.44 | 0.34 | 0.08 | 4.68 | 0.21 |
| KGE (ALL) | 1 | 0.30 | 4.55 | 0.52 | 0.11 | 5.54 | 0.36 |
| KGE (ALL) | 2 | 0.24 | 4.93 | 0.42 | 0.09 | 4.93 | 0.22 |
| KGE (ALL) | 3 | 0.22 | 5.08 | 0.37 | 0.08 | 4.89 | 0.20 |
| KGE (ALL) | 4 | 0.21 | 5.13 | 0.36 | 0.08 | 4.74 | 0.20 |
| KGE (ALL) | 5 | 0.21 | 4.98 | 0.34 | 0.08 | 4.69 | 0.21 |
| KGE (w.o LSTM) | 1 | 0.35 | 2.58 | 0.62 | 0.12 | 6.24 | 0.39 |
| KGE (w.o LSTM) | 2 | 0.28 | 1.96 | 0.50 | 0.10 | 5.54 | 0.29 |
| KGE (w.o LSTM) | 3 | 0.26 | 1.73 | 0.46 | 0.09 | 5.51 | 0.26 |
| KGE (w.o LSTM) | 4 | 0.26 | 1.66 | 0.41 | 0.08 | 5.25 | 0.25 |
| KGE (w.o LSTM) | 5 | 0.24 | 1.58 | 0.39 | 0.08 | 5.28 | 0.25 |
| KGE (w.o LSTM & w.o sharing) | 1 | 0.35 | 2.58 | 0.62 | 0.12 | 6.24 | 0.39 |
| KGE (w.o LSTM & w.o sharing) | 2 | 0.31 | 2.46 | 0.59 | 0.11 | 5.87 | 0.32 |
| KGE (w.o LSTM & w.o sharing) | 3 | 0.30 | 2.31 | 0.56 | 0.11 | 5.78 | 0.32 |
| KGE (w.o LSTM & w.o sharing) | 4 | 0.30 | 2.29 | 0.54 | 0.11 | 5.71 | 0.31 |
| KGE (w.o LSTM & w.o sharing) | 5 | 0.30 | 2.24 | 0.53 | 0.11 | 5.71 | 0.31 |
| KGE (LSTM only) | 1 | 0.40 | 6.31 | 0.62 | 0.11 | 5.50 | 0.53 |
| KGE (LSTM only) | 2 | 0.25 | 6.02 | 0.43 | 0.10 | 4.76 | 0.22 |
| KGE (LSTM only) | 3 | 0.23 | 5.63 | 0.36 | 0.09 | 4.79 | 0.20 |
| KGE (LSTM only) | 4 | 0.21 | 5.55 | 0.36 | 0.08 | 4.74 | 0.20 |
| KGE (LSTM only) | 5 | 0.21 | 5.34 | 0.34 | 0.08 | 4.69 | 0.21 |

**Table S13 |** Hydrologic diagnostic errors at the median performance across architecture-selection experiments and state dimensionalities.

| Experiment | State no. | 50th Performance Percentile | | | | | |
|---|---|---|---|---|---|---|---|
| | | FDC_NRMSE_MEANQ | FDC_LOG_RMSE | NABSERR_P95_MEANQ | NABSERR_P05_MEANQ | HIGH10_NRMSE_MEANQ | LOW10_NRMSE_MEANQ |
| ML (ALL) | 1 | 0.20 | 2.90 | 0.27 | 0.06 | 3.10 | 0.20 |
| ML (ALL) | 2 | 0.14 | 1.65 | 0.20 | 0.05 | 2.56 | 0.14 |
| ML (ALL) | 3 | 0.13 | 1.63 | 0.19 | 0.04 | 2.51 | 0.14 |
| ML (ALL) | 4 | 0.12 | 1.41 | 0.14 | 0.04 | 2.53 | 0.13 |
| ML (ALL) | 5 | 0.12 | 1.32 | 0.15 | 0.04 | 2.48 | 0.13 |
| ML (w.o LSTM) | 1 | 0.23 | 1.37 | 0.33 | 0.06 | 3.22 | 0.24 |
| ML (w.o LSTM) | 2 | 0.16 | 0.80 | 0.22 | 0.05 | 2.73 | 0.16 |
| ML (w.o LSTM) | 3 | 0.14 | 0.70 | 0.21 | 0.04 | 2.68 | 0.15 |
| ML (w.o LSTM) | 4 | 0.13 | 0.57 | 0.18 | 0.04 | 2.64 | 0.14 |
| ML (w.o LSTM) | 5 | 0.13 | 0.57 | 0.17 | 0.04 | 2.64 | 0.14 |
| ML (w.o LSTM & w.o sharing) | 1 | 0.23 | 1.37 | 0.33 | 0.06 | 3.22 | 0.24 |
| ML (w.o LSTM & w.o sharing) | 2 | 0.20 | 1.18 | 0.30 | 0.06 | 3.04 | 0.18 |
| ML (w.o LSTM & w.o sharing) | 3 | 0.19 | 1.12 | 0.29 | 0.05 | 3.04 | 0.17 |
| ML (w.o LSTM & w.o sharing) | 4 | 0.19 | 1.16 | 0.27 | 0.05 | 2.97 | 0.17 |
| ML (w.o LSTM & w.o sharing) | 5 | 0.18 | 1.06 | 0.27 | 0.05 | 2.95 | 0.17 |
| ML (LSTM only) | 1 | 0.22 | 4.85 | 0.28 | 0.06 | 3.17 | 0.26 |
| ML (LSTM only) | 2 | 0.15 | 4.21 | 0.20 | 0.05 | 2.63 | 0.14 |
| ML (LSTM only) | 3 | 0.14 | 3.76 | 0.19 | 0.04 | 2.53 | 0.13 |
| ML (LSTM only) | 4 | 0.13 | 3.52 | 0.16 | 0.04 | 2.55 | 0.12 |
| ML (LSTM only) | 5 | 0.13 | 3.43 | 0.17 | 0.04 | 2.57 | 0.13 |
| KGE (ALL) | 1 | 0.20 | 2.24 | 0.27 | 0.06 | 3.13 | 0.18 |
| KGE (ALL) | 2 | 0.14 | 2.00 | 0.20 | 0.05 | 2.57 | 0.13 |
| KGE (ALL) | 3 | 0.13 | 1.78 | 0.18 | 0.04 | 2.48 | 0.13 |
| KGE (ALL) | 4 | 0.13 | 1.71 | 0.15 | 0.04 | 2.50 | 0.13 |
| KGE (ALL) | 5 | 0.12 | 1.71 | 0.16 | 0.04 | 2.49 | 0.13 |
| KGE (w.o LSTM) | 1 | 0.23 | 1.33 | 0.33 | 0.06 | 3.22 | 0.22 |
| KGE (w.o LSTM) | 2 | 0.16 | 0.81 | 0.22 | 0.05 | 2.74 | 0.15 |
| KGE (w.o LSTM) | 3 | 0.14 | 0.70 | 0.20 | 0.04 | 2.62 | 0.14 |
| KGE (w.o LSTM) | 4 | 0.14 | 0.60 | 0.18 | 0.04 | 2.63 | 0.14 |
| KGE (w.o LSTM) | 5 | 0.14 | 0.58 | 0.17 | 0.04 | 2.58 | 0.14 |
| KGE (w.o LSTM & w.o sharing) | 1 | 0.23 | 1.33 | 0.33 | 0.06 | 3.22 | 0.22 |
| KGE (w.o LSTM & w.o sharing) | 2 | 0.19 | 1.17 | 0.28 | 0.06 | 3.00 | 0.18 |
| KGE (w.o LSTM & w.o sharing) | 3 | 0.18 | 1.12 | 0.26 | 0.05 | 2.99 | 0.17 |
| KGE (w.o LSTM & w.o sharing) | 4 | 0.18 | 1.14 | 0.26 | 0.05 | 2.99 | 0.17 |
| KGE (w.o LSTM & w.o sharing) | 5 | 0.18 | 1.10 | 0.26 | 0.05 | 2.99 | 0.17 |
| KGE (LSTM only) | 1 | 0.22 | 4.85 | 0.28 | 0.06 | 3.17 | 0.26 |
| KGE (LSTM only) | 2 | 0.15 | 4.21 | 0.20 | 0.05 | 2.61 | 0.13 |
| KGE (LSTM only) | 3 | 0.14 | 3.78 | 0.18 | 0.04 | 2.53 | 0.13 |
| KGE (LSTM only) | 4 | 0.13 | 3.54 | 0.16 | 0.04 | 2.55 | 0.13 |
| KGE (LSTM only) | 5 | 0.12 | 3.39 | 0.17 | 0.04 | 2.54 | 0.13 |

**Table S14 |** Hydrologic diagnostic errors at the 75th performance percentile across architecture-selection experiments and state dimensionalities.

| Experiment | State no. | 75th Performance Percentile | | | | | |
|---|---|---|---|---|---|---|---|
| | | FDC_NRMSE_MEANQ | FDC_LOG_RMSE | NABSERR_P95_MEANQ | NABSERR_P05_MEANQ | HIGH10_NRMSE_MEANQ | LOW10_NRMSE_MEANQ |
| ML (ALL) | 1 | 0.13 | 1.05 | 0.12 | 0.01 | 2.04 | 0.12 |
| ML (ALL) | 2 | 0.09 | 0.40 | 0.08 | 0.01 | 1.69 | 0.09 |
| ML (ALL) | 3 | 0.08 | 0.36 | 0.08 | 0.01 | 1.60 | 0.08 |
| ML (ALL) | 4 | 0.08 | 0.29 | 0.06 | 0.01 | 1.55 | 0.08 |
| ML (ALL) | 5 | 0.08 | 0.29 | 0.06 | 0.01 | 1.54 | 0.08 |
| ML (w.o LSTM) | 1 | 0.14 | 0.62 | 0.16 | 0.01 | 2.13 | 0.12 |
| ML (w.o LSTM) | 2 | 0.10 | 0.35 | 0.09 | 0.01 | 1.76 | 0.09 |
| ML (w.o LSTM) | 3 | 0.09 | 0.31 | 0.08 | 0.01 | 1.69 | 0.08 |
| ML (w.o LSTM) | 4 | 0.08 | 0.23 | 0.07 | 0.01 | 1.62 | 0.08 |
| ML (w.o LSTM) | 5 | 0.08 | 0.22 | 0.06 | 0.01 | 1.60 | 0.08 |
| ML (w.o LSTM & w.o sharing) | 1 | 0.14 | 0.62 | 0.16 | 0.01 | 2.13 | 0.12 |
| ML (w.o LSTM & w.o sharing) | 2 | 0.13 | 0.53 | 0.13 | 0.01 | 1.95 | 0.11 |
| ML (w.o LSTM & w.o sharing) | 3 | 0.12 | 0.49 | 0.12 | 0.01 | 1.96 | 0.10 |
| ML (w.o LSTM & w.o sharing) | 4 | 0.12 | 0.48 | 0.11 | 0.01 | 1.93 | 0.10 |
| ML (w.o LSTM & w.o sharing) | 5 | 0.12 | 0.47 | 0.12 | 0.01 | 1.92 | 0.10 |
| ML (LSTM only) | 1 | 0.14 | 2.83 | 0.12 | 0.01 | 2.13 | 0.14 |
| ML (LSTM only) | 2 | 0.10 | 1.61 | 0.08 | 0.01 | 1.74 | 0.08 |
| ML (LSTM only) | 3 | 0.09 | 1.27 | 0.08 | 0.01 | 1.68 | 0.08 |
| ML (LSTM only) | 4 | 0.09 | 1.13 | 0.07 | 0.01 | 1.69 | 0.08 |
| ML (LSTM only) | 5 | 0.08 | 1.15 | 0.07 | 0.01 | 1.64 | 0.08 |
| KGE (ALL) | 1 | 0.13 | 0.88 | 0.12 | 0.01 | 2.04 | 0.11 |
| KGE (ALL) | 2 | 0.09 | 0.45 | 0.08 | 0.01 | 1.68 | 0.08 |
| KGE (ALL) | 3 | 0.08 | 0.36 | 0.08 | 0.01 | 1.60 | 0.09 |
| KGE (ALL) | 4 | 0.08 | 0.29 | 0.06 | 0.01 | 1.57 | 0.08 |
| KGE (ALL) | 5 | 0.08 | 0.29 | 0.06 | 0.01 | 1.55 | 0.08 |
| KGE (w.o LSTM) | 1 | 0.14 | 0.62 | 0.15 | 0.01 | 2.12 | 0.11 |
| KGE (w.o LSTM) | 2 | 0.10 | 0.35 | 0.09 | 0.01 | 1.73 | 0.09 |
| KGE (w.o LSTM) | 3 | 0.09 | 0.32 | 0.08 | 0.01 | 1.64 | 0.08 |
| KGE (w.o LSTM) | 4 | 0.08 | 0.25 | 0.08 | 0.01 | 1.62 | 0.08 |
| KGE (w.o LSTM) | 5 | 0.08 | 0.24 | 0.08 | 0.01 | 1.59 | 0.08 |
| KGE (w.o LSTM & w.o sharing) | 1 | 0.14 | 0.62 | 0.15 | 0.01 | 2.12 | 0.11 |
| KGE (w.o LSTM & w.o sharing) | 2 | 0.13 | 0.54 | 0.12 | 0.01 | 1.95 | 0.10 |
| KGE (w.o LSTM & w.o sharing) | 3 | 0.12 | 0.49 | 0.11 | 0.01 | 1.96 | 0.10 |
| KGE (w.o LSTM & w.o sharing) | 4 | 0.12 | 0.48 | 0.11 | 0.01 | 1.94 | 0.10 |
| KGE (w.o LSTM & w.o sharing) | 5 | 0.12 | 0.46 | 0.11 | 0.01 | 1.93 | 0.10 |
| KGE (LSTM only) | 1 | 0.14 | 2.83 | 0.12 | 0.01 | 2.13 | 0.14 |
| KGE (LSTM only) | 2 | 0.10 | 1.55 | 0.08 | 0.01 | 1.74 | 0.08 |
| KGE (LSTM only) | 3 | 0.09 | 1.23 | 0.08 | 0.01 | 1.68 | 0.08 |
| KGE (LSTM only) | 4 | 0.09 | 1.05 | 0.06 | 0.01 | 1.69 | 0.08 |
| KGE (LSTM only) | 5 | 0.08 | 1.13 | 0.07 | 0.01 | 1.64 | 0.08 |

**Table S15 |** Hydrologic diagnostic errors at the 95th performance percentile across architecture-selection experiments and state dimensionalities.

| Experiment | State no. | 95th Performance Percentile | | | | | |
|---|---|---|---|---|---|---|---|
| | | FDC_NRMSE_MEANQ | FDC_LOG_RMSE | NABSERR_P95_MEANQ | NABSERR_P05_MEANQ | HIGH10_NRMSE_MEANQ | LOW10_NRMSE_MEANQ |
| ML (ALL) | 1 | 0.07 | 0.12 | 0.02 | 0.00 | 1.18 | 0.03 |
| ML (ALL) | 2 | 0.05 | 0.08 | 0.02 | 0.00 | 0.89 | 0.02 |
| ML (ALL) | 3 | 0.05 | 0.08 | 0.02 | 0.00 | 0.82 | 0.03 |
| ML (ALL) | 4 | 0.05 | 0.07 | 0.01 | 0.00 | 0.75 | 0.03 |
| ML (ALL) | 5 | 0.04 | 0.07 | 0.01 | 0.00 | 0.74 | 0.03 |
| ML (w.o LSTM) | 1 | 0.08 | 0.16 | 0.04 | 0.00 | 1.32 | 0.03 |
| ML (w.o LSTM) | 2 | 0.06 | 0.10 | 0.02 | 0.00 | 0.92 | 0.03 |
| ML (w.o LSTM) | 3 | 0.05 | 0.08 | 0.02 | 0.00 | 0.85 | 0.02 |
| ML (w.o LSTM) | 4 | 0.05 | 0.07 | 0.02 | 0.00 | 0.86 | 0.03 |
| ML (w.o LSTM) | 5 | 0.04 | 0.06 | 0.01 | 0.00 | 0.78 | 0.03 |
| ML (w.o LSTM & w.o sharing) | 1 | 0.08 | 0.16 | 0.04 | 0.00 | 1.32 | 0.03 |
| ML (w.o LSTM & w.o sharing) | 2 | 0.07 | 0.15 | 0.03 | 0.00 | 1.19 | 0.03 |
| ML (w.o LSTM & w.o sharing) | 3 | 0.07 | 0.12 | 0.02 | 0.00 | 1.14 | 0.03 |
| ML (w.o LSTM & w.o sharing) | 4 | 0.07 | 0.13 | 0.02 | 0.00 | 1.18 | 0.03 |
| ML (w.o LSTM & w.o sharing) | 5 | 0.07 | 0.12 | 0.02 | 0.00 | 1.17 | 0.03 |
| ML (LSTM only) | 1 | 0.08 | 0.12 | 0.02 | 0.00 | 1.19 | 0.05 |
| ML (LSTM only) | 2 | 0.06 | 0.08 | 0.02 | 0.00 | 0.97 | 0.03 |
| ML (LSTM only) | 3 | 0.05 | 0.09 | 0.02 | 0.00 | 0.88 | 0.03 |
| ML (LSTM only) | 4 | 0.05 | 0.07 | 0.02 | 0.00 | 0.81 | 0.04 |
| ML (LSTM only) | 5 | 0.05 | 0.06 | 0.02 | 0.00 | 0.78 | 0.03 |
| KGE (ALL) | 1 | 0.07 | 0.12 | 0.02 | 0.00 | 1.19 | 0.03 |
| KGE (ALL) | 2 | 0.05 | 0.08 | 0.02 | 0.00 | 0.90 | 0.02 |
| KGE (ALL) | 3 | 0.05 | 0.08 | 0.02 | 0.00 | 0.82 | 0.03 |
| KGE (ALL) | 4 | 0.05 | 0.07 | 0.02 | 0.00 | 0.75 | 0.03 |
| KGE (ALL) | 5 | 0.04 | 0.06 | 0.01 | 0.00 | 0.73 | 0.04 |
| KGE (w.o LSTM) | 1 | 0.08 | 0.16 | 0.04 | 0.00 | 1.34 | 0.02 |
| KGE (w.o LSTM) | 2 | 0.05 | 0.10 | 0.02 | 0.00 | 0.92 | 0.03 |
| KGE (w.o LSTM) | 3 | 0.05 | 0.09 | 0.02 | 0.00 | 0.85 | 0.03 |
| KGE (w.o LSTM) | 4 | 0.04 | 0.07 | 0.01 | 0.00 | 0.86 | 0.03 |
| KGE (w.o LSTM) | 5 | 0.04 | 0.06 | 0.01 | 0.00 | 0.80 | 0.03 |
| KGE (w.o LSTM & w.o sharing) | 1 | 0.08 | 0.16 | 0.04 | 0.00 | 1.34 | 0.02 |
| KGE (w.o LSTM & w.o sharing) | 2 | 0.07 | 0.13 | 0.03 | 0.00 | 1.21 | 0.02 |
| KGE (w.o LSTM & w.o sharing) | 3 | 0.07 | 0.11 | 0.02 | 0.00 | 1.19 | 0.02 |
| KGE (w.o LSTM & w.o sharing) | 4 | 0.07 | 0.12 | 0.02 | 0.00 | 1.19 | 0.02 |
| KGE (w.o LSTM & w.o sharing) | 5 | 0.07 | 0.12 | 0.03 | 0.00 | 1.19 | 0.02 |
| KGE (LSTM only) | 1 | 0.08 | 0.12 | 0.02 | 0.00 | 1.19 | 0.05 |
| KGE (LSTM only) | 2 | 0.06 | 0.08 | 0.02 | 0.00 | 0.97 | 0.03 |
| KGE (LSTM only) | 3 | 0.05 | 0.09 | 0.02 | 0.00 | 0.88 | 0.04 |
| KGE (LSTM only) | 4 | 0.05 | 0.07 | 0.02 | 0.00 | 0.81 | 0.04 |
| KGE (LSTM only) | 5 | 0.05 | 0.06 | 0.02 | 0.00 | 0.78 | 0.04 |

**Table S16 | Likelihood-based architecture selection within the sharing-enabled MCP candidate space with a maximum of one state.** For each basin grouping, the table reports the number of catchments assigned to each MCP-based model family, median testing-period $KGE_{ss}$, mean selected state dimensionality and mean number of trainable parameters. The candidate space excludes LSTMs while retaining MCP architectures with cross-state information sharing.

| Regional Categorization | Likelihood-based model selection | | | | | | |
|---|---|---|---|---|---|---|---|
| | HMCP | SNOWMCP | HMCPMR | HYDROMCP-Net | $KGE_{ss}^{50\%}$ | Ave State no | Ave Par no |
| CONUS | 79 | 273 | 161 | 0 | 0.80 | 1.00 | 10.54 |
| Eastern CONUS | 58 | 145 | 123 | 0 | 0.79 | 1.00 | 10.47 |
| Western CONUS | 21 | 128 | 38 | 0 | 0.86 | 1.00 | 10.66 |
| AM | 12 | 58 | 33 | 0 | 0.79 | 1.00 | 10.65 |
| RM | 1 | 52 | 1 | 0 | 0.89 | 1.00 | 10.94 |
| CRB | 4 | 23 | 7 | 0 | 0.83 | 1.00 | 10.65 |
| SN | 0 | 9 | 1 | 0 | 0.85 | 1.00 | 11.00 |
| CC | 1 | 26 | 6 | 0 | 0.88 | 1.00 | 10.91 |
| $EL < 1500$ m (low elevation) | 75 | 208 | 150 | 0 | 0.80 | 1.00 | 10.48 |
| $1500\ m <= EL <= 2500$ m (mid elevation) | 3 | 35 | 10 | 0 | 0.85 | 1.00 | 10.81 |
| $EL > 2500$ m (high elevation) | 1 | 30 | 1 | 0 | 0.90 | 1.00 | 10.91 |
| Aridity Index<1 (Energy Limited) | 42 | 174 | 110 | 0 | 0.82 | 1.00 | 10.61 |
| Aridity Index>1 (Water Limited) | 37 | 99 | 51 | 0 | 0.78 | 1.00 | 10.41 |
| Snowy | 43 | 233 | 84 | 0 | 0.80 | 1.00 | 10.64 |
| Non-Snowy | 36 | 40 | 77 | 0 | 0.82 | 1.00 | 10.29 |
| $Max\ SWE \leq$ 100mm | 64 | 94 | 123 | 0 | 0.80 | 1.00 | 10.32 |
| 100mm $< Max\ SWE \leq$ 200mm | 6 | 36 | 17 | 0 | 0.76 | 1.00 | 10.69 |
| 200mm $< Max\ SWE \leq$ 400mm | 6 | 59 | 15 | 0 | 0.79 | 1.00 | 10.78 |
| 400mm $< Max\ SWE \leq$ 600mm | 1 | 26 | 0 | 0 | 0.86 | 1.00 | 10.89 |
| $Max\ SWE >$ 600mm | 2 | 58 | 6 | 0 | 0.88 | 1.00 | 10.91 |
| Forest | 32 | 152 | 98 | 0 | 0.82 | 1.00 | 10.66 |
| Open | 47 | 121 | 63 | 0 | 0.78 | 1.00 | 10.39 |
| Climate Zone: Arid | 9 | 15 | 11 | 0 | 0.67 | 1.00 | 10.23 |
| Climate Zone: Temperate | 46 | 91 | 105 | 0 | 0.82 | 1.00 | 10.43 |
| Climate Zone: Cold | 24 | 165 | 45 | 0 | 0.80 | 1.00 | 10.69 |
| Climate Zone: Polar | 0 | 2 | 0 | 0 | 0.93 | 1.00 | 11.00 |

**Table S17 | Likelihood-based architecture selection within the sharing-enabled MCP candidate space with a maximum of 3 states.** For each basin grouping, the table reports the number of catchments assigned to each MCP-based model family, median testing-period $KGE_{ss}$, mean selected state dimensionality and mean number of trainable parameters. The candidate space excludes LSTMs while retaining MCP architectures with cross-state information sharing.

| Regional Categorization | Likelihood-based model selection | | | | | | |
|---|---|---|---|---|---|---|---|
| | HMCP | SNOWMCP | HMCPMR | HYDROMCP-Net | $KGE_{ss}^{50\%}$ | Ave State no | Ave Par no |
| CONUS | 93 | 250 | 111 | 59 | 0.88 | 2.75 | 41.18 |
| Eastern CONUS | 79 | 128 | 79 | 40 | 0.87 | 2.71 | 39.56 |
| Western CONUS | 14 | 122 | 32 | 19 | 0.92 | 2.82 | 44.01 |
| AM | 10 | 74 | 10 | 9 | 0.88 | 2.73 | 41.86 |
| RM | 0 | 52 | 2 | 0 | 0.96 | 2.96 | 49.28 |
| CRB | 3 | 21 | 6 | 4 | 0.94 | 2.85 | 43.68 |
| SN | 0 | 7 | 1 | 2 | 0.91 | 2.70 | 43.00 |
| CC | 0 | 32 | 0 | 1 | 0.92 | 2.91 | 47.73 |
| $EL < 1500$ m (low elevation) | 90 | 183 | 104 | 56 | 0.88 | 2.71 | 39.90 |
| $1500\ m <= EL <= 2500$ m (mid elevation) | 3 | 35 | 7 | 3 | 0.94 | 2.92 | 46.79 |
| $EL > 2500$ m (high elevation) | 0 | 32 | 0 | 0 | 0.96 | 3.00 | 50.06 |
| Aridity Index<1 (Energy Limited) | 57 | 174 | 58 | 37 | 0.89 | 2.75 | 41.25 |
| Aridity Index>1 (Water Limited) | 36 | 76 | 53 | 22 | 0.85 | 2.75 | 41.07 |
| Snowy | 47 | 234 | 42 | 37 | 0.89 | 2.78 | 42.82 |
| Non-Snowy | 46 | 16 | 69 | 22 | 0.86 | 2.67 | 37.33 |
| $Max\ SWE \leq$ 100mm | 82 | 54 | 101 | 44 | 0.85 | 2.65 | 37.26 |
| 100mm < $Max\ SWE \leq$ 200mm | 8 | 40 | 4 | 7 | 0.84 | 2.78 | 42.39 |
| 200mm < $Max\ SWE \leq$ 400mm | 2 | 68 | 4 | 6 | 0.91 | 2.88 | 46.24 |
| 400mm < $Max\ SWE \leq$ 600mm | 0 | 24 | 2 | 1 | 0.93 | 2.89 | 47.22 |
| $Max\ SWE$ > 600mm | 1 | 64 | 0 | 1 | 0.95 | 2.94 | 48.21 |
| Forest | 37 | 161 | 54 | 30 | 0.90 | 2.78 | 42.47 |
| Open | 56 | 89 | 57 | 29 | 0.85 | 2.71 | 39.61 |
| Climate Zone: Arid | 2 | 14 | 12 | 7 | 0.81 | 2.71 | 40.54 |
| Climate Zone: Temperate | 66 | 60 | 77 | 39 | 0.87 | 2.65 | 37.63 |
| Climate Zone: Cold | 25 | 174 | 22 | 13 | 0.90 | 2.85 | 44.88 |
| Climate Zone: Polar | 0 | 2 | 0 | 0 | 0.98 | 3.00 | 49.50 |

**Table S18 | Likelihood-based architecture selection within the sharing-enabled MCP candidate space with a maximum of 4 states.** For each basin grouping, the table reports the number of catchments assigned to each MCP-based model family, median testing-period $KGE_{ss}$, mean selected state dimensionality and mean number of trainable parameters. The candidate space excludes LSTMs while retaining MCP architectures with cross-state information sharing.

| Regional Categorization | Likelihood-based model selection | | | | | | |
|---|---|---|---|---|---|---|---|
| | HMCP | SNOWMCP | HMCPMR | HYDROMCP-Net | $KGE_{ss}^{50\%}$ | Ave State no | Ave Par no |
| CONUS | 98 | 256 | 120 | 39 | 0.88 | 3.48 | 57.59 |
| Eastern CONUS | 78 | 136 | 87 | 25 | 0.87 | 3.41 | 55.00 |
| Western CONUS | 20 | 120 | 33 | 14 | 0.92 | 3.59 | 62.12 |
| AM | 12 | 78 | 10 | 3 | 0.88 | 3.53 | 59.58 |
| RM | 0 | 52 | 2 | 0 | 0.96 | 3.89 | 71.63 |
| CRB | 2 | 19 | 10 | 3 | 0.94 | 3.65 | 63.09 |
| SN | 0 | 7 | 1 | 2 | 0.91 | 3.60 | 63.40 |
| CC | 0 | 32 | 0 | 1 | 0.91 | 3.52 | 63.00 |
| $EL < 1500$ m (low elevation) | 95 | 192 | 109 | 37 | 0.87 | 3.41 | 55.27 |
| $1500\ m\ <= EL <=\ 2500$ m (mid elevation) | 3 | 32 | 11 | 2 | 0.94 | 3.81 | 68.58 |
| $EL > 2500$ m (high elevation) | 0 | 32 | 0 | 0 | 0.96 | 3.94 | 72.59 |
| Aridity Index<1 (Energy Limited) | 55 | 183 | 62 | 26 | 0.89 | 3.44 | 56.96 |
| Aridity Index>1 (Water Limited) | 43 | 73 | 58 | 13 | 0.85 | 3.53 | 58.70 |
| Snowy | 53 | 238 | 48 | 21 | 0.89 | 3.54 | 60.37 |
| Non-Snowy | 45 | 18 | 72 | 18 | 0.86 | 3.32 | 51.07 |
| $Max\ SWE \leq$ 100mm | 92 | 56 | 104 | 29 | 0.85 | 3.33 | 51.27 |
| 100mm < $Max\ SWE\ \leq$ 200mm | 6 | 42 | 8 | 3 | 0.86 | 3.51 | 60.31 |
| 200mm < $Max\ SWE\ \leq$ 400mm | 0 | 71 | 4 | 5 | 0.91 | 3.61 | 64.49 |
| 400mm < $Max\ SWE\ \leq$ 600mm | 0 | 22 | 4 | 1 | 0.94 | 3.74 | 67.74 |
| $Max\ SWE$ > 600mm | 0 | 65 | 0 | 1 | 0.95 | 3.80 | 69.58 |
| Forest | 45 | 163 | 51 | 23 | 0.90 | 3.48 | 58.49 |
| Open | 53 | 93 | 69 | 16 | 0.85 | 3.48 | 56.51 |
| Climate Zone: Arid | 4 | 11 | 15 | 5 | 0.84 | 3.40 | 56.20 |
| Climate Zone: Temperate | 73 | 61 | 80 | 28 | 0.87 | 3.31 | 51.11 |
| Climate Zone: Cold | 21 | 182 | 25 | 6 | 0.90 | 3.67 | 64.49 |
| Climate Zone: Polar | 0 | 2 | 0 | 0 | 0.98 | 3.50 | 60.00 |

**Table S19 | Likelihood-based architecture selection within the sharing-enabled MCP candidate space with a maximum of 5 states.** For each basin grouping, the table reports the number of catchments assigned to each MCP-based model family, median testing-period $KGE_{ss}$, mean selected state dimensionality and mean number of trainable parameters. The candidate space excludes LSTMs while retaining MCP architectures with cross-state information sharing.

| Regional Categorization | Likelihood-based model selection | | | | | | |
|---|---|---|---|---|---|---|---|
| | HMCP | SNOWMCP | HMCPMR | HYDROMCP-Net | $KGE_{ss}^{50\%}$ | Ave State no | Ave Par no |
| CONUS | 112 | 252 | 125 | 24 | 0.89 | 4.19 | 75.96 |
| Eastern CONUS | 85 | 132 | 91 | 18 | 0.87 | 4.09 | 72.40 |
| Western CONUS | 27 | 120 | 34 | 6 | 0.92 | 4.37 | 82.17 |
| AM | 11 | 74 | 16 | 2 | 0.87 | 4.30 | 80.15 |
| RM | 0 | 53 | 1 | 0 | 0.97 | 4.70 | 95.28 |
| CRB | 2 | 20 | 10 | 2 | 0.94 | 4.41 | 84.18 |
| SN | 1 | 7 | 1 | 1 | 0.91 | 4.60 | 88.50 |
| CC | 0 | 33 | 0 | 0 | 0.92 | 4.30 | 84.91 |
| $EL < 1500$ m (low elevation) | 108 | 187 | 114 | 24 | 0.88 | 4.10 | 72.69 |
| $1500\ m <= EL <= 2500$ m (mid elevation) | 4 | 33 | 11 | 0 | 0.94 | 4.63 | 91.08 |
| $EL > 2500$ m (high elevation) | 0 | 32 | 0 | 0 | 0.97 | 4.81 | 97.59 |
| Aridity Index<1 (Energy Limited) | 60 | 181 | 68 | 17 | 0.90 | 4.15 | 75.25 |
| Aridity Index>1 (Water Limited) | 52 | 71 | 57 | 7 | 0.87 | 4.27 | 77.21 |
| Snowy | 58 | 232 | 61 | 9 | 0.90 | 4.34 | 81.12 |
| Non-Snowy | 54 | 20 | 64 | 15 | 0.86 | 3.85 | 63.83 |
| $Max\ SWE \leq$ 100mm | 103 | 51 | 107 | 20 | 0.85 | 4.03 | 68.48 |
| 100mm < $Max\ SWE \leq$ 200mm | 8 | 41 | 9 | 1 | 0.85 | 4.08 | 74.58 |
| 200mm < $Max\ SWE \leq$ 400mm | 1 | 69 | 7 | 3 | 0.91 | 4.36 | 84.45 |
| 400mm < $Max\ SWE \leq$ 600mm | 0 | 25 | 2 | 0 | 0.93 | 4.41 | 86.37 |
| $Max\ SWE$ > 600mm | 0 | 66 | 0 | 0 | 0.95 | 4.68 | 94.53 |
| Forest | 56 | 159 | 53 | 14 | 0.90 | 4.23 | 78.12 |
| Open | 56 | 93 | 72 | 10 | 0.85 | 4.14 | 73.33 |
| Climate Zone: Arid | 8 | 11 | 14 | 2 | 0.83 | 4.34 | 79.31 |
| Climate Zone: Temperate | 79 | 60 | 84 | 19 | 0.87 | 3.97 | 67.69 |
| Climate Zone: Cold | 25 | 179 | 27 | 3 | 0.90 | 4.40 | 84.04 |
| Climate Zone: Polar | 0 | 2 | 0 | 0 | 0.98 | 4.00 | 74.00 |

**Table S20 | KGE-based architecture selection within the sharing-enabled MCP candidate space with a maximum of one state.** For each basin grouping, the table reports the number of catchments assigned to each MCP-based model family, median testing-period $KGE_{ss}$, mean selected state dimensionality and mean number of trainable parameters. The candidate space excludes LSTMs while retaining MCP architectures with cross-state information sharing.

| Regional Categorization | KGE-based model selection | | | | | | |
|---|---|---|---|---|---|---|---|
| | HMCP | SNOWMCP | HMCPMR | HYDROMCP-Net | $KGE_{ss}^{50\%}$ | Ave State no | Ave Par no |
| CONUS | 70 | 247 | 196 | 0 | 0.81 | 1.00 | 10.59 |
| Eastern CONUS | 55 | 128 | 143 | 0 | 0.80 | 1.00 | 10.49 |
| Western CONUS | 15 | 119 | 53 | 0 | 0.86 | 1.00 | 10.76 |
| AM | 10 | 59 | 34 | 0 | 0.78 | 1.00 | 10.71 |
| RM | 0 | 53 | 1 | 0 | 0.89 | 1.00 | 11.00 |
| CRB | 4 | 22 | 8 | 0 | 0.83 | 1.00 | 10.65 |
| SN | 0 | 9 | 1 | 0 | 0.85 | 1.00 | 11.00 |
| CC | 0 | 22 | 11 | 0 | 0.88 | 1.00 | 11.00 |
| $EL < 1500$ m (low elevation) | 68 | 178 | 187 | 0 | 0.80 | 1.00 | 10.53 |
| $1500\ m <= EL <= 2500$ m (mid elevation) | 2 | 38 | 8 | 0 | 0.85 | 1.00 | 10.88 |
| $EL > 2500$ m (high elevation) | 0 | 31 | 1 | 0 | 0.90 | 1.00 | 11.00 |
| Aridity Index<1 (Energy Limited) | 36 | 162 | 128 | 0 | 0.82 | 1.00 | 10.67 |
| Aridity Index>1 (Water Limited) | 34 | 85 | 68 | 0 | 0.78 | 1.00 | 10.45 |
| Snowy | 36 | 223 | 101 | 0 | 0.81 | 1.00 | 10.70 |
| Non-Snowy | 34 | 24 | 95 | 0 | 0.82 | 1.00 | 10.33 |
| $Max\ SWE \leq$ 100mm | 63 | 68 | 150 | 0 | 0.80 | 1.00 | 10.33 |
| 100mm < $Max\ SWE \leq$ 200mm | 2 | 38 | 19 | 0 | 0.77 | 1.00 | 10.90 |
| 200mm < $Max\ SWE \leq$ 400mm | 5 | 58 | 17 | 0 | 0.79 | 1.00 | 10.81 |
| 400mm < $Max\ SWE \leq$ 600mm | 0 | 25 | 2 | 0 | 0.86 | 1.00 | 11.00 |
| $Max\ SWE$ > 600mm | 0 | 58 | 8 | 0 | 0.88 | 1.00 | 11.00 |
| Forest | 24 | 142 | 116 | 0 | 0.82 | 1.00 | 10.74 |
| Open | 46 | 105 | 80 | 0 | 0.79 | 1.00 | 10.40 |
| Climate Zone: Arid | 5 | 13 | 17 | 0 | 0.67 | 1.00 | 10.57 |
| Climate Zone: Temperate | 45 | 73 | 124 | 0 | 0.82 | 1.00 | 10.44 |
| Climate Zone: Cold | 20 | 159 | 55 | 0 | 0.80 | 1.00 | 10.74 |
| Climate Zone: Polar | 0 | 2 | 0 | 0 | 0.93 | 1.00 | 11.00 |

**Table S21 | KGE-based architecture selection within the sharing-enabled MCP candidate space with a maximum of 3 states.** For each basin grouping, the table reports the number of catchments assigned to each MCP-based model family, median testing-period $KGE_{ss}$, mean selected state dimensionality and mean number of trainable parameters. The candidate space excludes LSTMs while retaining MCP architectures with cross-state information sharing.

| Regional Categorization | KGE-based model selection | | | | | | |
|---|---|---|---|---|---|---|---|
| | HMCP | SNOWMCP | HMCPMR | HYDROMCP-Net | $KGE_{ss}^{50\%}$ | Ave State no | Ave Par no |
| CONUS | 81 | 256 | 121 | 55 | 0.88 | 2.79 | 42.54 |
| Eastern CONUS | 62 | 133 | 92 | 39 | 0.87 | 2.75 | 41.02 |
| Western CONUS | 19 | 123 | 29 | 16 | 0.92 | 2.87 | 45.18 |
| AM | 11 | 76 | 8 | 8 | 0.88 | 2.80 | 43.54 |
| RM | 1 | 52 | 1 | 0 | 0.96 | 2.96 | 49.17 |
| CRB | 4 | 21 | 5 | 4 | 0.93 | 2.82 | 43.38 |
| SN | 0 | 7 | 1 | 2 | 0.91 | 2.70 | 42.10 |
| CC | 0 | 33 | 0 | 0 | 0.92 | 3.00 | 49.91 |
| $EL < 1500$ m (low elevation) | 77 | 188 | 116 | 52 | 0.88 | 2.76 | 41.54 |
| $1500\ m <= EL <= 2500$ m (mid elevation) | 4 | 36 | 5 | 3 | 0.93 | 2.92 | 46.63 |
| $EL > 2500$ m (high elevation) | 0 | 32 | 0 | 0 | 0.96 | 3.00 | 49.97 |
| Aridity Index<1 (Energy Limited) | 42 | 183 | 68 | 33 | 0.89 | 2.81 | 43.25 |
| Aridity Index>1 (Water Limited) | 39 | 73 | 53 | 22 | 0.85 | 2.76 | 41.29 |
| Snowy | 41 | 240 | 49 | 30 | 0.89 | 2.84 | 44.32 |
| Non-Snowy | 40 | 16 | 72 | 25 | 0.87 | 2.69 | 38.35 |
| $Max\ SWE \leq$ 100mm | 71 | 51 | 114 | 45 | 0.85 | 2.70 | 38.86 |
| 100mm < $Max\ SWE \leq$ 200mm | 7 | 44 | 4 | 4 | 0.84 | 2.81 | 44.05 |
| 200mm < $Max\ SWE \leq$ 400mm | 3 | 71 | 1 | 5 | 0.91 | 2.91 | 47.21 |
| 400mm < $Max\ SWE \leq$ 600mm | 0 | 24 | 2 | 1 | 0.93 | 2.89 | 47.11 |
| $Max\ SWE$ > 600mm | 0 | 66 | 0 | 0 | 0.95 | 2.97 | 49.32 |
| Forest | 31 | 166 | 57 | 28 | 0.90 | 2.81 | 43.61 |
| Open | 50 | 90 | 64 | 27 | 0.85 | 2.77 | 41.23 |
| Climate Zone: Arid | 7 | 10 | 11 | 7 | 0.81 | 2.71 | 39.60 |
| Climate Zone: Temperate | 52 | 66 | 88 | 36 | 0.87 | 2.72 | 39.81 |
| Climate Zone: Cold | 22 | 178 | 22 | 12 | 0.90 | 2.88 | 45.74 |
| Climate Zone: Polar | 0 | 2 | 0 | 0 | 0.98 | 3.00 | 49.50 |

**Table S22 | KGE-based architecture selection within the sharing-enabled MCP candidate space with a maximum of 4 states.** For each basin grouping, the table reports the number of catchments assigned to each MCP-based model family, median testing-period $KGE_{ss}$, mean selected state dimensionality and mean number of trainable parameters. The candidate space excludes LSTMs while retaining MCP architectures with cross-state information sharing.

| Regional Categorization | KGE-based model selection | | | | | | |
|---|---|---|---|---|---|---|---|
| | HMCP | SNOWMCP | HMCPMR | HYDROMCP-Net | $KGE_{ss}^{50\%}$ | Ave State no | Ave Par no |
| CONUS | 96 | 259 | 124 | 34 | 0.89 | 3.54 | 59.62 |
| Eastern CONUS | 71 | 140 | 92 | 23 | 0.87 | 3.47 | 57.32 |
| Western CONUS | 25 | 119 | 32 | 11 | 0.92 | 3.66 | 63.63 |
| AM | 13 | 77 | 10 | 3 | 0.88 | 3.61 | 62.28 |
| RM | 1 | 52 | 1 | 0 | 0.96 | 3.85 | 70.81 |
| CRB | 5 | 19 | 7 | 3 | 0.94 | 3.59 | 61.03 |
| SN | 0 | 7 | 1 | 2 | 0.91 | 3.50 | 61.10 |
| CC | 0 | 33 | 0 | 0 | 0.92 | 3.76 | 68.82 |
| $EL < 1500$ m (low elevation) | 91 | 194 | 116 | 32 | 0.88 | 3.49 | 57.98 |
| $1500\ m <= EL <= 2500$ m (mid elevation) | 5 | 33 | 8 | 2 | 0.94 | 3.73 | 66.10 |
| $EL > 2500$ m (high elevation) | 0 | 32 | 0 | 0 | 0.96 | 3.91 | 72.03 |
| Aridity Index<1 (Energy Limited) | 47 | 189 | 70 | 20 | 0.90 | 3.56 | 60.64 |
| Aridity Index>1 (Water Limited) | 49 | 70 | 54 | 14 | 0.86 | 3.50 | 57.83 |
| Snowy | 50 | 240 | 53 | 17 | 0.89 | 3.61 | 62.28 |
| Non-Snowy | 46 | 19 | 71 | 17 | 0.85 | 3.38 | 53.37 |
| $Max\ SWE \leq$ 100mm | 86 | 54 | 114 | 27 | 0.85 | 3.41 | 54.12 |
| 100mm < $Max\ SWE \leq$ 200mm | 7 | 46 | 4 | 2 | 0.85 | 3.54 | 61.44 |
| 200mm < $Max\ SWE \leq$ 400mm | 3 | 71 | 2 | 4 | 0.90 | 3.65 | 65.23 |
| 400mm < $Max\ SWE \leq$ 600mm | 0 | 22 | 4 | 1 | 0.94 | 3.81 | 69.70 |
| $Max\ SWE$ > 600mm | 0 | 66 | 0 | 0 | 0.95 | 3.83 | 70.48 |
| Forest | 38 | 167 | 58 | 19 | 0.90 | 3.54 | 60.56 |
| Open | 58 | 92 | 66 | 15 | 0.85 | 3.54 | 58.47 |
| Climate Zone: Arid | 11 | 8 | 11 | 5 | 0.84 | 3.40 | 54.37 |
| Climate Zone: Temperate | 65 | 67 | 86 | 24 | 0.87 | 3.42 | 54.98 |
| Climate Zone: Cold | 20 | 182 | 27 | 5 | 0.90 | 3.68 | 65.09 |
| Climate Zone: Polar | 0 | 2 | 0 | 0 | 0.97 | 4.00 | 72.00 |

**Table S23 | KGE-based architecture selection within the sharing-enabled MCP candidate space with a maximum of 5 states.** For each basin grouping, the table reports the number of catchments assigned to each MCP-based model family, median testing-period $KGE_{ss}$, mean selected state dimensionality and mean number of trainable parameters. The candidate space excludes LSTMs while retaining MCP architectures with cross-state information sharing.

| Regional Categorization | KGE-based model selection | | | | | | |
|---|---|---|---|---|---|---|---|
| | HMCP | SNOWMCP | HMCPMR | HYDROMCP-Net | $KGE_{ss}^{50\%}$ | Ave State no | Ave Par no |
| CONUS | 110 | 254 | 120 | 29 | 0.89 | 4.25 | 77.97 |
| Eastern CONUS | 82 | 138 | 86 | 20 | 0.87 | 4.15 | 74.78 |
| Western CONUS | 28 | 116 | 34 | 9 | 0.92 | 4.41 | 83.52 |
| AM | 13 | 77 | 11 | 2 | 0.88 | 4.27 | 79.65 |
| RM | 1 | 52 | 1 | 0 | 0.97 | 4.72 | 94.98 |
| CRB | 5 | 19 | 7 | 3 | 0.95 | 4.38 | 82.88 |
| SN | 0 | 7 | 1 | 2 | 0.91 | 4.00 | 74.90 |
| CC | 0 | 32 | 1 | 0 | 0.92 | 4.52 | 89.30 |
| $EL < 1500$ m (low elevation) | 106 | 189 | 110 | 28 | 0.88 | 4.17 | 75.24 |
| $1500\ m <= EL <= 2500$ m (mid elevation) | 4 | 33 | 10 | 1 | 0.94 | 4.60 | 90.08 |
| $EL > 2500$ m (high elevation) | 0 | 32 | 0 | 0 | 0.97 | 4.78 | 96.66 |
| Aridity Index<1 (Energy Limited) | 53 | 186 | 69 | 18 | 0.90 | 4.23 | 78.15 |
| Aridity Index>1 (Water Limited) | 57 | 68 | 51 | 11 | 0.87 | 4.27 | 77.65 |
| Snowy | 56 | 233 | 59 | 12 | 0.90 | 4.38 | 82.78 |
| Non-Snowy | 54 | 21 | 61 | 17 | 0.85 | 3.93 | 66.65 |
| $Max\ SWE \leq$ 100mm | 100 | 53 | 104 | 24 | 0.85 | 4.11 | 71.35 |
| 100mm < $Max\ SWE \leq$ 200mm | 7 | 45 | 6 | 1 | 0.85 | 4.19 | 78.61 |
| 200mm < $Max\ SWE \leq$ 400mm | 3 | 67 | 6 | 4 | 0.91 | 4.39 | 85.28 |
| 400mm < $Max\ SWE \leq$ 600mm | 0 | 24 | 3 | 0 | 0.94 | 4.56 | 88.59 |
| $Max\ SWE$ > 600mm | 0 | 65 | 1 | 0 | 0.95 | 4.61 | 92.38 |
| Forest | 49 | 165 | 50 | 18 | 0.90 | 4.22 | 78.35 |
| Open | 61 | 89 | 70 | 11 | 0.85 | 4.28 | 77.50 |
| Climate Zone: Arid | 13 | 7 | 13 | 2 | 0.84 | 4.46 | 81.23 |
| Climate Zone: Temperate | 74 | 67 | 78 | 23 | 0.87 | 4.05 | 70.63 |
| Climate Zone: Cold | 23 | 178 | 29 | 4 | 0.90 | 4.41 | 84.88 |
| Climate Zone: Polar | 0 | 2 | 0 | 0 | 0.98 | 5.00 | 100.00 |

**Table S24 | Likelihood-based architecture selection within the no-sharing MCP candidate space with a maximum of one state.** For each basin grouping, the table reports the number of catchments assigned to each MCP-based model family, median testing-period $KGE_{ss}$, mean selected state dimensionality and mean number of trainable parameters. The C$N$ candidate space excludes LSTMs and cross-state information sharing.

| Regional Categorization | Likelihood-based model selection | | | | | | |
|---|---|---|---|---|---|---|---|
| | HMCP | SNOWMCP | HMCPMR | HYDROMCP-Net | $KGE_{ss}^{50\%}$ | Ave State no | Ave Par no |
| CONUS | 79 | 273 | 161 | 0 | 0.80 | 1.00 | 10.54 |
| Eastern CONUS | 58 | 145 | 123 | 0 | 0.79 | 1.00 | 10.47 |
| Western CONUS | 21 | 128 | 38 | 0 | 0.86 | 1.00 | 10.66 |
| AM | 12 | 58 | 33 | 0 | 0.79 | 1.00 | 10.65 |
| RM | 1 | 52 | 1 | 0 | 0.89 | 1.00 | 10.94 |
| CRB | 4 | 23 | 7 | 0 | 0.83 | 1.00 | 10.65 |
| SN | 0 | 9 | 1 | 0 | 0.85 | 1.00 | 11.00 |
| CC | 1 | 26 | 6 | 0 | 0.88 | 1.00 | 10.91 |
| $EL < 1500$ m (low elevation) | 75 | 208 | 150 | 0 | 0.80 | 1.00 | 10.48 |
| $1500\ m <= EL <= 2500$ m (mid elevation) | 3 | 35 | 10 | 0 | 0.85 | 1.00 | 10.81 |
| $EL > 2500$ m (high elevation) | 1 | 30 | 1 | 0 | 0.90 | 1.00 | 10.91 |
| Aridity Index<1 (Energy Limited) | 42 | 174 | 110 | 0 | 0.82 | 1.00 | 10.61 |
| Aridity Index>1 (Water Limited) | 37 | 99 | 51 | 0 | 0.78 | 1.00 | 10.41 |
| Snowy | 43 | 233 | 84 | 0 | 0.80 | 1.00 | 10.64 |
| Non-Snowy | 36 | 40 | 77 | 0 | 0.82 | 1.00 | 10.29 |
| $Max\ SWE \leq$ 100mm | 64 | 94 | 123 | 0 | 0.80 | 1.00 | 10.32 |
| 100mm < $Max\ SWE \leq$ 200mm | 6 | 36 | 17 | 0 | 0.76 | 1.00 | 10.69 |
| 200mm < $Max\ SWE \leq$ 400mm | 6 | 59 | 15 | 0 | 0.79 | 1.00 | 10.78 |
| 400mm < $Max\ SWE \leq$ 600mm | 1 | 26 | 0 | 0 | 0.86 | 1.00 | 10.89 |
| $Max\ SWE$ > 600mm | 2 | 58 | 6 | 0 | 0.88 | 1.00 | 10.91 |
| Forest | 32 | 152 | 98 | 0 | 0.82 | 1.00 | 10.66 |
| Open | 47 | 121 | 63 | 0 | 0.78 | 1.00 | 10.39 |
| Climate Zone: Arid | 9 | 15 | 11 | 0 | 0.67 | 1.00 | 10.23 |
| Climate Zone: Temperate | 46 | 91 | 105 | 0 | 0.82 | 1.00 | 10.43 |
| Climate Zone: Cold | 24 | 165 | 45 | 0 | 0.80 | 1.00 | 10.69 |
| Climate Zone: Polar | 0 | 2 | 0 | 0 | 0.93 | 1.00 | 11.00 |

**Table S25 | Likelihood-based architecture selection within the no-sharing MCP candidate space with a maximum of 2 states.** For each basin grouping, the table reports the number of catchments assigned to each MCP-based model family, median testing-period $KGE_{ss}$, mean selected state dimensionality and mean number of trainable parameters. The C$N$ candidate space excludes LSTMs and cross-state information sharing.

| Regional Categorization | Likelihood-based model selection | | | | | | |
|---|---|---|---|---|---|---|---|
| | HMCP | SNOWMCP | HMCPMR | HYDROMCP-Net | $KGE_{ss}^{50\%}$ | Ave State no | Ave Par no |
| CONUS | 94 | 162 | 132 | 125 | 0.84 | 1.92 | 21.16 |
| Eastern CONUS | 78 | 78 | 102 | 68 | 0.82 | 1.90 | 20.68 |
| Western CONUS | 16 | 84 | 30 | 57 | 0.88 | 1.96 | 21.98 |
| AM | 16 | 41 | 23 | 23 | 0.82 | 1.96 | 21.89 |
| RM | 1 | 31 | 1 | 21 | 0.90 | 1.96 | 22.24 |
| CRB | 2 | 14 | 4 | 14 | 0.85 | 1.91 | 21.26 |
| SN | 1 | 5 | 1 | 3 | 0.89 | 2.00 | 22.50 |
| CC | 0 | 22 | 3 | 8 | 0.89 | 2.00 | 23.27 |
| $EL < 1500$ m (low elevation) | 91 | 119 | 125 | 98 | 0.83 | 1.91 | 20.95 |
| $1500\ m <= EL <= 2500$ m (mid elevation) | 2 | 19 | 6 | 21 | 0.87 | 1.98 | 22.17 |
| $EL > 2500$ m (high elevation) | 1 | 24 | 1 | 6 | 0.91 | 1.94 | 22.44 |
| Aridity Index<1 (Energy Limited) | 56 | 108 | 84 | 78 | 0.85 | 1.94 | 21.56 |
| Aridity Index>1 (Water Limited) | 38 | 54 | 48 | 47 | 0.80 | 1.87 | 20.45 |
| Snowy | 47 | 133 | 75 | 105 | 0.84 | 1.96 | 21.85 |
| Non-Snowy | 47 | 29 | 57 | 20 | 0.84 | 1.82 | 19.52 |
| $Max\ SWE \leq$ 100mm | 80 | 47 | 101 | 53 | 0.82 | 1.87 | 20.11 |
| 100mm < $Max\ SWE \leq$ 200mm | 8 | 19 | 16 | 16 | 0.78 | 1.97 | 21.93 |
| 200mm < $Max\ SWE \leq$ 400mm | 5 | 34 | 13 | 28 | 0.84 | 2.00 | 22.58 |
| 400mm < $Max\ SWE \leq$ 600mm | 0 | 15 | 0 | 12 | 0.89 | 1.96 | 22.19 |
| $Max\ SWE >$ 600mm | 1 | 47 | 2 | 16 | 0.90 | 1.97 | 22.79 |
| Forest | 43 | 95 | 69 | 75 | 0.86 | 1.97 | 21.92 |
| Open | 51 | 67 | 63 | 50 | 0.80 | 1.85 | 20.23 |
| Climate Zone: Arid | 5 | 7 | 8 | 15 | 0.71 | 1.89 | 20.29 |
| Climate Zone: Temperate | 63 | 63 | 75 | 41 | 0.85 | 1.89 | 20.64 |
| Climate Zone: Cold | 26 | 90 | 49 | 69 | 0.84 | 1.95 | 21.79 |
| Climate Zone: Polar | 0 | 2 | 0 | 0 | 0.94 | 2.00 | 24.00 |

**Table S26 | Likelihood-based architecture selection within the no-sharing MCP candidate space with a maximum of 3 states.** For each basin grouping, the table reports the number of catchments assigned to each MCP-based model family, median testing-period $KGE_{ss}$, mean selected state dimensionality and mean number of trainable parameters. The C$N$ candidate space excludes LSTMs and cross-state information sharing.

| Regional Categorization | Likelihood-based model selection | | | | | | |
|---|---|---|---|---|---|---|---|
| | HMCP | SNOWMCP | HMCPMR | HYDROMCP-Net | $KGE_{ss}^{50\%}$ | Ave State no | Ave Par no |
| CONUS | 99 | 164 | 159 | 91 | 0.84 | 2.50 | 27.84 |
| Eastern CONUS | 82 | 71 | 124 | 49 | 0.83 | 2.48 | 27.26 |
| Western CONUS | 17 | 93 | 35 | 42 | 0.89 | 2.53 | 28.86 |
| AM | 22 | 40 | 25 | 16 | 0.83 | 2.65 | 29.34 |
| RM | 1 | 35 | 1 | 17 | 0.90 | 2.56 | 29.48 |
| CRB | 2 | 15 | 5 | 12 | 0.86 | 2.32 | 26.32 |
| SN | 0 | 5 | 3 | 2 | 0.85 | 2.50 | 29.40 |
| CC | 0 | 27 | 2 | 4 | 0.89 | 2.70 | 32.00 |
| $EL < 1500$ m (low elevation) | 97 | 114 | 150 | 72 | 0.84 | 2.48 | 27.43 |
| $1500\ m <= EL <= 2500$ m (mid elevation) | 1 | 25 | 8 | 14 | 0.87 | 2.56 | 29.69 |
| $EL > 2500$ m (high elevation) | 1 | 25 | 1 | 5 | 0.91 | 2.63 | 30.63 |
| Aridity Index<1 (Energy Limited) | 62 | 108 | 98 | 58 | 0.86 | 2.57 | 28.70 |
| Aridity Index>1 (Water Limited) | 37 | 56 | 61 | 33 | 0.81 | 2.37 | 26.36 |
| Snowy | 53 | 140 | 93 | 74 | 0.84 | 2.55 | 28.73 |
| Non-Snowy | 46 | 24 | 66 | 17 | 0.85 | 2.37 | 25.77 |
| $Max\ SWE \leq$ 100mm | 83 | 38 | 123 | 37 | 0.82 | 2.44 | 26.44 |
| 100mm < $Max\ SWE \leq$ 200mm | 10 | 18 | 19 | 12 | 0.80 | 2.54 | 28.41 |
| 200mm < $Max\ SWE \leq$ 400mm | 5 | 37 | 14 | 24 | 0.83 | 2.54 | 28.99 |
| 400mm < $Max\ SWE \leq$ 600mm | 0 | 17 | 1 | 9 | 0.89 | 2.48 | 28.56 |
| $Max\ SWE$ > 600mm | 1 | 54 | 2 | 9 | 0.90 | 2.68 | 31.62 |
| Forest | 46 | 99 | 82 | 55 | 0.86 | 2.60 | 29.13 |
| Open | 53 | 65 | 77 | 36 | 0.81 | 2.38 | 26.28 |
| Climate Zone: Arid | 3 | 8 | 11 | 13 | 0.71 | 2.26 | 25.29 |
| Climate Zone: Temperate | 64 | 57 | 91 | 30 | 0.86 | 2.48 | 27.21 |
| Climate Zone: Cold | 32 | 97 | 57 | 48 | 0.84 | 2.56 | 28.86 |
| Climate Zone: Polar | 0 | 2 | 0 | 0 | 0.94 | 2.50 | 30.00 |

**Table S27 | Likelihood-based architecture selection within the no-sharing MCP candidate space with a maximum of 4 states.** For each basin grouping, the table reports the number of catchments assigned to each MCP-based model family, median testing-period $KGE_{ss}$, mean selected state dimensionality and mean number of trainable parameters. The C$N$ candidate space excludes LSTMs and cross-state information sharing.

| Regional Categorization | Likelihood-based model selection | | | | | | |
|---|---|---|---|---|---|---|---|
| | HMCP | SNOWMCP | HMCPMR | HYDROMCP-Net | $KGE_{ss}^{50\%}$ | Ave State no | Ave Par no |
| CONUS | 97 | 179 | 168 | 69 | 0.84 | 3.07 | 34.53 |
| Eastern CONUS | 81 | 78 | 130 | 37 | 0.83 | 3.06 | 33.87 |
| Western CONUS | 16 | 101 | 38 | 32 | 0.88 | 3.09 | 35.68 |
| AM | 22 | 44 | 26 | 11 | 0.83 | 3.29 | 36.75 |
| RM | 0 | 38 | 1 | 15 | 0.91 | 3.17 | 37.11 |
| CRB | 1 | 15 | 7 | 11 | 0.86 | 2.82 | 32.44 |
| SN | 0 | 6 | 2 | 2 | 0.87 | 3.20 | 37.80 |
| CC | 0 | 27 | 2 | 4 | 0.89 | 3.09 | 36.73 |
| $EL < 1500$ m (low elevation) | 95 | 125 | 159 | 54 | 0.83 | 3.03 | 33.76 |
| $1500\ m <= EL <= 2500$ m (mid elevation) | 2 | 27 | 8 | 11 | 0.88 | 3.23 | 37.63 |
| $EL > 2500$ m (high elevation) | 0 | 27 | 1 | 4 | 0.91 | 3.41 | 40.41 |
| Aridity Index<1 (Energy Limited) | 64 | 118 | 103 | 41 | 0.86 | 3.14 | 35.23 |
| Aridity Index>1 (Water Limited) | 33 | 61 | 65 | 28 | 0.81 | 2.95 | 33.33 |
| Snowy | 55 | 152 | 94 | 59 | 0.84 | 3.12 | 35.36 |
| Non-Snowy | 42 | 27 | 74 | 10 | 0.85 | 2.96 | 32.60 |
| $Max\ SWE \leq$ 100mm | 85 | 40 | 132 | 24 | 0.82 | 2.99 | 32.67 |
| 100mm < $Max\ SWE \leq$ 200mm | 9 | 17 | 21 | 12 | 0.81 | 3.15 | 35.58 |
| 200mm < $Max\ SWE \leq$ 400mm | 3 | 47 | 13 | 17 | 0.84 | 3.10 | 36.11 |
| 400mm < $Max\ SWE \leq$ 600mm | 0 | 20 | 0 | 7 | 0.89 | 3.07 | 36.00 |
| $Max\ SWE$ > 600mm | 0 | 55 | 2 | 9 | 0.90 | 3.29 | 39.05 |
| Forest | 46 | 111 | 85 | 40 | 0.86 | 3.17 | 35.75 |
| Open | 51 | 68 | 83 | 29 | 0.81 | 2.95 | 33.05 |
| Climate Zone: Arid | 3 | 10 | 12 | 10 | 0.72 | 2.71 | 31.14 |
| Climate Zone: Temperate | 65 | 60 | 98 | 19 | 0.86 | 3.05 | 33.61 |
| Climate Zone: Cold | 29 | 107 | 58 | 40 | 0.85 | 3.15 | 36.00 |
| Climate Zone: Polar | 0 | 2 | 0 | 0 | 0.94 | 3.00 | 34.50 |

**Table S28 | KGE-based architecture selection within the no-sharing MCP candidate space with a maximum of one state.** For each basin grouping, the table reports the number of catchments assigned to each MCP-based model family, median testing-period $KGE_{ss}$, mean selected state dimensionality and mean number of trainable parameters. The C$N$ candidate space excludes LSTMs and cross-state information sharing.

| Regional Categorization | KGE-based model selection | | | | | | |
|---|---|---|---|---|---|---|---|
| | HMCP | SNOWMCP | HMCPMR | HYDROMCP-Net | $KGE_{ss}^{50\%}$ | Ave State no | Ave Par no |
| CONUS | 70 | 247 | 196 | 0 | 0.81 | 1.00 | 10.59 |
| Eastern CONUS | 55 | 128 | 143 | 0 | 0.80 | 1.00 | 10.49 |
| Western CONUS | 15 | 119 | 53 | 0 | 0.86 | 1.00 | 10.76 |
| AM | 10 | 59 | 34 | 0 | 0.78 | 1.00 | 10.71 |
| RM | 0 | 53 | 1 | 0 | 0.89 | 1.00 | 11.00 |
| CRB | 4 | 22 | 8 | 0 | 0.83 | 1.00 | 10.65 |
| SN | 0 | 9 | 1 | 0 | 0.85 | 1.00 | 11.00 |
| CC | 0 | 22 | 11 | 0 | 0.88 | 1.00 | 11.00 |
| $EL < 1500$ m (low elevation) | 68 | 178 | 187 | 0 | 0.80 | 1.00 | 10.53 |
| $1500\ m <= EL <= 2500$ m (mid elevation) | 2 | 38 | 8 | 0 | 0.85 | 1.00 | 10.88 |
| $EL > 2500$ m (high elevation) | 0 | 31 | 1 | 0 | 0.90 | 1.00 | 11.00 |
| Aridity Index<1 (Energy Limited) | 36 | 162 | 128 | 0 | 0.82 | 1.00 | 10.67 |
| Aridity Index>1 (Water Limited) | 34 | 85 | 68 | 0 | 0.78 | 1.00 | 10.45 |
| Snowy | 36 | 223 | 101 | 0 | 0.81 | 1.00 | 10.70 |
| Non-Snowy | 34 | 24 | 95 | 0 | 0.82 | 1.00 | 10.33 |
| $Max\ SWE \leq$ 100mm | 63 | 68 | 150 | 0 | 0.80 | 1.00 | 10.33 |
| 100mm < $Max\ SWE \leq$ 200mm | 2 | 38 | 19 | 0 | 0.77 | 1.00 | 10.90 |
| 200mm < $Max\ SWE \leq$ 400mm | 5 | 58 | 17 | 0 | 0.79 | 1.00 | 10.81 |
| 400mm < $Max\ SWE \leq$ 600mm | 0 | 25 | 2 | 0 | 0.86 | 1.00 | 11.00 |
| $Max\ SWE$ > 600mm | 0 | 58 | 8 | 0 | 0.88 | 1.00 | 11.00 |
| Forest | 24 | 142 | 116 | 0 | 0.82 | 1.00 | 10.74 |
| Open | 46 | 105 | 80 | 0 | 0.79 | 1.00 | 10.40 |
| Climate Zone: Arid | 5 | 13 | 17 | 0 | 0.67 | 1.00 | 10.57 |
| Climate Zone: Temperate | 45 | 73 | 124 | 0 | 0.82 | 1.00 | 10.44 |
| Climate Zone: Cold | 20 | 159 | 55 | 0 | 0.80 | 1.00 | 10.74 |
| Climate Zone: Polar | 0 | 2 | 0 | 0 | 0.93 | 1.00 | 11.00 |

**Table S29 | KGE-based architecture selection within the no-sharing MCP candidate space with a maximum of 2 states.** For each basin grouping, the table reports the number of catchments assigned to each MCP-based model family, median testing-period $KGE_{ss}$, mean selected state dimensionality and mean number of trainable parameters. The $CN$ candidate space excludes LSTMs and cross-state information sharing.

| Regional Categorization | KGE-based model selection | | | | | | |
|---|---|---|---|---|---|---|---|
| | HMCP | SNOWMCP | HMCPMR | HYDROMCP-Net | $KGE_{ss}^{50\%}$ | Ave State no | Ave Par no |
| CONUS | 73 | 137 | 155 | 148 | 0.84 | 1.97 | 21.98 |
| Eastern CONUS | 58 | 67 | 119 | 82 | 0.83 | 1.97 | 21.83 |
| Western CONUS | 15 | 70 | 36 | 66 | 0.88 | 1.98 | 22.25 |
| AM | 11 | 42 | 23 | 27 | 0.83 | 2.00 | 22.57 |
| RM | 0 | 27 | 0 | 27 | 0.90 | 1.96 | 22.02 |
| CRB | 4 | 11 | 4 | 15 | 0.85 | 1.97 | 21.59 |
| SN | 0 | 6 | 1 | 3 | 0.85 | 2.00 | 23.10 |
| CC | 0 | 19 | 6 | 8 | 0.89 | 2.00 | 23.27 |
| $EL < 1500$ m (low elevation) | 71 | 98 | 150 | 114 | 0.83 | 1.97 | 21.93 |
| $1500\ m <= EL <= 2500$ m (mid elevation) | 2 | 18 | 5 | 23 | 0.87 | 1.96 | 21.77 |
| $EL > 2500$ m (high elevation) | 0 | 21 | 0 | 11 | 0.91 | 2.00 | 22.97 |
| Aridity Index<1 (Energy Limited) | 44 | 98 | 99 | 85 | 0.86 | 1.98 | 22.12 |
| Aridity Index>1 (Water Limited) | 29 | 39 | 56 | 63 | 0.81 | 1.97 | 21.74 |
| Snowy | 32 | 119 | 83 | 126 | 0.84 | 1.98 | 22.21 |
| Non-Snowy | 41 | 18 | 72 | 22 | 0.85 | 1.95 | 21.44 |
| $Max\ SWE \leq$ 100mm | 65 | 33 | 122 | 61 | 0.83 | 1.96 | 21.51 |
| 100mm < $Max\ SWE \leq$ 200mm | 7 | 17 | 15 | 20 | 0.79 | 2.00 | 22.27 |
| 200mm < $Max\ SWE \leq$ 400mm | 1 | 34 | 13 | 32 | 0.84 | 1.99 | 22.56 |
| 400mm < $Max\ SWE \leq$ 600mm | 0 | 9 | 2 | 16 | 0.89 | 2.00 | 22.22 |
| $Max\ SWE$ > 600mm | 0 | 44 | 3 | 19 | 0.90 | 1.98 | 22.94 |
| Forest | 31 | 89 | 81 | 81 | 0.86 | 1.99 | 22.32 |
| Open | 42 | 48 | 74 | 67 | 0.81 | 1.96 | 21.57 |
| Climate Zone: Arid | 5 | 4 | 7 | 19 | 0.71 | 1.97 | 21.14 |
| Climate Zone: Temperate | 56 | 51 | 92 | 43 | 0.86 | 1.97 | 21.70 |
| Climate Zone: Cold | 12 | 81 | 56 | 85 | 0.84 | 1.98 | 22.39 |
| Climate Zone: Polar | 0 | 1 | 0 | 1 | 0.94 | 2.00 | 22.50 |

**Table S30 | KGE-based architecture selection within the no-sharing MCP candidate space with a maximum of 3 states.** For each basin grouping, the table reports the number of catchments assigned to each MCP-based model family, median testing-period $KGE_{ss}$, mean selected state dimensionality and mean number of trainable parameters. The C$N$ candidate space excludes LSTMs and cross-state information sharing.

| Regional Categorization | KGE-based model selection | | | | | | |
|---|---|---|---|---|---|---|---|
| | HMCP | SNOWMCP | HMCPMR | HYDROMCP-Net | $KGE_{ss}^{50\%}$ | Ave State no | Ave Par no |
| CONUS | 79 | 153 | 173 | 108 | 0.85 | 2.57 | 28.97 |
| Eastern CONUS | 64 | 67 | 132 | 63 | 0.83 | 2.57 | 28.70 |
| Western CONUS | 15 | 86 | 41 | 45 | 0.89 | 2.57 | 29.45 |
| AM | 18 | 42 | 22 | 21 | 0.83 | 2.62 | 29.33 |
| RM | 0 | 33 | 1 | 20 | 0.90 | 2.48 | 28.63 |
| CRB | 1 | 12 | 7 | 14 | 0.86 | 2.32 | 26.35 |
| SN | 0 | 5 | 3 | 2 | 0.86 | 2.50 | 29.40 |
| CC | 0 | 25 | 5 | 3 | 0.89 | 2.76 | 32.82 |
| $EL < 1500$ m (low elevation) | 78 | 105 | 164 | 86 | 0.84 | 2.58 | 28.85 |
| $1500\ m <= EL <= 2500$ m (mid elevation) | 1 | 25 | 8 | 14 | 0.87 | 2.50 | 28.90 |
| $EL > 2500$ m (high elevation) | 0 | 23 | 1 | 8 | 0.91 | 2.63 | 30.75 |
| Aridity Index<1 (Energy Limited) | 54 | 108 | 104 | 60 | 0.86 | 2.63 | 29.61 |
| Aridity Index>1 (Water Limited) | 25 | 45 | 69 | 48 | 0.81 | 2.47 | 27.87 |
| Snowy | 45 | 133 | 92 | 90 | 0.85 | 2.57 | 29.06 |
| Non-Snowy | 34 | 20 | 81 | 18 | 0.85 | 2.57 | 28.76 |
| $Max\ SWE \leq$ 100mm | 68 | 32 | 136 | 45 | 0.83 | 2.57 | 28.42 |
| 100mm < $Max\ SWE \leq$ 200mm | 7 | 16 | 20 | 16 | 0.81 | 2.56 | 28.88 |
| 200mm < $Max\ SWE \leq$ 400mm | 4 | 40 | 11 | 25 | 0.84 | 2.54 | 29.05 |
| 400mm < $Max\ SWE \leq$ 600mm | 0 | 11 | 3 | 13 | 0.89 | 2.44 | 27.89 |
| $Max\ SWE$ > 600mm | 0 | 54 | 3 | 9 | 0.90 | 2.68 | 31.76 |
| Forest | 36 | 100 | 91 | 55 | 0.86 | 2.66 | 30.19 |
| Open | 43 | 53 | 82 | 53 | 0.81 | 2.47 | 27.49 |
| Climate Zone: Arid | 3 | 6 | 10 | 16 | 0.77 | 2.34 | 26.03 |
| Climate Zone: Temperate | 56 | 51 | 107 | 28 | 0.86 | 2.63 | 29.33 |
| Climate Zone: Cold | 20 | 95 | 56 | 63 | 0.85 | 2.55 | 29.06 |
| Climate Zone: Polar | 0 | 1 | 0 | 1 | 0.94 | 2.50 | 28.50 |

**Table S31 | KGE-based architecture selection within the no-sharing MCP candidate space with a maximum of 4 states.** For each basin grouping, the table reports the number of catchments assigned to each MCP-based model family, median testing-period $KGE_{ss}$, mean selected state dimensionality and mean number of trainable parameters. The $CN$ candidate space excludes LSTMs and cross-state information sharing.

| Regional Categorization | KGE-based model selection | | | | | | |
|---|---|---|---|---|---|---|---|
| | HMCP | SNOWMCP | HMCPMR | HYDROMCP-Net | $KGE_{ss}^{50\%}$ | Ave State no | Ave Par no |
| CONUS | 91 | 161 | 176 | 85 | 0.85 | 3.22 | 36.27 |
| Eastern CONUS | 77 | 70 | 131 | 48 | 0.83 | 3.20 | 35.54 |
| Western CONUS | 14 | 91 | 45 | 37 | 0.89 | 3.25 | 37.53 |
| AM | 21 | 46 | 22 | 14 | 0.83 | 3.32 | 37.25 |
| RM | 0 | 37 | 1 | 16 | 0.91 | 3.17 | 37.09 |
| CRB | 1 | 12 | 8 | 13 | 0.86 | 2.94 | 33.79 |
| SN | 0 | 6 | 2 | 2 | 0.87 | 3.50 | 41.40 |
| CC | 0 | 26 | 5 | 2 | 0.90 | 3.45 | 41.27 |
| $EL < 1500$ m (low elevation) | 89 | 109 | 166 | 69 | 0.84 | 3.20 | 35.78 |
| $1500\ m <= EL <= 2500$ m (mid elevation) | 2 | 26 | 9 | 11 | 0.88 | 3.19 | 37.10 |
| $EL > 2500$ m (high elevation) | 0 | 26 | 1 | 5 | 0.91 | 3.50 | 41.53 |
| Aridity Index<1 (Energy Limited) | 60 | 113 | 107 | 46 | 0.86 | 3.28 | 36.94 |
| Aridity Index>1 (Water Limited) | 31 | 48 | 69 | 39 | 0.81 | 3.11 | 35.09 |
| Snowy | 48 | 141 | 98 | 73 | 0.84 | 3.24 | 36.86 |
| Non-Snowy | 43 | 20 | 78 | 12 | 0.86 | 3.16 | 34.86 |
| $Max\ SWE \leq$ 100mm | 81 | 30 | 136 | 34 | 0.84 | 3.20 | 35.01 |
| 100mm < $Max\ SWE\ \leq$ 200mm | 7 | 14 | 23 | 15 | 0.80 | 3.14 | 35.64 |
| 200mm < $Max\ SWE\ \leq$400mm | 3 | 46 | 12 | 19 | 0.84 | 3.24 | 37.73 |
| 400mm < $Max\ SWE\ \leq$600mm | 0 | 15 | 2 | 10 | 0.89 | 2.96 | 34.44 |
| $Max\ SWE$ > 600mm | 0 | 56 | 3 | 7 | 0.91 | 3.45 | 41.12 |
| Forest | 45 | 106 | 90 | 41 | 0.86 | 3.33 | 37.70 |
| Open | 46 | 55 | 86 | 44 | 0.82 | 3.08 | 34.51 |
| Climate Zone: Arid | 3 | 7 | 12 | 13 | 0.73 | 2.83 | 32.06 |
| Climate Zone: Temperate | 66 | 50 | 105 | 21 | 0.86 | 3.27 | 36.20 |
| Climate Zone: Cold | 22 | 103 | 59 | 50 | 0.85 | 3.22 | 36.98 |
| Climate Zone: Polar | 0 | 1 | 0 | 1 | 0.94 | 3.00 | 34.50 |

**Table S32 | Testing performance of representative MCP architectures across hydroclimatic basin groupings.** Median testing-period $KGE_{ss}$ is reported across 513 CAMELS-US catchments and by geographic region, elevation, aridity, snow influence, land cover and climate zone. The number of trainable parameters and state variables associated with each architecture is also reported.

| Regional Categorization | $HMCP$ $(1, AL)$ | $SNOWMCP$ $(1, BQ)$ | $HMCP$ $(1, ALMR)$ | $M_{Best}^{HYDRO}$ | $M_{Best}^{HYDRO-Net}$ | $HMCP_{none}^{DS}$ $(3, AL)$ | $HMCP_{SALO}^{DS}$ $(5, AL)$ | $SNOWMCP_{none}^{DS}$ $(5, BQ)$ | $SNOWMCP_{SALO}^{DS}$ $(3, BQ)$ | $HMCP_{none}^{DS}$ $(4, ALMR)$ | $HMCP_{SALO}^{DS}$ $(3, ALMR)$ |
|---|---|---|---|---|---|---|---|---|---|---|---|
| CONUS | 0.74 | 0.78 | 0.77 | 0.85 | 0.86 | 0.78 | 0.86 | 0.83 | 0.85 | 0.81 | 0.85 |
| Eastern CONUS | 0.77 | 0.76 | 0.78 | 0.84 | 0.85 | 0.80 | 0.84 | 0.81 | 0.82 | 0.82 | 0.83 |
| Western CONUS | 0.67 | 0.85 | 0.75 | 0.89 | 0.89 | 0.70 | 0.88 | 0.88 | 0.91 | 0.79 | 0.88 |
| AM | 0.75 | 0.78 | 0.75 | 0.83 | 0.85 | 0.78 | 0.82 | 0.83 | 0.87 | 0.80 | 0.82 |
| RM | 0.55 | 0.89 | 0.72 | 0.93 | 0.93 | 0.55 | 0.93 | 0.90 | 0.96 | 0.75 | 0.92 |
| CRB | 0.55 | 0.83 | 0.68 | 0.90 | 0.89 | 0.54 | 0.92 | 0.84 | 0.93 | 0.73 | 0.90 |
| SN | 0.59 | 0.85 | 0.70 | 0.91 | 0.91 | 0.61 | 0.91 | 0.89 | 0.93 | 0.71 | 0.85 |
| CC | 0.81 | 0.88 | 0.83 | 0.89 | 0.90 | 0.83 | 0.86 | 0.90 | 0.91 | 0.85 | 0.86 |
| $EL < 1500$ m (low elevation) | 0.77 | 0.77 | 0.78 | 0.84 | 0.86 | 0.81 | 0.85 | 0.82 | 0.84 | 0.82 | 0.84 |
| $1500\ m <= EL <= 2500$ m (mid elevation) | 0.59 | 0.85 | 0.69 | 0.89 | 0.89 | 0.60 | 0.91 | 0.86 | 0.93 | 0.74 | 0.89 |
| $EL > 2500$ m (high elevation) | 0.51 | 0.90 | 0.73 | 0.93 | 0.93 | 0.51 | 0.93 | 0.91 | 0.96 | 0.75 | 0.92 |
| Aridity Index<1 (Energy Limited) | 0.78 | 0.79 | 0.79 | 0.86 | 0.87 | 0.82 | 0.86 | 0.84 | 0.87 | 0.83 | 0.86 |
| Aridity Index>1 (Water Limited) | 0.64 | 0.74 | 0.72 | 0.83 | 0.83 | 0.68 | 0.86 | 0.78 | 0.74 | 0.76 | 0.83 |
| Snowy | 0.72 | 0.80 | 0.75 | 0.85 | 0.87 | 0.76 | 0.86 | 0.83 | 0.88 | 0.79 | 0.85 |
| Non-Snowy | 0.81 | 0.68 | 0.81 | 0.85 | 0.86 | 0.85 | 0.86 | 0.78 | 0.74 | 0.85 | 0.86 |
| $Max\ SWE \leq$ 100mm | 0.79 | 0.72 | 0.79 | 0.84 | 0.85 | 0.83 | 0.85 | 0.80 | 0.76 | 0.84 | 0.85 |
| 100mm $< Max\ SWE \leq$ 200mm | 0.72 | 0.75 | 0.72 | 0.82 | 0.83 | 0.75 | 0.83 | 0.81 | 0.85 | 0.78 | 0.82 |
| 200mm $< Max\ SWE \leq$ 400mm | 0.71 | 0.79 | 0.73 | 0.84 | 0.85 | 0.76 | 0.86 | 0.83 | 0.90 | 0.79 | 0.84 |
| 400mm $< Max\ SWE \leq$ 600mm | 0.66 | 0.86 | 0.74 | 0.90 | 0.90 | 0.68 | 0.90 | 0.88 | 0.93 | 0.76 | 0.88 |
| $Max\ SWE >$ 600mm | 0.60 | 0.88 | 0.75 | 0.92 | 0.91 | 0.61 | 0.91 | 0.90 | 0.94 | 0.77 | 0.90 |
| Forest | 0.77 | 0.79 | 0.78 | 0.87 | 0.88 | 0.81 | 0.87 | 0.85 | 0.88 | 0.82 | 0.86 |
| Open | 0.71 | 0.76 | 0.74 | 0.83 | 0.84 | 0.74 | 0.84 | 0.78 | 0.79 | 0.77 | 0.82 |
| Climate Zone: Arid | 0.60 | 0.51 | 0.62 | 0.78 | 0.78 | 0.58 | 0.84 | 0.63 | 0.59 | 0.70 | 0.75 |
| Climate Zone: Temperate | 0.81 | 0.77 | 0.82 | 0.86 | 0.87 | 0.85 | 0.86 | 0.82 | 0.81 | 0.85 | 0.86 |
| Climate Zone: Cold | 0.68 | 0.80 | 0.74 | 0.85 | 0.86 | 0.74 | 0.86 | 0.83 | 0.89 | 0.78 | 0.85 |
| Climate Zone: Polar | 0.43 | 0.93 | 0.82 | 0.94 | 0.94 | 0.43 | 0.95 | 0.94 | 0.98 | 0.82 | 0.89 |
| No. of Parameters | 8 | 11 | 11 | 8 to 23 | 8 to 25 | 27 | 85 | 60 | 48 | 48 | 48 |
| No. of State Variables | 1 | 1 | 1 | 1 to 2 | 1 to 2 | 3 | 5 | 5 | 3 | 4 | 3 |

**Table S33 | Likelihood-based architecture selection within the LSTM-only candidate space with a maximum of one state.** For each basin grouping, the table reports the number of catchments assigned to each LSTM state dimensionality, median testing-period $KGE_{ss}$, mean selected state dimensionality and mean number of trainable parameters.

| Regional Categorization | Likelihood-based model selection | | | | | | | |
|---|---|---|---|---|---|---|---|---|
| | $LSTM(1)$ | $LSTM(2)$ | $LSTM(3)$ | $LSTM(4)$ | $LSTM(5)$ | $KGE_{ss}^{50\%}$ | Ave State no | Ave Par no |
| CONUS | 513 | 0 | 0 | 0 | 0 | 0.81 | 1.00 | 12.00 |
| Eastern CONUS | 326 | 0 | 0 | 0 | 0 | 0.79 | 1.00 | 12.00 |
| Western CONUS | 187 | 0 | 0 | 0 | 0 | 0.85 | 1.00 | 12.00 |
| AM | 103 | 0 | 0 | 0 | 0 | 0.80 | 1.00 | 12.00 |
| RM | 54 | 0 | 0 | 0 | 0 | 0.83 | 1.00 | 12.00 |
| CRB | 34 | 0 | 0 | 0 | 0 | 0.74 | 1.00 | 12.00 |
| SN | 10 | 0 | 0 | 0 | 0 | 0.73 | 1.00 | 12.00 |
| CC | 33 | 0 | 0 | 0 | 0 | 0.87 | 1.00 | 12.00 |
| $EL < 1500$ m (low elevation) | 433 | 0 | 0 | 0 | 0 | 0.81 | 1.00 | 12.00 |
| $1500\ m <= EL <= 2500$ m (mid elevation) | 48 | 0 | 0 | 0 | 0 | 0.84 | 1.00 | 12.00 |
| $EL > 2500$ m (high elevation) | 32 | 0 | 0 | 0 | 0 | 0.79 | 1.00 | 12.00 |
| Aridity Index<1 (Energy Limited) | 326 | 0 | 0 | 0 | 0 | 0.82 | 1.00 | 12.00 |
| Aridity Index>1 (Water Limited) | 187 | 0 | 0 | 0 | 0 | 0.74 | 1.00 | 12.00 |
| Snowy | 360 | 0 | 0 | 0 | 0 | 0.80 | 1.00 | 12.00 |
| Non-Snowy | 153 | 0 | 0 | 0 | 0 | 0.82 | 1.00 | 12.00 |
| $Max\ SWE \leq$ 100mm | 281 | 0 | 0 | 0 | 0 | 0.79 | 1.00 | 12.00 |
| 100mm $<\ Max\ SWE\ \leq$ 200mm | 59 | 0 | 0 | 0 | 0 | 0.77 | 1.00 | 12.00 |
| 200mm $<\ Max\ SWE\ \leq$ 400mm | 80 | 0 | 0 | 0 | 0 | 0.81 | 1.00 | 12.00 |
| 400mm $<\ Max\ SWE\ \leq$ 600mm | 27 | 0 | 0 | 0 | 0 | 0.82 | 1.00 | 12.00 |
| $Max\ SWE >$ 600mm | 66 | 0 | 0 | 0 | 0 | 0.85 | 1.00 | 12.00 |
| Forest | 282 | 0 | 0 | 0 | 0 | 0.83 | 1.00 | 12.00 |
| Open | 231 | 0 | 0 | 0 | 0 | 0.77 | 1.00 | 12.00 |
| Climate Zone: Arid | 35 | 0 | 0 | 0 | 0 | 0.70 | 1.00 | 12.00 |
| Climate Zone: Temperate | 242 | 0 | 0 | 0 | 0 | 0.83 | 1.00 | 12.00 |
| Climate Zone: Cold | 234 | 0 | 0 | 0 | 0 | 0.79 | 1.00 | 12.00 |
| Climate Zone: Polar | 2 | 0 | 0 | 0 | 0 | 0.83 | 1.00 | 12.00 |

**Table S34 | Likelihood-based architecture selection within the LSTM-only candidate space with a maximum of** 2 **states.** For each basin grouping, the table reports the number of catchments assigned to each LSTM state dimensionality, median testing-period $KGE_{ss}$, mean selected state dimensionality and mean number of trainable parameters.

| Regional Categorization | Likelihood-based model selection | | | | | | | |
|---|---|---|---|---|---|---|---|---|
| | $LSTM(1)$ | $LSTM(2)$ | $LSTM(3)$ | $LSTM(4)$ | $LSTM(5)$ | $KGE_{ss}^{50\%}$ | Ave State no | Ave Par no |
| CONUS | 35 | 478 | 0 | 0 | 0 | 0.88 | 1.93 | 40.88 |
| Eastern CONUS | 26 | 300 | 0 | 0 | 0 | 0.86 | 1.92 | 40.53 |
| Western CONUS | 9 | 178 | 0 | 0 | 0 | 0.91 | 1.95 | 41.51 |
| AM | 10 | 93 | 0 | 0 | 0 | 0.86 | 1.90 | 39.99 |
| RM | 0 | 54 | 0 | 0 | 0 | 0.95 | 2.00 | 43.00 |
| CRB | 2 | 32 | 0 | 0 | 0 | 0.93 | 1.94 | 41.18 |
| SN | 0 | 10 | 0 | 0 | 0 | 0.91 | 2.00 | 43.00 |
| CC | 1 | 32 | 0 | 0 | 0 | 0.91 | 1.97 | 42.06 |
| $EL < 1500$ m (low elevation) | 34 | 399 | 0 | 0 | 0 | 0.87 | 1.92 | 40.57 |
| $1500\ m <= EL <= 2500$ m (mid elevation) | 1 | 47 | 0 | 0 | 0 | 0.93 | 1.98 | 42.35 |
| $EL > 2500$ m (high elevation) | 0 | 32 | 0 | 0 | 0 | 0.95 | 2.00 | 43.00 |
| Aridity Index<1 (Energy Limited) | 16 | 310 | 0 | 0 | 0 | 0.88 | 1.95 | 41.48 |
| Aridity Index>1 (Water Limited) | 19 | 168 | 0 | 0 | 0 | 0.86 | 1.90 | 39.85 |
| Snowy | 21 | 339 | 0 | 0 | 0 | 0.88 | 1.94 | 41.19 |
| Non-Snowy | 14 | 139 | 0 | 0 | 0 | 0.87 | 1.91 | 40.16 |
| $Max\ SWE \leq$ 100mm | 24 | 257 | 0 | 0 | 0 | 0.87 | 1.91 | 40.35 |
| 100mm < $Max\ SWE \leq$ 200mm | 7 | 52 | 0 | 0 | 0 | 0.86 | 1.88 | 39.32 |
| 200mm < $Max\ SWE \leq$ 400mm | 4 | 76 | 0 | 0 | 0 | 0.88 | 1.95 | 41.45 |
| 400mm < $Max\ SWE \leq$ 600mm | 0 | 27 | 0 | 0 | 0 | 0.92 | 2.00 | 43.00 |
| $Max\ SWE$ > 600mm | 0 | 66 | 0 | 0 | 0 | 0.93 | 2.00 | 43.00 |
| Forest | 12 | 270 | 0 | 0 | 0 | 0.89 | 1.96 | 41.68 |
| Open | 23 | 208 | 0 | 0 | 0 | 0.87 | 1.90 | 39.91 |
| Climate Zone: Arid | 5 | 30 | 0 | 0 | 0 | 0.83 | 1.86 | 38.57 |
| Climate Zone: Temperate | 19 | 223 | 0 | 0 | 0 | 0.88 | 1.92 | 40.57 |
| Climate Zone: Cold | 11 | 223 | 0 | 0 | 0 | 0.88 | 1.95 | 41.54 |
| Climate Zone: Polar | 0 | 2 | 0 | 0 | 0 | 0.97 | 2.00 | 43.00 |

**Table S35 | Likelihood-based architecture selection within the LSTM-only candidate space with a maximum of 3 states.** For each basin grouping, the table reports the number of catchments assigned to each LSTM state dimensionality, median testing-period $KGE_{ss}$, mean selected state dimensionality and mean number of trainable parameters.

| Regional Categorization | Likelihood-based model selection | | | | | | | |
|---|---|---|---|---|---|---|---|---|
| | $LSTM(1)$ | $LSTM(2)$ | $LSTM(3)$ | $LSTM(4)$ | $LSTM(5)$ | $KGE_{ss}^{50\%}$ | Ave State no | Ave Par no |
| CONUS | 11 | 75 | 427 | 0 | 0 | 0.88 | 2.81 | 69.80 |
| Eastern CONUS | 9 | 46 | 271 | 0 | 0 | 0.87 | 2.80 | 69.58 |
| Western CONUS | 2 | 29 | 156 | 0 | 0 | 0.92 | 2.82 | 70.20 |
| AM | 3 | 15 | 85 | 0 | 0 | 0.86 | 2.80 | 69.33 |
| RM | 0 | 0 | 54 | 0 | 0 | 0.95 | 3.00 | 76.00 |
| CRB | 1 | 3 | 30 | 0 | 0 | 0.94 | 2.85 | 71.21 |
| SN | 0 | 2 | 8 | 0 | 0 | 0.93 | 2.80 | 69.40 |
| CC | 0 | 6 | 27 | 0 | 0 | 0.91 | 2.82 | 70.00 |
| $EL < 1500$ m (low elevation) | 10 | 73 | 350 | 0 | 0 | 0.87 | 2.79 | 68.96 |
| $1500\ m <= EL <= 2500$ m (mid elevation) | 1 | 2 | 45 | 0 | 0 | 0.94 | 2.92 | 73.29 |
| $EL > 2500$ m (high elevation) | 0 | 0 | 32 | 0 | 0 | 0.95 | 3.00 | 76.00 |
| Aridity Index<1 (Energy Limited) | 5 | 46 | 275 | 0 | 0 | 0.88 | 2.83 | 70.36 |
| Aridity Index>1 (Water Limited) | 6 | 29 | 152 | 0 | 0 | 0.87 | 2.78 | 68.83 |
| Snowy | 7 | 49 | 304 | 0 | 0 | 0.88 | 2.83 | 70.26 |
| Non-Snowy | 4 | 26 | 123 | 0 | 0 | 0.88 | 2.78 | 68.72 |
| $Max\ SWE \leq$ 100mm | 6 | 52 | 223 | 0 | 0 | 0.87 | 2.77 | 68.53 |
| 100mm < $Max\ SWE \leq$ 200mm | 4 | 5 | 50 | 0 | 0 | 0.87 | 2.78 | 68.86 |
| 200mm < $Max\ SWE \leq$ 400mm | 1 | 9 | 70 | 0 | 0 | 0.89 | 2.86 | 71.49 |
| 400mm < $Max\ SWE \leq$ 600mm | 0 | 1 | 26 | 0 | 0 | 0.93 | 2.96 | 74.78 |
| $Max\ SWE >$ 600mm | 0 | 8 | 58 | 0 | 0 | 0.94 | 2.88 | 72.00 |
| Forest | 4 | 41 | 237 | 0 | 0 | 0.89 | 2.83 | 70.29 |
| Open | 7 | 34 | 190 | 0 | 0 | 0.87 | 2.79 | 69.20 |
| Climate Zone: Arid | 2 | 5 | 28 | 0 | 0 | 0.83 | 2.74 | 67.63 |
| Climate Zone: Temperate | 7 | 43 | 192 | 0 | 0 | 0.88 | 2.76 | 68.29 |
| Climate Zone: Cold | 2 | 27 | 205 | 0 | 0 | 0.89 | 2.87 | 71.65 |
| Climate Zone: Polar | 0 | 0 | 2 | 0 | 0 | 0.97 | 3.00 | 76.00 |

**Table S36 | Likelihood-based architecture selection within the LSTM-only candidate space with a maximum of** 4 **states.** For each basin grouping, the table reports the number of catchments assigned to each LSTM state dimensionality, median testing-period $KGE_{ss}$, mean selected state dimensionality and mean number of trainable parameters.

| Regional Categorization | Likelihood-based model selection | | | | | | | |
|---|---|---|---|---|---|---|---|---|
| | $LSTM(1)$ | $LSTM(2)$ | $LSTM(3)$ | $LSTM(4)$ | $LSTM(5)$ | $KGE_{ss}^{50\%}$ | Ave State no | Ave Par no |
| CONUS | 4 | 23 | 138 | 348 | 0 | 0.89 | 3.62 | 101.83 |
| Eastern CONUS | 4 | 18 | 95 | 209 | 0 | 0.87 | 3.56 | 99.68 |
| Western CONUS | 0 | 5 | 43 | 139 | 0 | 0.92 | 3.72 | 105.59 |
| AM | 1 | 5 | 23 | 74 | 0 | 0.86 | 3.65 | 103.23 |
| RM | 0 | 0 | 3 | 51 | 0 | 0.96 | 3.94 | 114.72 |
| CRB | 0 | 0 | 8 | 26 | 0 | 0.94 | 3.76 | 107.35 |
| SN | 0 | 0 | 5 | 5 | 0 | 0.94 | 3.50 | 96.50 |
| CC | 0 | 0 | 10 | 23 | 0 | 0.90 | 3.70 | 104.58 |
| $EL < 1500$ m (low elevation) | 4 | 23 | 125 | 281 | 0 | 0.88 | 3.58 | 100.26 |
| $1500\ m <= EL <= 2500$ m (mid elevation) | 0 | 0 | 11 | 37 | 0 | 0.94 | 3.77 | 107.60 |
| $EL > 2500$ m (high elevation) | 0 | 0 | 2 | 30 | 0 | 0.96 | 3.94 | 114.44 |
| Aridity Index<1 (Energy Limited) | 2 | 13 | 90 | 221 | 0 | 0.89 | 3.63 | 102.09 |
| Aridity Index>1 (Water Limited) | 2 | 10 | 48 | 127 | 0 | 0.87 | 3.60 | 101.40 |
| Snowy | 2 | 17 | 91 | 250 | 0 | 0.89 | 3.64 | 102.56 |
| Non-Snowy | 2 | 6 | 47 | 98 | 0 | 0.88 | 3.58 | 100.13 |
| $Max\ SWE \leq$ 100mm | 2 | 21 | 90 | 168 | 0 | 0.87 | 3.51 | 97.59 |
| 100mm < $Max\ SWE \leq$ 200mm | 2 | 1 | 18 | 38 | 0 | 0.87 | 3.56 | 99.68 |
| 200mm < $Max\ SWE \leq$ 400mm | 0 | 1 | 16 | 63 | 0 | 0.90 | 3.78 | 107.88 |
| 400mm < $Max\ SWE \leq$ 600mm | 0 | 0 | 7 | 20 | 0 | 0.94 | 3.74 | 106.37 |
| $Max\ SWE$ > 600mm | 0 | 0 | 7 | 59 | 0 | 0.94 | 3.89 | 112.65 |
| Forest | 1 | 11 | 66 | 204 | 0 | 0.90 | 3.68 | 104.15 |
| Open | 3 | 12 | 72 | 144 | 0 | 0.87 | 3.55 | 99.01 |
| Climate Zone: Arid | 0 | 1 | 13 | 21 | 0 | 0.85 | 3.57 | 99.66 |
| Climate Zone: Temperate | 3 | 15 | 75 | 149 | 0 | 0.88 | 3.53 | 98.40 |
| Climate Zone: Cold | 1 | 7 | 50 | 176 | 0 | 0.90 | 3.71 | 105.58 |
| Climate Zone: Polar | 0 | 0 | 0 | 2 | 0 | 0.97 | 4.00 | 117.00 |

**Table S37 | Likelihood-based architecture selection within the LSTM-only candidate space with a maximum of** 5 **states.** For each basin grouping, the table reports the number of catchments assigned to each LSTM state dimensionality, median testing-period $KGE_{ss}$, mean selected state dimensionality and mean number of trainable parameters.

| Regional Categorization | Likelihood-based model selection | | | | | | | |
|---|---|---|---|---|---|---|---|---|
| | $LSTM(1)$ | $LSTM(2)$ | $LSTM(3)$ | $LSTM(4)$ | $LSTM(5)$ | $KGE_{ss}^{50\%}$ | Ave State no | Ave Par no |
| CONUS | 1 | 12 | 67 | 150 | 283 | 0.89 | 4.37 | 136.74 |
| Eastern CONUS | 1 | 10 | 44 | 101 | 170 | 0.87 | 4.32 | 134.43 |
| Western CONUS | 0 | 2 | 23 | 49 | 113 | 0.92 | 4.46 | 140.78 |
| AM | 0 | 2 | 8 | 36 | 57 | 0.87 | 4.44 | 139.50 |
| RM | 0 | 0 | 0 | 10 | 44 | 0.96 | 4.81 | 156.93 |
| CRB | 0 | 0 | 2 | 7 | 25 | 0.95 | 4.68 | 150.62 |
| SN | 0 | 0 | 0 | 3 | 7 | 0.93 | 4.70 | 151.30 |
| CC | 0 | 0 | 6 | 9 | 18 | 0.91 | 4.36 | 136.27 |
| $EL < 1500$ m (low elevation) | 1 | 12 | 65 | 133 | 222 | 0.88 | 4.30 | 133.67 |
| $1500\ m\ <= EL <=\ 2500$ m (mid elevation) | 0 | 0 | 2 | 10 | 36 | 0.95 | 4.71 | 152.04 |
| $EL > 2500$ m (high elevation) | 0 | 0 | 0 | 7 | 25 | 0.96 | 4.78 | 155.28 |
| Aridity Index<1 (Energy Limited) | 0 | 7 | 44 | 100 | 175 | 0.89 | 4.36 | 136.18 |
| Aridity Index>1 (Water Limited) | 1 | 5 | 23 | 50 | 108 | 0.88 | 4.39 | 137.72 |
| Snowy | 0 | 7 | 37 | 93 | 223 | 0.89 | 4.48 | 141.70 |
| Non-Snowy | 1 | 5 | 30 | 57 | 60 | 0.88 | 4.11 | 125.07 |
| $Max\ SWE \leq$ 100mm | 1 | 11 | 51 | 90 | 128 | 0.87 | 4.19 | 128.61 |
| 100mm < $Max\ SWE\ \leq$ 200mm | 0 | 0 | 7 | 15 | 37 | 0.87 | 4.51 | 142.86 |
| 200mm < $Max\ SWE\ \leq$400mm | 0 | 1 | 6 | 25 | 48 | 0.90 | 4.50 | 142.40 |
| 400mm < $Max\ SWE\ \leq$600mm | 0 | 0 | 2 | 6 | 19 | 0.94 | 4.63 | 148.44 |
| $Max\ SWE$ > 600mm | 0 | 0 | 1 | 14 | 51 | 0.94 | 4.76 | 154.24 |
| Forest | 0 | 6 | 29 | 87 | 160 | 0.90 | 4.42 | 139.01 |
| Open | 1 | 6 | 38 | 63 | 123 | 0.87 | 4.30 | 133.97 |
| Climate Zone: Arid | 0 | 1 | 6 | 8 | 20 | 0.87 | 4.34 | 135.86 |
| Climate Zone: Temperate | 1 | 8 | 41 | 84 | 108 | 0.88 | 4.20 | 129.04 |
| Climate Zone: Cold | 0 | 3 | 20 | 58 | 153 | 0.90 | 4.54 | 144.59 |
| Climate Zone: Polar | 0 | 0 | 0 | 0 | 2 | 0.97 | 5.00 | 166.00 |

**Table S38 | KGE-based architecture selection within the LSTM-only candidate space with a maximum of one state.** For each basin grouping, the table reports the number of catchments assigned to each LSTM state dimensionality, median testing-period $KGE_{ss}$, mean selected state dimensionality and mean number of trainable parameters.

| Regional Categorization | KGE-based model selection | | | | | | | |
|---|---|---|---|---|---|---|---|---|
| | $LSTM(1)$ | $LSTM(2)$ | $LSTM(3)$ | $LSTM(4)$ | $LSTM(5)$ | $KGE_{ss}^{50\%}$ | Ave State no | Ave Par no |
| CONUS | 513 | 0 | 0 | 0 | 0 | 0.81 | 1.00 | 12.00 |
| Eastern CONUS | 326 | 0 | 0 | 0 | 0 | 0.79 | 1.00 | 12.00 |
| Western CONUS | 187 | 0 | 0 | 0 | 0 | 0.85 | 1.00 | 12.00 |
| AM | 103 | 0 | 0 | 0 | 0 | 0.80 | 1.00 | 12.00 |
| RM | 54 | 0 | 0 | 0 | 0 | 0.83 | 1.00 | 12.00 |
| CRB | 34 | 0 | 0 | 0 | 0 | 0.74 | 1.00 | 12.00 |
| SN | 10 | 0 | 0 | 0 | 0 | 0.73 | 1.00 | 12.00 |
| CC | 33 | 0 | 0 | 0 | 0 | 0.87 | 1.00 | 12.00 |
| $EL < 1500$ m (low elevation) | 433 | 0 | 0 | 0 | 0 | 0.81 | 1.00 | 12.00 |
| $1500\,m <= EL <= 2500$ m (mid elevation) | 48 | 0 | 0 | 0 | 0 | 0.84 | 1.00 | 12.00 |
| $EL > 2500$ m (high elevation) | 32 | 0 | 0 | 0 | 0 | 0.79 | 1.00 | 12.00 |
| Aridity Index<1 (Energy Limited) | 326 | 0 | 0 | 0 | 0 | 0.82 | 1.00 | 12.00 |
| Aridity Index>1 (Water Limited) | 187 | 0 | 0 | 0 | 0 | 0.74 | 1.00 | 12.00 |
| Snowy | 360 | 0 | 0 | 0 | 0 | 0.80 | 1.00 | 12.00 |
| Non-Snowy | 153 | 0 | 0 | 0 | 0 | 0.82 | 1.00 | 12.00 |
| $Max\ SWE \leq$ 100mm | 281 | 0 | 0 | 0 | 0 | 0.79 | 1.00 | 12.00 |
| 100mm $<\ Max\ SWE\ \leq$ 200mm | 59 | 0 | 0 | 0 | 0 | 0.77 | 1.00 | 12.00 |
| 200mm $<\ Max\ SWE\ \leq$ 400mm | 80 | 0 | 0 | 0 | 0 | 0.81 | 1.00 | 12.00 |
| 400mm $<\ Max\ SWE\ \leq$ 600mm | 27 | 0 | 0 | 0 | 0 | 0.82 | 1.00 | 12.00 |
| $Max\ SWE >$ 600mm | 66 | 0 | 0 | 0 | 0 | 0.85 | 1.00 | 12.00 |
| Forest | 282 | 0 | 0 | 0 | 0 | 0.83 | 1.00 | 12.00 |
| Open | 231 | 0 | 0 | 0 | 0 | 0.77 | 1.00 | 12.00 |
| Climate Zone: Arid | 35 | 0 | 0 | 0 | 0 | 0.70 | 1.00 | 12.00 |
| Climate Zone: Temperate | 242 | 0 | 0 | 0 | 0 | 0.83 | 1.00 | 12.00 |
| Climate Zone: Cold | 234 | 0 | 0 | 0 | 0 | 0.79 | 1.00 | 12.00 |
| Climate Zone: Polar | 2 | 0 | 0 | 0 | 0 | 0.83 | 1.00 | 12.00 |

**Table S39 | KGE-based architecture selection within the LSTM-only candidate space with a maximum of 2 states.** For each basin grouping, the table reports the number of catchments assigned to each LSTM state dimensionality, median testing-period $KGE_{ss}$, mean selected state dimensionality and mean number of trainable parameters.

| Regional Categorization | KGE-based model selection | | | | | | | |
|---|---|---|---|---|---|---|---|---|
| | $LSTM(1)$ | $LSTM(2)$ | $LSTM(3)$ | $LSTM(4)$ | $LSTM(5)$ | $KGE_{ss}^{50\%}$ | Ave State no | Ave Par no |
| CONUS | 1 | 512 | 0 | 0 | 0 | 0.88 | 2.00 | 42.94 |
| Eastern CONUS | 1 | 325 | 0 | 0 | 0 | 0.87 | 2.00 | 42.90 |
| Western CONUS | 0 | 187 | 0 | 0 | 0 | 0.91 | 2.00 | 43.00 |
| AM | 0 | 103 | 0 | 0 | 0 | 0.86 | 2.00 | 43.00 |
| RM | 0 | 54 | 0 | 0 | 0 | 0.95 | 2.00 | 43.00 |
| CRB | 0 | 34 | 0 | 0 | 0 | 0.93 | 2.00 | 43.00 |
| SN | 0 | 10 | 0 | 0 | 0 | 0.91 | 2.00 | 43.00 |
| CC | 0 | 33 | 0 | 0 | 0 | 0.91 | 2.00 | 43.00 |
| $EL < 1500$ m (low elevation) | 1 | 432 | 0 | 0 | 0 | 0.87 | 2.00 | 42.93 |
| $1500\ m <= EL <= 2500$ m (mid elevation) | 0 | 48 | 0 | 0 | 0 | 0.93 | 2.00 | 43.00 |
| $EL > 2500$ m (high elevation) | 0 | 32 | 0 | 0 | 0 | 0.95 | 2.00 | 43.00 |
| Aridity Index<1 (Energy Limited) | 0 | 326 | 0 | 0 | 0 | 0.88 | 2.00 | 43.00 |
| Aridity Index>1 (Water Limited) | 1 | 186 | 0 | 0 | 0 | 0.87 | 1.99 | 42.83 |
| Snowy | 0 | 360 | 0 | 0 | 0 | 0.88 | 2.00 | 43.00 |
| Non-Snowy | 1 | 152 | 0 | 0 | 0 | 0.88 | 1.99 | 42.80 |
| $Max\ SWE \leq$ 100mm | 1 | 280 | 0 | 0 | 0 | 0.87 | 2.00 | 42.89 |
| 100mm < $Max\ SWE \leq$ 200mm | 0 | 59 | 0 | 0 | 0 | 0.86 | 2.00 | 43.00 |
| 200mm < $Max\ SWE \leq$ 400mm | 0 | 80 | 0 | 0 | 0 | 0.88 | 2.00 | 43.00 |
| 400mm < $Max\ SWE \leq$ 600mm | 0 | 27 | 0 | 0 | 0 | 0.92 | 2.00 | 43.00 |
| $Max\ SWE$ > 600mm | 0 | 66 | 0 | 0 | 0 | 0.93 | 2.00 | 43.00 |
| Forest | 1 | 281 | 0 | 0 | 0 | 0.89 | 2.00 | 42.89 |
| Open | 0 | 231 | 0 | 0 | 0 | 0.87 | 2.00 | 43.00 |
| Climate Zone: Arid | 0 | 35 | 0 | 0 | 0 | 0.84 | 2.00 | 43.00 |
| Climate Zone: Temperate | 1 | 241 | 0 | 0 | 0 | 0.88 | 2.00 | 42.87 |
| Climate Zone: Cold | 0 | 234 | 0 | 0 | 0 | 0.88 | 2.00 | 43.00 |
| Climate Zone: Polar | 0 | 2 | 0 | 0 | 0 | 0.97 | 2.00 | 43.00 |

**Table S40 | KGE-based architecture selection within the LSTM-only candidate space with a maximum of 3 states.** For each basin grouping, the table reports the number of catchments assigned to each LSTM state dimensionality, median testing-period $KGE_{ss}$, mean selected state dimensionality and mean number of trainable parameters.

| Regional Categorization | KGE-based model selection | | | | | | | |
|---|---|---|---|---|---|---|---|---|
| | $LSTM(1)$ | $LSTM(2)$ | $LSTM(3)$ | $LSTM(4)$ | $LSTM(5)$ | $KGE_{ss}^{50\%}$ | Ave State no | Ave Par no |
| CONUS | 0 | 49 | 464 | 0 | 0 | 0.88 | 2.90 | 72.85 |
| Eastern CONUS | 0 | 33 | 293 | 0 | 0 | 0.87 | 2.90 | 72.66 |
| Western CONUS | 0 | 16 | 171 | 0 | 0 | 0.92 | 2.91 | 73.18 |
| AM | 0 | 9 | 94 | 0 | 0 | 0.86 | 2.91 | 73.12 |
| RM | 0 | 1 | 53 | 0 | 0 | 0.95 | 2.98 | 75.39 |
| CRB | 0 | 1 | 33 | 0 | 0 | 0.94 | 2.97 | 75.03 |
| SN | 0 | 0 | 10 | 0 | 0 | 0.93 | 3.00 | 76.00 |
| CC | 0 | 3 | 30 | 0 | 0 | 0.91 | 2.91 | 73.00 |
| $EL < 1500$ m (low elevation) | 0 | 47 | 386 | 0 | 0 | 0.87 | 2.89 | 72.42 |
| $1500\ m <= EL <= 2500$ m (mid elevation) | 0 | 2 | 46 | 0 | 0 | 0.94 | 2.96 | 74.63 |
| $EL > 2500$ m (high elevation) | 0 | 0 | 32 | 0 | 0 | 0.95 | 3.00 | 76.00 |
| Aridity Index<1 (Energy Limited) | 0 | 30 | 296 | 0 | 0 | 0.88 | 2.91 | 72.96 |
| Aridity Index>1 (Water Limited) | 0 | 19 | 168 | 0 | 0 | 0.87 | 2.90 | 72.65 |
| Snowy | 0 | 29 | 331 | 0 | 0 | 0.88 | 2.92 | 73.34 |
| Non-Snowy | 0 | 20 | 133 | 0 | 0 | 0.88 | 2.87 | 71.69 |
| $Max\ SWE \leq$ 100mm | 0 | 38 | 243 | 0 | 0 | 0.87 | 2.86 | 71.54 |
| 100mm < $Max\ SWE \leq$ 200mm | 0 | 4 | 55 | 0 | 0 | 0.86 | 2.93 | 73.76 |
| 200mm < $Max\ SWE \leq$ 400mm | 0 | 3 | 77 | 0 | 0 | 0.89 | 2.96 | 74.76 |
| 400mm < $Max\ SWE \leq$ 600mm | 0 | 0 | 27 | 0 | 0 | 0.93 | 3.00 | 76.00 |
| $Max\ SWE$ > 600mm | 0 | 4 | 62 | 0 | 0 | 0.94 | 2.94 | 74.00 |
| Forest | 0 | 23 | 259 | 0 | 0 | 0.89 | 2.92 | 73.31 |
| Open | 0 | 26 | 205 | 0 | 0 | 0.87 | 2.89 | 72.29 |
| Climate Zone: Arid | 0 | 1 | 34 | 0 | 0 | 0.86 | 2.97 | 75.06 |
| Climate Zone: Temperate | 0 | 29 | 213 | 0 | 0 | 0.88 | 2.88 | 72.05 |
| Climate Zone: Cold | 0 | 19 | 215 | 0 | 0 | 0.89 | 2.92 | 73.32 |
| Climate Zone: Polar | 0 | 0 | 2 | 0 | 0 | 0.97 | 3.00 | 76.00 |

**Table S41 | KGE-based architecture selection within the LSTM-only candidate space with a maximum of 4 states.** For each basin grouping, the table reports the number of catchments assigned to each LSTM state dimensionality, median testing-period $KGE_{ss}$, mean selected state dimensionality and mean number of trainable parameters.

| Regional Categorization | KGE-based model selection | | | | | | | |
|---|---|---|---|---|---|---|---|---|
| | $LSTM(1)$ | $LSTM(2)$ | $LSTM(3)$ | $LSTM(4)$ | $LSTM(5)$ | $KGE_{ss}^{50\%}$ | Ave State no | Ave Par no |
| CONUS | 0 | 15 | 114 | 384 | 0 | 0.88 | 3.72 | 105.73 |
| Eastern CONUS | 0 | 12 | 82 | 232 | 0 | 0.87 | 3.67 | 103.96 |
| Western CONUS | 0 | 3 | 32 | 152 | 0 | 0.92 | 3.80 | 108.80 |
| AM | 0 | 2 | 23 | 78 | 0 | 0.86 | 3.74 | 106.41 |
| RM | 0 | 0 | 0 | 54 | 0 | 0.96 | 4.00 | 117.00 |
| CRB | 0 | 0 | 5 | 29 | 0 | 0.95 | 3.85 | 110.97 |
| SN | 0 | 0 | 4 | 6 | 0 | 0.94 | 3.60 | 100.60 |
| CC | 0 | 0 | 5 | 28 | 0 | 0.90 | 3.85 | 110.79 |
| $EL < 1500$ m (low elevation) | 0 | 15 | 106 | 312 | 0 | 0.87 | 3.69 | 104.40 |
| $1500\ m <= EL <= 2500$ m (mid elevation) | 0 | 0 | 7 | 41 | 0 | 0.95 | 3.85 | 111.02 |
| $EL > 2500$ m (high elevation) | 0 | 0 | 1 | 31 | 0 | 0.96 | 3.97 | 115.72 |
| Aridity Index<1 (Energy Limited) | 0 | 8 | 78 | 240 | 0 | 0.89 | 3.71 | 105.37 |
| Aridity Index>1 (Water Limited) | 0 | 7 | 36 | 144 | 0 | 0.87 | 3.73 | 106.34 |
| Snowy | 0 | 9 | 67 | 284 | 0 | 0.89 | 3.76 | 107.52 |
| Non-Snowy | 0 | 6 | 47 | 100 | 0 | 0.88 | 3.61 | 101.50 |
| $Max\ SWE \leq$ 100mm | 0 | 12 | 85 | 184 | 0 | 0.87 | 3.61 | 101.44 |
| 100mm < $Max\ SWE \leq$ 200mm | 0 | 3 | 8 | 48 | 0 | 0.87 | 3.76 | 107.68 |
| 200mm < $Max\ SWE \leq$ 400mm | 0 | 0 | 13 | 67 | 0 | 0.90 | 3.84 | 110.34 |
| 400mm < $Max\ SWE \leq$ 600mm | 0 | 0 | 4 | 23 | 0 | 0.94 | 3.85 | 110.93 |
| $Max\ SWE$ > 600mm | 0 | 0 | 4 | 62 | 0 | 0.94 | 3.94 | 114.52 |
| Forest | 0 | 6 | 60 | 216 | 0 | 0.90 | 3.74 | 106.70 |
| Open | 0 | 9 | 54 | 168 | 0 | 0.87 | 3.69 | 104.53 |
| Climate Zone: Arid | 0 | 0 | 6 | 29 | 0 | 0.85 | 3.83 | 109.97 |
| Climate Zone: Temperate | 0 | 9 | 72 | 161 | 0 | 0.88 | 3.63 | 102.05 |
| Climate Zone: Cold | 0 | 6 | 36 | 192 | 0 | 0.89 | 3.79 | 108.79 |
| Climate Zone: Polar | 0 | 0 | 0 | 2 | 0 | 0.97 | 4.00 | 117.00 |

**Table S42 | KGE-based architecture selection within the LSTM-only candidate space with a maximum of 5 states.** For each basin grouping, the table reports the number of catchments assigned to each LSTM state dimensionality, median testing-period $KGE_{ss}$, mean selected state dimensionality and mean number of trainable parameters.

| Regional Categorization | KGE-based model selection | | | | | | | |
|---|---|---|---|---|---|---|---|---|
| | $LSTM(1)$ | $LSTM(2)$ | $LSTM(3)$ | $LSTM(4)$ | $LSTM(5)$ | $KGE_{ss}^{50\%}$ | Ave State no | Ave Par no |
| CONUS | 0 | 5 | 53 | 136 | 319 | 0.89 | 4.50 | 142.51 |
| Eastern CONUS | 0 | 5 | 37 | 89 | 195 | 0.87 | 4.45 | 140.52 |
| Western CONUS | 0 | 0 | 16 | 47 | 124 | 0.92 | 4.58 | 145.98 |
| AM | 0 | 0 | 9 | 22 | 72 | 0.87 | 4.61 | 147.67 |
| RM | 0 | 0 | 0 | 9 | 45 | 0.96 | 4.83 | 157.83 |
| CRB | 0 | 0 | 1 | 8 | 25 | 0.95 | 4.71 | 151.82 |
| SN | 0 | 0 | 0 | 2 | 8 | 0.93 | 4.80 | 156.20 |
| CC | 0 | 0 | 1 | 12 | 20 | 0.91 | 4.58 | 145.45 |
| $EL < 1500$ m (low elevation) | 0 | 5 | 52 | 122 | 254 | 0.88 | 4.44 | 139.97 |
| $1500\ m <= EL <= 2500$ m (mid elevation) | 0 | 0 | 1 | 8 | 39 | 0.95 | 4.79 | 155.96 |
| $EL > 2500$ m (high elevation) | 0 | 0 | 0 | 6 | 26 | 0.96 | 4.81 | 156.81 |
| Aridity Index<1 (Energy Limited) | 0 | 2 | 37 | 88 | 199 | 0.89 | 4.48 | 141.80 |
| Aridity Index>1 (Water Limited) | 0 | 3 | 16 | 48 | 120 | 0.87 | 4.52 | 143.75 |
| Snowy | 0 | 4 | 25 | 86 | 245 | 0.89 | 4.59 | 146.68 |
| Non-Snowy | 0 | 1 | 28 | 50 | 74 | 0.88 | 4.29 | 132.71 |
| $Max\ SWE \leq$ 100mm | 0 | 5 | 44 | 82 | 150 | 0.87 | 4.34 | 135.42 |
| 100mm < $Max\ SWE \leq$ 200mm | 0 | 0 | 3 | 17 | 39 | 0.87 | 4.61 | 147.31 |
| 200mm < $Max\ SWE \leq$ 400mm | 0 | 0 | 6 | 17 | 57 | 0.90 | 4.64 | 148.84 |
| 400mm < $Max\ SWE \leq$ 600mm | 0 | 0 | 0 | 5 | 22 | 0.94 | 4.81 | 156.93 |
| $Max\ SWE$ > 600mm | 0 | 0 | 0 | 15 | 51 | 0.94 | 4.77 | 154.86 |
| Forest | 0 | 2 | 25 | 70 | 185 | 0.90 | 4.55 | 144.99 |
| Open | 0 | 3 | 28 | 66 | 134 | 0.87 | 4.43 | 139.49 |
| Climate Zone: Arid | 0 | 0 | 2 | 8 | 25 | 0.86 | 4.66 | 149.66 |
| Climate Zone: Temperate | 0 | 2 | 38 | 76 | 126 | 0.88 | 4.35 | 135.46 |
| Climate Zone: Cold | 0 | 3 | 13 | 51 | 167 | 0.89 | 4.63 | 148.74 |
| Climate Zone: Polar | 0 | 0 | 0 | 1 | 1 | 0.97 | 4.50 | 141.50 |

**Table S43 | Testing performance of KGE-selected best-performing representations across model families and hydroclimatic basin groupings.** Median testing-period $KGE_{ss}$ is reported for each model family across 513 CAMELS-US catchments and by geographic region, elevation, aridity, snow influence, land cover and climate zone.

| Regional Categorization | $MCPNet(1)$ | $MCPNet(2)$ | $MCPNet(3)$ | $MCPNet(4)$ | $MCPNet(5)$ | $OPTMCPNet$ | $OPTLSTM$ |
|---|---|---|---|---|---|---|---|
| CONUS | 0.82 | 0.89 | 0.89 | 0.90 | 0.90 | 0.90 | 0.90 |
| Eastern CONUS | 0.80 | 0.88 | 0.88 | 0.89 | 0.88 | 0.90 | 0.89 |
| Western CONUS | 0.86 | 0.91 | 0.92 | 0.92 | 0.93 | 0.93 | 0.93 |
| AM | 0.79 | 0.87 | 0.89 | 0.89 | 0.88 | 0.89 | 0.88 |
| RM | 0.89 | 0.95 | 0.96 | 0.96 | 0.97 | 0.97 | 0.96 |
| CRB | 0.83 | 0.91 | 0.94 | 0.94 | 0.95 | 0.95 | 0.95 |
| SN | 0.85 | 0.93 | 0.94 | 0.93 | 0.94 | 0.95 | 0.94 |
| CC | 0.88 | 0.91 | 0.92 | 0.92 | 0.92 | 0.92 | 0.91 |
| $EL < 1500$ m (low elevation) | 0.81 | 0.88 | 0.89 | 0.89 | 0.89 | 0.90 | 0.89 |
| $1500\ m <= EL <= 2500$ m (mid elevation) | 0.85 | 0.92 | 0.94 | 0.94 | 0.95 | 0.95 | 0.95 |
| $EL > 2500$ m (high elevation) | 0.90 | 0.95 | 0.96 | 0.96 | 0.96 | 0.97 | 0.97 |
| Aridity Index<1 (Energy Limited) | 0.82 | 0.89 | 0.90 | 0.90 | 0.90 | 0.91 | 0.91 |
| Aridity Index>1 (Water Limited) | 0.79 | 0.86 | 0.87 | 0.87 | 0.88 | 0.89 | 0.90 |
| Snowy | 0.81 | 0.89 | 0.90 | 0.90 | 0.90 | 0.91 | 0.90 |
| Non-Snowy | 0.82 | 0.88 | 0.88 | 0.88 | 0.88 | 0.90 | 0.90 |
| $Max\ SWE \leq$ 100mm | 0.82 | 0.87 | 0.87 | 0.88 | 0.88 | 0.89 | 0.89 |
| 100mm $< Max\ SWE \leq$ 200mm | 0.78 | 0.85 | 0.87 | 0.86 | 0.86 | 0.87 | 0.88 |
| 200mm $< Max\ SWE \leq$ 400mm | 0.79 | 0.90 | 0.92 | 0.91 | 0.92 | 0.92 | 0.90 |
| 400mm $< Max\ SWE \leq$ 600mm | 0.86 | 0.92 | 0.93 | 0.94 | 0.94 | 0.94 | 0.94 |
| $Max\ SWE >$ 600mm | 0.88 | 0.94 | 0.95 | 0.95 | 0.96 | 0.96 | 0.95 |
| Forest | 0.83 | 0.90 | 0.91 | 0.91 | 0.91 | 0.92 | 0.91 |
| Open | 0.80 | 0.86 | 0.87 | 0.87 | 0.87 | 0.89 | 0.89 |
| Climate Zone: Arid | 0.67 | 0.82 | 0.84 | 0.85 | 0.86 | 0.86 | 0.89 |
| Climate Zone: Temperate | 0.83 | 0.89 | 0.89 | 0.89 | 0.89 | 0.90 | 0.90 |
| Climate Zone: Cold | 0.80 | 0.89 | 0.90 | 0.90 | 0.91 | 0.91 | 0.90 |
| Climate Zone: Polar | 0.93 | 0.97 | 0.98 | 0.98 | 0.98 | 0.98 | 0.97 |

**Table S44 | Parameter complexity of KGE-selected best-performing representations across model families and hydroclimatic basin groupings.** Mean numbers of trainable parameters are reported across 513 CAMELS-US catchments and by geographic region, elevation, aridity, snow influence, land cover and climate zone. The five-state LSTM contains 166 trainable parameters.

| | $MCPNet(1)$ | $MCPNet(2)$ | $MCPNet(3)$ | $MCPNet(4)$ | $MCPNet(5)$ | $OPTMCPNet$ | $OPTLSTM$ |
|---|---|---|---|---|---|---|---|
| CONUS | 10.53 | 26.21 | 45.09 | 66.41 | 89.77 | 69.45 | 115.86 |
| Eastern CONUS | 10.42 | 25.72 | 44.21 | 65.33 | 88.30 | 65.35 | 113.57 |
| Western CONUS | 10.73 | 27.07 | 46.64 | 68.30 | 92.33 | 76.59 | 119.84 |
| AM | 10.56 | 26.80 | 46.17 | 69.55 | 93.54 | 70.98 | 113.84 |
| RM | 11.00 | 28.70 | 50.06 | 74.15 | 102.41 | 92.02 | 142.24 |
| CRB | 10.74 | 27.38 | 47.12 | 68.82 | 92.50 | 75.71 | 125.35 |
| SN | 10.70 | 28.00 | 47.40 | 69.60 | 94.50 | 84.10 | 102.20 |
| CC | 10.91 | 27.82 | 47.82 | 69.58 | 97.42 | 78.36 | 108.21 |
| $EL < 1500$ m (low elevation) | 10.45 | 25.84 | 44.31 | 65.19 | 87.97 | 65.98 | 112.82 |
| $1500\ m <= EL <= 2500$ m (mid elevation) | 10.94 | 27.90 | 49.00 | 72.08 | 97.50 | 84.00 | 127.50 |
| $EL > 2500$ m (high elevation) | 11.00 | 28.75 | 49.78 | 74.38 | 102.50 | 94.50 | 139.44 |
| Aridity Index<1 (Energy Limited) | 10.56 | 26.38 | 45.45 | 66.77 | 91.32 | 69.33 | 114.97 |
| Aridity Index>1 (Water Limited) | 10.49 | 25.93 | 44.47 | 65.78 | 87.06 | 69.66 | 117.40 |
| Snowy | 10.68 | 26.74 | 46.62 | 68.78 | 93.10 | 75.71 | 120.40 |
| Non-Snowy | 10.20 | 24.99 | 41.51 | 60.84 | 81.93 | 54.71 | 105.16 |
| $Max\ SWE \leq$ 100mm | 10.24 | 25.02 | 42.34 | 62.33 | 82.97 | 58.30 | 105.65 |
| 100mm $< Max\ SWE \leq$ 200mm | 10.95 | 26.61 | 46.98 | 70.24 | 95.08 | 78.12 | 125.17 |
| 200mm $< Max\ SWE \leq$ 400mm | 10.74 | 27.81 | 48.38 | 71.75 | 97.25 | 80.64 | 130.61 |
| 400mm $< Max\ SWE \leq$ 600mm | 11.00 | 28.41 | 49.22 | 70.67 | 100.37 | 85.74 | 134.19 |
| $Max\ SWE >$ 600mm | 10.95 | 28.11 | 49.45 | 72.12 | 100.53 | 88.95 | 125.61 |
| Forest | 10.61 | 26.58 | 45.71 | 67.29 | 91.99 | 71.21 | 116.66 |
| Open | 10.44 | 25.77 | 44.34 | 65.33 | 87.06 | 67.29 | 114.87 |
| Climate Zone: Arid | 10.31 | 26.00 | 44.31 | 66.17 | 84.86 | 66.00 | 124.40 |
| Climate Zone: Temperate | 10.34 | 25.25 | 42.63 | 62.13 | 83.90 | 58.20 | 102.76 |
| Climate Zone: Cold | 10.76 | 27.23 | 47.73 | 70.79 | 96.45 | 81.44 | 127.70 |
| Climate Zone: Polar | 11.00 | 28.00 | 48.00 | 76.00 | 102.50 | 88.00 | 166.00 |

**Table S45 | State dimensionality of KGE-selected best-performing representations across model families and hydroclimatic basin groupings.** Mean numbers of selected state variables are reported across 513 CAMELS-US catchments and by geographic region, elevation, aridity, snow influence, land cover and climate zone.

| | $MCPNet(1)$ | $MCPNet(2)$ | $MCPNet(3)$ | $MCPNet(4)$ | $MCPNet(5)$ | $OPTMCPNet$ | $OPTLSTM$ |
|---|---|---|---|---|---|---|---|
| CONUS | 1.00 | 2.00 | 3.00 | 4.00 | 5.00 | 3.91 | 3.87 |
| Eastern CONUS | 1.00 | 2.00 | 3.00 | 4.00 | 5.00 | 3.79 | 3.80 |
| Western CONUS | 1.00 | 2.00 | 3.00 | 4.00 | 5.00 | 4.12 | 3.98 |
| AM | 1.00 | 2.00 | 3.00 | 4.00 | 5.00 | 3.96 | 3.80 |
| RM | 1.00 | 2.00 | 3.00 | 4.00 | 5.00 | 4.59 | 4.50 |
| CRB | 1.00 | 2.00 | 3.00 | 4.00 | 5.00 | 4.06 | 4.12 |
| SN | 1.00 | 2.00 | 3.00 | 4.00 | 5.00 | 4.40 | 3.60 |
| CC | 1.00 | 2.00 | 3.00 | 4.00 | 5.00 | 4.12 | 3.70 |
| $EL < 1500$ m (low elevation) | 1.00 | 2.00 | 3.00 | 4.00 | 5.00 | 3.80 | 3.79 |
| $1500\ m <= EL <= 2500$ m (mid elevation) | 1.00 | 2.00 | 3.00 | 4.00 | 5.00 | 4.33 | 4.17 |
| $EL > 2500$ m (high elevation) | 1.00 | 2.00 | 3.00 | 4.00 | 5.00 | 4.69 | 4.44 |
| Aridity Index<1 (Energy Limited) | 1.00 | 2.00 | 3.00 | 4.00 | 5.00 | 3.88 | 3.84 |
| Aridity Index>1 (Water Limited) | 1.00 | 2.00 | 3.00 | 4.00 | 5.00 | 3.96 | 3.90 |
| Snowy | 1.00 | 2.00 | 3.00 | 4.00 | 5.00 | 4.12 | 3.97 |
| Non-Snowy | 1.00 | 2.00 | 3.00 | 4.00 | 5.00 | 3.41 | 3.63 |
| $Max\ SWE \leq$ 100mm | 1.00 | 2.00 | 3.00 | 4.00 | 5.00 | 3.58 | 3.62 |
| 100mm $<\ Max\ SWE\ \leq$ 200mm | 1.00 | 2.00 | 3.00 | 4.00 | 5.00 | 4.14 | 4.07 |
| 200mm $<\ Max\ SWE\ \leq$ 400mm | 1.00 | 2.00 | 3.00 | 4.00 | 5.00 | 4.24 | 4.21 |
| 400mm $<\ Max\ SWE\ \leq$ 600mm | 1.00 | 2.00 | 3.00 | 4.00 | 5.00 | 4.37 | 4.30 |
| $Max\ SWE$ > 600mm | 1.00 | 2.00 | 3.00 | 4.00 | 5.00 | 4.50 | 4.12 |
| Forest | 1.00 | 2.00 | 3.00 | 4.00 | 5.00 | 3.95 | 3.89 |
| Open | 1.00 | 2.00 | 3.00 | 4.00 | 5.00 | 3.86 | 3.84 |
| Climate Zone: Arid | 1.00 | 2.00 | 3.00 | 4.00 | 5.00 | 3.77 | 4.09 |
| Climate Zone: Temperate | 1.00 | 2.00 | 3.00 | 4.00 | 5.00 | 3.53 | 3.56 |
| Climate Zone: Cold | 1.00 | 2.00 | 3.00 | 4.00 | 5.00 | 4.31 | 4.14 |
| Climate Zone: Polar | 1.00 | 2.00 | 3.00 | 4.00 | 5.00 | 4.50 | 5.00 |

**Table S46 |** MCP-based model and network configurations evaluated in this study.

| Model/Network Group | Representative Model Names | Node Number | Sharing-Configuration |
|---|---|---|---|
| Single-Node HMCP | $HMCP(1, AL)$ | 1 | N/A |
| Single-Node SNOWMCP | $SNOWMCP(1, BQ)$ | 1 | |
| Single-Node HMCP with Mass Relaxation | $HMCP(1, ALMR)$ | 1 | |
| Two-node HYDROMCP (HMCP+SNOWMCP) P: Parallel (Bypass); S: Serial (Routing) + denotes loss-gate information augmentation using the Current time-step cell state. | $HYDROMCP(2, B_+^P)$ | 2 | None |
| | $HYDROMCP(2, B_+^S)$ | 2 | None |
| | $HYDROMCP(2, B_+^P, H^{SALO})$ | 2 | $H^{SALO}$: HMCP Augmented loss & output gate |
| | $HYDROMCP(2, B_+^P, H^{SALO})$ | 2 | |
| | $HYDROMCP(2, B_+^P, S^{SALO})$ | 2 | $S^{SALO}$: SNOWMCP Augmented loss & output gate |
| | $HYDROMCP(2, B_+^S, S^{SALO})$ | 2 | |
| | $HYDROMCP(2, B_+^P, HS^{SALO})$ | 2 | $HS^{SALO}$: Augmented for both HMCP & SNOWMCP |
| | $HYDROMCP(2, B_+^S, HS^{SALO})$ | 2 | |
| Single-Layer HMCP network | $HMCP_{None}^{DS}(N, AL)$ | $N = 2$ to 5 | None |
| | $HMCP_{SALO}^{DS}(N, AL)$ | $N = 2$ to 5 | Augmented loss & output gate |
| | $HMCP_{SALO_P}^{DS}(N, AL)$ | $N = 3$ to 5 | Partial augmentation with adjacent loss & output gate |
| Single-Layer SNOWMCP Network *2LO denotes a two-output linear layer with softmax activation for partitioning precipitation bypass and snowmelt fluxes. | $SNOWMCP_{None}^{DS}(N, BQ)$ | $N = 2$ to 5 | None |
| | $SNOWMCP_{SALO}^{DS}(N, BQ)$ | $N = 2$ to 5 | Augmented loss & output gate |
| | $SNOWMCP_{SALO_{2LO}}^{DS}(N, BQ)$ | $N = 2$ to 5 | |
| Single-Layer HMCP with mass-relaxation network | $HMCP_{None}^{DS}(N, ALMR)$ | $N = 2$ to 5 | None |
| | $HMCP_{SALO}^{DS}(N, ALMR)$ | $N = 2$ to 5 | Augmented loss & output gate |
| Two-node HYDROMCP (HMCP+SNOWMCP) Network + denotes loss-gate information augmentation using the current time-step cell state | $HYDROMCP(2, NB_+^P)$ | 2 | None |
| | $HYDROMCP(2, NB_+^P, H^{SALO})$ | 2 | $H^{SALO}$: HMCP Augmented loss & output gate |
| | $HYDROMCP(2, NB_+^P, S^{SALO})$ | 2 | $S^{SALO}$: SNOWMCP Augmented loss & output gate |
| | $HYDROMCP(2, NB_+^P, HS^{SALO})$ | 2 | $HS^{SALO}$: Augmented for both HMCP & SNOWMCP |
| Single-Layer Mixed HMCP–SNOWMCP network | $H(2)SNOW(1)_{None}^{DS}$ | 3 | None |
| | $H(2)SNOW(1)_{SALO}^{DS}$ | 3 | Augmented for both HMCP & SNOWMCP |
| | $H(1)SNOW(2)_{None}^{DS}$ | 3 | None |
| | $H(1)SNOW(2)_{SALO}^{DS}$ | 3 | Augmented for both HMCP & SNOWMCP |
| Single-Layer LSTM benchmark | $LSTM(N)$ | $N = 1$ to 5 | Sharing among Input/Forget/Output gate and cell state update |

**Table S47 |** Model and network classes and scientific questions addressed by the experimental framework.

| Model/Network class | Scientific question addressed |
|---|---|
| $HMCP$ / $SNOWMCP$ | Is a single dominant storage state sufficient? |
| $HYDROMCP$ model | Does explicit snow–soil process coupling improve streamflow prediction? |
| $HYDROMCP$ network | Can snow–water dynamics be represented through functional approximation? |
| $DS$ multi-node networks | How many flow paths or storage states are needed? |
| State Information-Sharing ($SALO$ / $SALO_P$/ $SALO_{2LO}$ etc.) | Do state sharing and output partitioning improve predictive accuracy and process-relevant representation discovery? |

**Table S48 | Variability-component performance of representative MCP architectures across hydroclimatic basin groupings.** Median $\alpha_*^{KGE}$ is reported across 513 CAMELS-US catchments and by geographic region, elevation, aridity, snow influence, land cover and climate zone, where $\alpha_*^{KGE} = 1 - | 1 - \alpha^{KGE} |$ and values closer to 1 indicate better performance.

| | $HMCP$ $(1, AL)$ | $SNOWMCP$ $(1, BQ)$ | $HMCP$ $(1, ALMR)$ | $M_{Best}^{HYDRO}$ | $M_{Best}^{HYDRO-Net}$ | $HMCP_{none}^{DS}$ $(3, AL)$ | $HMCP_{SALO}^{DS}$ $(5, AL)$ | $SNOWMCP_{none}^{DS}$ $(5, BQ)$ | $SNOWMCP_{SALO}^{DS}$ $(3, BQ)$ | $HMCP_{none}^{DS}$ $(4, ALMR)$ | $HMCP_{SALO}^{DS}$ $(3, ALMR)$ |
|---|---|---|---|---|---|---|---|---|---|---|---|
| CONUS | 0.92 | 0.95 | 0.94 | 0.96 | 0.96 | 0.92 | 0.95 | 0.95 | 0.95 | 0.93 | 0.95 |
| Eastern CONUS | 0.93 | 0.94 | 0.93 | 0.95 | 0.95 | 0.92 | 0.93 | 0.94 | 0.94 | 0.93 | 0.93 |
| Western CONUS | 0.90 | 0.97 | 0.94 | 0.97 | 0.98 | 0.91 | 0.96 | 0.97 | 0.97 | 0.95 | 0.97 |
| AM | 0.93 | 0.96 | 0.94 | 0.95 | 0.96 | 0.93 | 0.93 | 0.95 | 0.96 | 0.93 | 0.94 |
| RM | 0.81 | 0.99 | 0.96 | 0.98 | 0.99 | 0.77 | 0.98 | 0.99 | 0.99 | 0.95 | 0.99 |
| CRB | 0.79 | 0.98 | 0.93 | 0.97 | 0.98 | 0.73 | 0.97 | 0.97 | 0.98 | 0.94 | 0.97 |
| SN | 0.81 | 0.98 | 0.88 | 0.97 | 0.96 | 0.79 | 0.96 | 0.98 | 0.98 | 0.88 | 0.96 |
| CC | 0.94 | 0.96 | 0.95 | 0.98 | 0.98 | 0.95 | 0.96 | 0.97 | 0.97 | 0.96 | 0.96 |
| $EL < 1500$ m (low elevation) | 0.93 | 0.94 | 0.94 | 0.95 | 0.96 | 0.93 | 0.93 | 0.94 | 0.94 | 0.93 | 0.94 |
| $1500\ m <= EL <= 2500$ m (mid elevation) | 0.86 | 0.98 | 0.92 | 0.98 | 0.98 | 0.86 | 0.98 | 0.98 | 0.98 | 0.93 | 0.98 |
| $EL > 2500$ m (high elevation) | 0.65 | 0.98 | 0.96 | 0.98 | 0.98 | 0.66 | 0.99 | 0.99 | 0.99 | 0.98 | 0.99 |
| Aridity Index<1 (Energy Limited) | 0.93 | 0.95 | 0.94 | 0.96 | 0.96 | 0.94 | 0.95 | 0.95 | 0.96 | 0.94 | 0.95 |
| Aridity Index>1 (Water Limited) | 0.89 | 0.95 | 0.92 | 0.96 | 0.96 | 0.87 | 0.95 | 0.95 | 0.92 | 0.92 | 0.95 |
| Snowy | 0.92 | 0.96 | 0.94 | 0.96 | 0.97 | 0.92 | 0.96 | 0.96 | 0.96 | 0.94 | 0.96 |
| Non-Snowy | 0.93 | 0.92 | 0.93 | 0.94 | 0.95 | 0.93 | 0.92 | 0.91 | 0.90 | 0.93 | 0.92 |
| $Max\ SWE \leq$ 100mm | 0.93 | 0.92 | 0.93 | 0.95 | 0.95 | 0.92 | 0.92 | 0.93 | 0.91 | 0.93 | 0.93 |
| 100mm $< Max\ SWE \leq$ 200mm | 0.94 | 0.94 | 0.93 | 0.95 | 0.96 | 0.93 | 0.94 | 0.94 | 0.96 | 0.94 | 0.95 |
| 200mm $< Max\ SWE \leq$ 400mm | 0.92 | 0.97 | 0.95 | 0.97 | 0.97 | 0.93 | 0.97 | 0.96 | 0.97 | 0.95 | 0.96 |
| 400mm $< Max\ SWE \leq$ 600mm | 0.93 | 0.98 | 0.96 | 0.98 | 0.98 | 0.93 | 0.97 | 0.98 | 0.98 | 0.96 | 0.98 |
| $Max\ SWE >$ 600mm | 0.81 | 0.98 | 0.95 | 0.98 | 0.98 | 0.83 | 0.98 | 0.98 | 0.98 | 0.94 | 0.98 |
| Forest | 0.93 | 0.96 | 0.94 | 0.96 | 0.97 | 0.93 | 0.95 | 0.96 | 0.96 | 0.94 | 0.95 |
| Open | 0.90 | 0.93 | 0.93 | 0.96 | 0.95 | 0.90 | 0.95 | 0.94 | 0.94 | 0.93 | 0.94 |
| Climate Zone: Arid | 0.89 | 0.92 | 0.91 | 0.95 | 0.94 | 0.87 | 0.95 | 0.95 | 0.92 | 0.89 | 0.96 |
| Climate Zone: Temperate | 0.93 | 0.94 | 0.94 | 0.95 | 0.95 | 0.94 | 0.93 | 0.94 | 0.93 | 0.93 | 0.93 |
| Climate Zone: Cold | 0.91 | 0.96 | 0.94 | 0.97 | 0.97 | 0.91 | 0.96 | 0.96 | 0.97 | 0.94 | 0.97 |
| Climate Zone: Polar | 0.47 | 0.99 | 0.98 | 0.99 | 0.99 | 0.49 | 0.99 | 0.99 | 0.99 | 0.98 | 0.99 |
| No. of Parameters | 8 | 11 | 11 | 8 to 23 | 8 to 25 | 27 | 85 | 60 | 48 | 48 | 48 |
| No. of State Variables | 1 | 1 | 1 | 1 to 2 | 1 to 2 | 3 | 5 | 5 | 3 | 4 | 3 |

**Table S49 | Bias-component performance of representative MCP architectures across hydroclimatic basin groupings.** Median $\beta_*^{KGE}$ is reported across 513 CAMELS-US catchments and by geographic region, elevation, aridity, snow influence, land cover and climate zone, where $\beta_*^{KGE} = 1 - | 1 - \beta^{KGE} |$ and values closer to 1 indicate better performance.

| | $HMCP$ $(1, AL)$ | $SNOWMCP$ $(1, BQ)$ | $HMCP$ $(1, ALMR)$ | $M_{Best}^{HYDRO}$ | $M_{Best}^{HYDRO-Net}$ | $HMCP_{none}^{DS}$ $(3, AL)$ | $HMCP_{SALO}^{DS}$ $(5, AL)$ | $SNOWMCP_{none}^{DS}$ $(5, BQ)$ | $SNOWMCP_{SALO}^{DS}$ $(3, BQ)$ | $HMCP_{none}^{DS}$ $(4, ALMR)$ | $HMCP_{SALO}^{DS}$ $(3, ALMR)$ |
|---|---|---|---|---|---|---|---|---|---|---|---|
| CONUS | 0.96 | 0.97 | 0.97 | 0.98 | 0.98 | 0.97 | 0.98 | 0.97 | 0.98 | 0.98 | 0.98 |
| Eastern CONUS | 0.97 | 0.97 | 0.97 | 0.98 | 0.98 | 0.98 | 0.98 | 0.97 | 0.97 | 0.97 | 0.97 |
| Western CONUS | 0.95 | 0.98 | 0.97 | 0.99 | 0.99 | 0.97 | 0.99 | 0.98 | 0.98 | 0.98 | 0.99 |
| AM | 0.97 | 0.98 | 0.98 | 0.98 | 0.98 | 0.98 | 0.98 | 0.98 | 0.99 | 0.98 | 0.98 |
| RM | 0.91 | 0.99 | 0.97 | 0.99 | 0.99 | 0.90 | 0.99 | 0.99 | 0.99 | 0.96 | 0.99 |
| CRB | 0.92 | 0.97 | 0.96 | 0.98 | 0.99 | 0.91 | 0.99 | 0.97 | 0.98 | 0.95 | 0.98 |
| SN | 0.96 | 0.97 | 0.97 | 0.98 | 0.98 | 0.98 | 0.99 | 0.98 | 0.98 | 0.95 | 0.98 |
| CC | 0.96 | 0.98 | 0.98 | 0.99 | 0.99 | 0.97 | 0.98 | 0.99 | 0.99 | 0.98 | 0.99 |
| $EL < 1500$ m (low elevation) | 0.97 | 0.97 | 0.97 | 0.98 | 0.98 | 0.98 | 0.98 | 0.97 | 0.97 | 0.98 | 0.98 |
| $1500\ m <= EL <= 2500$ m (mid elevation) | 0.95 | 0.98 | 0.97 | 0.99 | 0.99 | 0.96 | 0.99 | 0.98 | 0.99 | 0.96 | 0.99 |
| $EL > 2500$ m (high elevation) | 0.91 | 0.99 | 0.97 | 0.98 | 0.99 | 0.90 | 0.99 | 0.99 | 0.99 | 0.98 | 0.99 |
| Aridity Index<1 (Energy Limited) | 0.97 | 0.98 | 0.98 | 0.98 | 0.98 | 0.98 | 0.98 | 0.98 | 0.98 | 0.98 | 0.98 |
| Aridity Index>1 (Water Limited) | 0.95 | 0.96 | 0.95 | 0.97 | 0.98 | 0.96 | 0.97 | 0.96 | 0.97 | 0.96 | 0.97 |
| Snowy | 0.96 | 0.98 | 0.97 | 0.98 | 0.98 | 0.97 | 0.98 | 0.98 | 0.98 | 0.98 | 0.98 |
| Non-Snowy | 0.96 | 0.94 | 0.97 | 0.98 | 0.98 | 0.97 | 0.97 | 0.96 | 0.95 | 0.97 | 0.98 |
| $Max\ SWE \leq$ 100mm | 0.96 | 0.95 | 0.97 | 0.97 | 0.98 | 0.97 | 0.97 | 0.96 | 0.96 | 0.97 | 0.97 |
| 100mm < $Max\ SWE \leq$ 200mm | 0.98 | 0.98 | 0.98 | 0.98 | 0.98 | 0.98 | 0.98 | 0.98 | 0.98 | 0.98 | 0.98 |
| 200mm < $Max\ SWE \leq$ 400mm | 0.97 | 0.98 | 0.97 | 0.99 | 0.99 | 0.98 | 0.99 | 0.98 | 0.99 | 0.98 | 0.99 |
| 400mm < $Max\ SWE \leq$ 600mm | 0.97 | 0.99 | 0.98 | 0.98 | 0.99 | 0.96 | 0.98 | 0.99 | 0.99 | 0.99 | 0.99 |
| $Max\ SWE$ > 600mm | 0.89 | 0.98 | 0.97 | 0.99 | 0.99 | 0.93 | 0.99 | 0.99 | 0.99 | 0.98 | 0.99 |
| Forest | 0.97 | 0.98 | 0.98 | 0.98 | 0.98 | 0.98 | 0.98 | 0.98 | 0.98 | 0.98 | 0.98 |
| Open | 0.96 | 0.96 | 0.96 | 0.97 | 0.98 | 0.96 | 0.98 | 0.97 | 0.97 | 0.97 | 0.97 |
| Climate Zone: Arid | 0.93 | 0.95 | 0.92 | 0.97 | 0.95 | 0.93 | 0.94 | 0.95 | 0.95 | 0.94 | 0.96 |
| Climate Zone: Temperate | 0.97 | 0.96 | 0.97 | 0.98 | 0.98 | 0.98 | 0.98 | 0.97 | 0.97 | 0.98 | 0.98 |
| Climate Zone: Cold | 0.96 | 0.98 | 0.97 | 0.98 | 0.98 | 0.97 | 0.98 | 0.98 | 0.99 | 0.98 | 0.98 |
| Climate Zone: Polar | 0.78 | 0.99 | 0.98 | 0.98 | 0.99 | 0.80 | 0.99 | 1.00 | 0.99 | 0.99 | 0.99 |
| No. of Parameters | 8 | 11 | 11 | 8 to 23 | 8 to 25 | 27 | 85 | 60 | 48 | 48 | 48 |
| No. of State Variables | 1 | 1 | 1 | 1 to 2 | 1 to 2 | 3 | 5 | 5 | 3 | 4 | 3 |

**Table S50 | Correlation performance of representative MCP architectures across hydroclimatic basin groupings.** Median $r^{KGE}$ is reported across 513 CAMELS-US catchments and by geographic region, elevation, aridity, snow influence, land cover and climate zone.

| | $HMCP$ $(1, AL)$ | $SNOWMCP$ $(1, BQ)$ | $HMCP$ $(1, ALMR)$ | $M^{HYDRO}_{Best}$ | $M^{HYDRO-Net}_{Best}$ | $HMCP^{DS}_{none}$ $(3, AL)$ | $HMCP^{DS}_{SALO}$ $(5, AL)$ | $SNOWMCP^{DS}_{none}$ $(5, BQ)$ | $SNOWMCP^{DS}_{SALO}$ $(3, BQ)$ | $HMCP^{DS}_{none}$ $(4, ALMR)$ | $HMCP^{DS}_{SALO}$ $(3, ALMR)$ |
|---|---|---|---|---|---|---|---|---|---|---|---|
| CONUS | 0.67 | 0.71 | 0.69 | 0.81 | 0.82 | 0.72 | 0.82 | 0.77 | 0.81 | 0.75 | 0.82 |
| Eastern CONUS | 0.70 | 0.68 | 0.71 | 0.79 | 0.81 | 0.76 | 0.82 | 0.75 | 0.78 | 0.77 | 0.80 |
| Western CONUS | 0.56 | 0.80 | 0.66 | 0.86 | 0.85 | 0.62 | 0.85 | 0.84 | 0.89 | 0.71 | 0.83 |
| AM | 0.67 | 0.70 | 0.67 | 0.77 | 0.79 | 0.73 | 0.78 | 0.77 | 0.83 | 0.72 | 0.77 |
| RM | 0.46 | 0.85 | 0.61 | 0.91 | 0.90 | 0.47 | 0.90 | 0.87 | 0.95 | 0.65 | 0.89 |
| CRB | 0.45 | 0.76 | 0.58 | 0.86 | 0.85 | 0.47 | 0.89 | 0.77 | 0.90 | 0.64 | 0.86 |
| SN | 0.49 | 0.81 | 0.59 | 0.88 | 0.88 | 0.49 | 0.89 | 0.85 | 0.92 | 0.61 | 0.82 |
| CC | 0.75 | 0.84 | 0.77 | 0.85 | 0.86 | 0.78 | 0.81 | 0.86 | 0.87 | 0.81 | 0.82 |
| $EL < 1500$ m (low elevation) | 0.70 | 0.69 | 0.71 | 0.80 | 0.81 | 0.76 | 0.81 | 0.76 | 0.80 | 0.77 | 0.80 |
| $1500\ m <= EL <= 2500$ m (mid elevation) | 0.47 | 0.79 | 0.57 | 0.85 | 0.85 | 0.50 | 0.88 | 0.81 | 0.90 | 0.65 | 0.85 |
| $EL > 2500$ m (high elevation) | 0.46 | 0.86 | 0.63 | 0.91 | 0.90 | 0.46 | 0.91 | 0.88 | 0.94 | 0.66 | 0.88 |
| Aridity Index<1 (Energy Limited) | 0.71 | 0.72 | 0.73 | 0.82 | 0.84 | 0.78 | 0.83 | 0.80 | 0.84 | 0.79 | 0.82 |
| Aridity Index>1 (Water Limited) | 0.54 | 0.65 | 0.63 | 0.78 | 0.79 | 0.62 | 0.82 | 0.71 | 0.65 | 0.69 | 0.81 |
| Snowy | 0.62 | 0.73 | 0.66 | 0.81 | 0.82 | 0.69 | 0.82 | 0.78 | 0.84 | 0.73 | 0.81 |
| Non-Snowy | 0.76 | 0.57 | 0.77 | 0.81 | 0.82 | 0.81 | 0.84 | 0.72 | 0.66 | 0.82 | 0.84 |
| $Max\ SWE \leq$ 100mm | 0.74 | 0.64 | 0.75 | 0.80 | 0.81 | 0.80 | 0.83 | 0.74 | 0.70 | 0.80 | 0.82 |
| 100mm $< Max\ SWE \leq$ 200mm | 0.61 | 0.67 | 0.63 | 0.75 | 0.78 | 0.68 | 0.78 | 0.73 | 0.80 | 0.71 | 0.77 |
| 200mm $< Max\ SWE \leq$ 400mm | 0.61 | 0.70 | 0.63 | 0.78 | 0.80 | 0.68 | 0.81 | 0.77 | 0.87 | 0.71 | 0.79 |
| 400mm $< Max\ SWE \leq$ 600mm | 0.53 | 0.81 | 0.63 | 0.86 | 0.85 | 0.56 | 0.86 | 0.83 | 0.90 | 0.66 | 0.85 |
| $Max\ SWE >$ 600mm | 0.50 | 0.84 | 0.65 | 0.89 | 0.88 | 0.51 | 0.89 | 0.87 | 0.93 | 0.68 | 0.86 |
| Forest | 0.70 | 0.72 | 0.71 | 0.83 | 0.84 | 0.75 | 0.84 | 0.80 | 0.84 | 0.77 | 0.82 |
| Open | 0.62 | 0.69 | 0.66 | 0.78 | 0.80 | 0.67 | 0.81 | 0.72 | 0.74 | 0.72 | 0.80 |
| Climate Zone: Arid | 0.46 | 0.39 | 0.55 | 0.73 | 0.72 | 0.57 | 0.81 | 0.58 | 0.58 | 0.63 | 0.75 |
| Climate Zone: Temperate | 0.76 | 0.70 | 0.77 | 0.82 | 0.83 | 0.81 | 0.83 | 0.77 | 0.75 | 0.82 | 0.83 |
| Climate Zone: Cold | 0.59 | 0.72 | 0.64 | 0.80 | 0.81 | 0.65 | 0.82 | 0.78 | 0.86 | 0.70 | 0.80 |
| Climate Zone: Polar | 0.44 | 0.91 | 0.75 | 0.91 | 0.92 | 0.42 | 0.94 | 0.92 | 0.97 | 0.75 | 0.85 |
| No. of Parameters | 8 | 11 | 11 | 8 to 23 | 8 to 25 | 27 | 85 | 60 | 48 | 48 | 48 |
| No. of State Variables | 1 | 1 | 1 | 1 to 2 | 1 to 2 | 3 | 5 | 5 | 3 | 4 | 3 |